\documentclass[review]{cas-sc}
\usepackage[nolist]{acronym}
\usepackage{amsmath,amssymb,amsfonts}
\usepackage[para,online,flushleft]{threeparttablex}
\usepackage{multirow}
\usepackage{textcomp}
\usepackage{xcolor}
\usepackage[linesnumbered,ruled,boxed,commentsnumbered,noend]{algorithm2e}

\SetKwComment{Comment}{$\triangleright$\ }{}

\usepackage{tikz}
\usepackage{pgfplots}
\usepackage{pgf-pie}
\usetikzlibrary{plotmarks}
\pgfplotsset{compat=newest}
\usetikzlibrary{patterns}
\usetikzlibrary{shapes,arrows}

\usepackage{natbib}

\usepackage{epsfig}
\usepackage{epstopdf}
\usepackage{graphicx}
\usepackage{amssymb}
\usepackage{array}
\usepackage{arydshln}
\usepackage[utf8]{inputenc}
\usepackage{balance}
\usepackage{url}
\usepackage{float}

\usepackage{makecell}

\usetikzlibrary{plotmarks}
\usetikzlibrary{patterns}
\usetikzlibrary{pgfplots.dateplot}
\usetikzlibrary{external}
\usetikzlibrary{quotes, angles}
\usetikzlibrary{shapes.geometric}
\usetikzlibrary{external}
\usetikzlibrary{calc}

\usepackage{booktabs}

\usepackage{pgf-pie}
\usepackage{etoolbox}
\newtoggle{showpct}
\makeatletter
\patchcmd{\pgfpie@slice}%
{\scalefont{#3}\beforenumber#3\afternumber}%
{\iftoggle{showpct}{\scalefont{#3}\beforenumber#3\afternumber}{}}%
{}{}
\makeatother

\usepackage{subfig}

\newlength\figureheight
\newlength\figurewidth

\makeatletter
\def\endthebibliography{%
	\def\@noitemerr{\@latex@warning{Empty `thebibliography' environment}}%
	\endlist
}
\makeatother

\usepackage{todonotes}

\begin{document}

\begin{acronym}[TDMA]
    \acro{DCS}{Dynamic Classifier Selection}
    \acro{DES}{Dynamic Ensemble Selection}
    \acro{DS}{Dynamic Selection}
    \acro{DSEL}{Dynamic Selection Set}
    \acro{DSEW}{Dynamic Selection Window}
    \acro{DDM}{Drift Detection Method}
    \acro{EDDM}{Early Drift Detection Method}
    \acro{ADWIN}{Adaptive Windowing}
    \acro{RDDM}{Reactive Drift Detection Method}
    \acro{HT}{Hoeffding Tree}
    \acro{RoC}{Region of Competence}
    \acro{kNN}{k-Nearest Neighbors}
    \acro{ML}{Machine Learning}
    \acro{STEPD}{Statistical Test of Equal Proportions Detector}
    \acro{DDCS}{Double Dynamic Classifier Selection}
    \acro{DES-ICD}{Dynamic Ensemble Selection for Imbalanced Data Streams with Concept Drift}
    \acro{DESW-ID}{Dynamic Ensemble Selection based on Window over Imbalanced Drift Data Stream}
    \acro{DynEd}{Dynamic Ensemble Diversification}
    \acro{LAE}{Local Accuracy Estimates}
    \acro{OLA}{Overall Local Accuracy}
    \acro{KNORA}{k-Nearest Oracles}
    \acro{KNORAE}{KNORA-Eliminate}
    \acro{KNORAU}{KNORA-Union}
    \acro{PH}{Page-Hinkley Test}
    \acro{HDDM}{Hoeffding Drift Detection Method}
    \acro{ACDDM}{Accurate Concept Drift Detection Method}
    \acro{ECDD}{EWMA for Concept Drift Detection}
    \acro{FHDDM}{Fast Hoeffding Drift Detection Method}
    \acro{FHDDMS}{Stacking Fast Hoeffding Drift Detection Method}
    \acro{FHDDMS$_{add}$}{Additive FHDDMS}
    \acro{STUDD}{Student-Teacher approach for Unsupervised Drift Detection}
    \acro{iid}{Independent and Identically Distributed}
    \acro{MDDM}{McDiarmid Drift Detection Method}
    \acro{UDD}{Uncertainty Drift Detection}
    \acro{LD3}{Label Dependency Drift Detector}
    \acro{MSTS}{Mass-Based Short Term Selection}
    \acro{kdtree}{k-dimensional tree}
    \acro{IncA-DES}{Incremental Adaptive Dynamic Ensemble Selection}
    \acro{AB-DES}{Adaptive Bagging-based Dynamic Ensemble Selection}
    \acro{ARF}{Adaptive Random Forest}
    \acro{kDN}{k-Disagreeing Neighbors}
    \acro{AO-DCS}{Attribute-Oriented Dynamic Classifier Selection}
    \acro{OAUE}{Online Accuracy Updated Ensemble}
    \acro{MOA}{Massive Online Analysis}
    \acro{ARTE}{Adaptive Random Tree Ensemble}
    \acro{DDD}{Diversity for Dealing with Drifts}
    \acro{ARE}{Adaptive Regularized Ensemble}
    \acro{EOCD}{Ensemble Optimization for Concept Drift}
    \acro{CALMID}{Comprehensive Active Learning Framework for Multiclass Imbalanced Data Streams with Concept Drift}
    \acro{SEA}{Streaming Ensemble Algorithm}
    \acro{AWE}{Accuracy Weighted Ensemble}
    \acro{AUE}{Accuracy Updated Ensemble}
    \acro{SEOA}{Selective Ensemble-based Online Adaptive Neural Network}
    \acro{HE-CDTL}{Hybrid Ensemble approach to concept drift-tolerate transfer learning}
    \acro{CCEKL}{Cost-sensitive Continuous Ensemble Kernel Learning method}
    \acro{SCHCDOE}{Streaming data classification algorithm based on hierarchical concept drift and online ensemble}
\end{acronym}

\bibliographystyle{cas-model2-names}
\shorttitle{}

\title[mode=title]{IncA-DES: An incremental and adaptive dynamic ensemble selection approach using online K-d tree neighborhood search for data streams with concept drift.}

\shortauthors{E.V.L. Barboza et~al.}

\author[1,2]{Eduardo V.L. Barboza}\corref{correspondingauthor}
\cortext[correspondingauthor]{Corresponding author}
\ead{eduardo.lima-barboza.1@ens.etsmtl.ca}
\author[2]{Paulo R. Lisboa de Almeida}
\author[3]{Alceu de Souza Britto Jr.}
\author[1]{Robert Sabourin}
\author[1]{Rafael M.O. Cruz}

\address[1]{LIVIA, École de Technologie Supérieure, University of Qu\'ebec, Montreal, Qu\'ebec, Canada}

\address[2]{Department of Informatics, 
Universidade Federal do Paran\'{a} (UFPR), Curitiba (PR), Brazil}

\address[3]{Graduate Program in Informatics (PPGIa), Pontif\'{i}cia Universidade Cat\'{o}lica do Paran\'{a} (PUCPR), Curitiba (PR), Brazil}

\begin{abstract}
Data streams pose challenges not usually encountered in batch-based \ac{ML}. One of them is concept drift, which is characterized by the change in data distribution over time. Among many approaches explored in literature, the fusion of classifiers has been showing good results and is getting growing attention. More specifically, \ac{DS} methods, due to the ensemble being instance-based, seem to be a more efficient choice under drifting scenarios. However, some attention must be paid to adapting such methods for concept drift. The training must be done in order to create local experts, and the commonly used neighborhood-search \ac{DS} may become prohibitive with the continuous arrival of data. In this work, we propose \ac{IncA-DES}, which employs a training strategy that promotes the generation of local experts with the assumption that different regions of the feature space become available with time. Additionally, the fusion of a concept drift detector supports the maintenance of information and adaptation to a new concept. An overlap-based classification filter is also employed in order to avoid using the \ac{DS} method when there is a consensus in the neighborhood, a strategy that we argue every \ac{DS} method should employ, as it was shown to make them more applicable and quicker. Moreover, aiming to reduce the processing time of the \ac{kNN}, we propose an Online K-d tree algorithm, which can quickly remove instances without becoming inconsistent and deals with unbalancing concerns that may occur in data streams. Experimental results showed that the proposed framework got the best average accuracy compared to seven state-of-the-art methods considering different levels of label availability and presented the smaller processing time between the most accurate methods. Additionally, the fusion with the Online K-d tree has improved processing time with a negligible loss in accuracy. We have made our framework available in an online repository\footnote{https://github.com/eduardovlb/IncA-DES}.
\end{abstract}


\maketitle

\section{Introduction}\label{sec:introduction}

When dealing with data streams, where both training and testing instances arrive over time, we may encounter some scenarios usually not found in classic batch-based \acf{ML}. One of them is concept drift, where data distribution changes over time \citep{gama2014, lu2019}. Under such scenario, the \ac{ML} models must constantly learn with the newly arrived data and adapt to changing distributions, which may involve discard old and possibly conflicting information and/or to learn on the newest data. Concept drift may be encountered in, for instance, cybersecurity-related problems \citep{ceschinEtAl2020, shya2024}, climate prediction \citep{ditzlerEtAl2012}, spam prediction \citep{kuncheva2004, Mohammad2024}, and people's opinion \citep{muller2020}.

In literature, there are two main approaches to dealing with concept drift. In the first one, called the blind adaptation strategy, the classification models are kept up to date only with the most recent data, discarding past samples unaware of concept drift \citep{Guo2021, yang2022}. The second strategy, active strategy, tries to track concept drift, primarily by monitoring the classification error or perceiving changes in the data distribution through concept drift detectors, also known as triggers \citep{gomes2017, kozal2021}. 

Recently, \acf{DS}-based approaches have been getting growing attention for concept drift scenarios \citep{cavalheiro2021, Han2023, abadifard2023}, as they can select the most promising ensemble of classifiers based on the characteristics of the test instance. In many methods, the most competent candidate classifiers in the \ac{RoC} of the test instance, commonly gathered through a \ac{kNN} algorithm from the named \ac{DSEL}, are chosen according to their performance. The \ac{DSEL} is the set of labeled instances used by the \ac{DS} method for selecting the classifiers. Therefore, the candidate classifiers from the pool $C$ that are more competent in the \ac{RoC} are chosen to classify the test instance, and local experts of the \ac{RoC} are more likely to be chosen if they exist.

Given the possibility of concept drift, many \ac{DS}-based approaches rely on blind adaptation strategies, keeping a sliding window of recent instances or data chunks as a \ac{DSEL} \citep{almeidaEtAl2018, cavalheiro2021,jiao2022}, making \ac{DS} both space and time-dependent. Many of them employ the fusion of concept drift detectors as well \citep{jiao2022, abadifard2023, wei2024} to drop old instances from the \ac{DSEL} or to update the pool of candidate classifiers. To differ the sliding window employed in \ac{DS} for data streams to the \ac{DSEL} used in static \ac{DS}, we call it \ac{DSEW}.

\begin{figure}[!htb]
    \centering
    \includegraphics{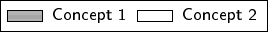}\\
    
    \subfloat[Big Sliding Window: Adaptation to a New Concept Takes Longer.]{
        \label{subfig:big-dsew}
        \includegraphics{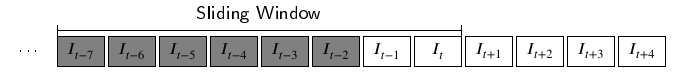}
        }\\
    \subfloat[Small Sliding Window: Adaptation to a New Concept is Quicker.]{
        \label{subfig:small-dsew}
        \includegraphics{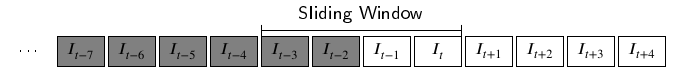}
        }\\
    \caption{The Impact of Window Size on Adapting to Concept Drift.}
    \label{fig:window-concept}
\end{figure}

However, the \ac{DSEW} may suffer from the well-known stability-plasticity dilemma problem \citep{elwell2011} since, on the one hand, a big \ac{DSEW} will have more information for the \ac{DS} mechanism, but adapting to a new concept may take longer. On the other hand, a smaller \ac{DSEW} may lead to a quicker adaption but results in a lack of information to estimate the \ac{RoC}. As exemplified in Figure \ref{fig:window-concept}, as the window slides through the new arriving instances, the old concept remains until the new concept has enough instances to overlap it completely. The larger the sliding window, the longer the adaptation will take. This can even become a bigger problem under a limited label availability that happens in real-world scenarios, mainly when human action is needed to label the data. Thus, leveraging helpful information while aware of possible changes is crucial in drifting scenarios.

Further, the training approach that \ac{DS} methods for data streams is usually employ is based on some resampling strategy, such as online bagging \citep{cavalheiro2021, abadifard2023}. However, such an approach may not be the best one for \ac{DS}. It has a global perspective of the data and does not guarantee the creation of local experts, preferred for ensemble candidates in \ac{DS} \citep{cruz2018}. \citet{cruz2018} also mentions that techniques leveraged from static \ac{ML} may not be the best choices for training models for \ac{DES} applications.

In this work, we propose \acf{IncA-DES}, which adopts an incremental training approach that favors local experts' generation as the feature space regions become available over time. Besides, considering a limited label availability, focusing the training on only one classifier may bring advantages compared to resampling, as the latter may cause classifiers not to receive enough information, leading to a pool of underfitted candidate classifiers.

In addition, concept drift is addressed through drift detection. The idea is to have a \ac{DSEW} that can keep as many instances as possible. When a drift detector triggers a concept drift, the \ac{DSEW} shrinks to the warning level given by the drift detector to adapt to the new concept. Thus, we can retain useful information and drop potentially outdated instances when a concept drift is detected. The rationale is to let the drift detector be in charge of dealing with \textit{real concept drift}, while under stable concepts or \textit{virtual concept drift}, information is leveraged for \ac{DS}. Furthermore, faced with the possibility of false alarms from the drift detector, old classifiers are still kept as ensemble candidates, making \ac{IncA-DES} robust in these scenarios.

To enhance the effectiveness of the proposed framework, we present the overlap-based classification, which can choose whether it is necessary to use the \ac{DS} method for classifying a test instance. As we use a neighborhood-based \ac{DS} method, and the \ac{kNN} suffers more on classifying regions with a higher instance hardness \citep{cruz2017}, we use the \ac{DS} method when the neighbors' classes are different. Otherwise, the \ac{kNN} is used. Experimental analysis has shown that this mechanism has substantially decreased the processing time of the framework, while improving accuracy performance. Thus, we argue that every neighborhood-based \ac{DS} method should employ this strategy.

\begin{figure}
    \centering
    \includegraphics[width=0.8\linewidth]{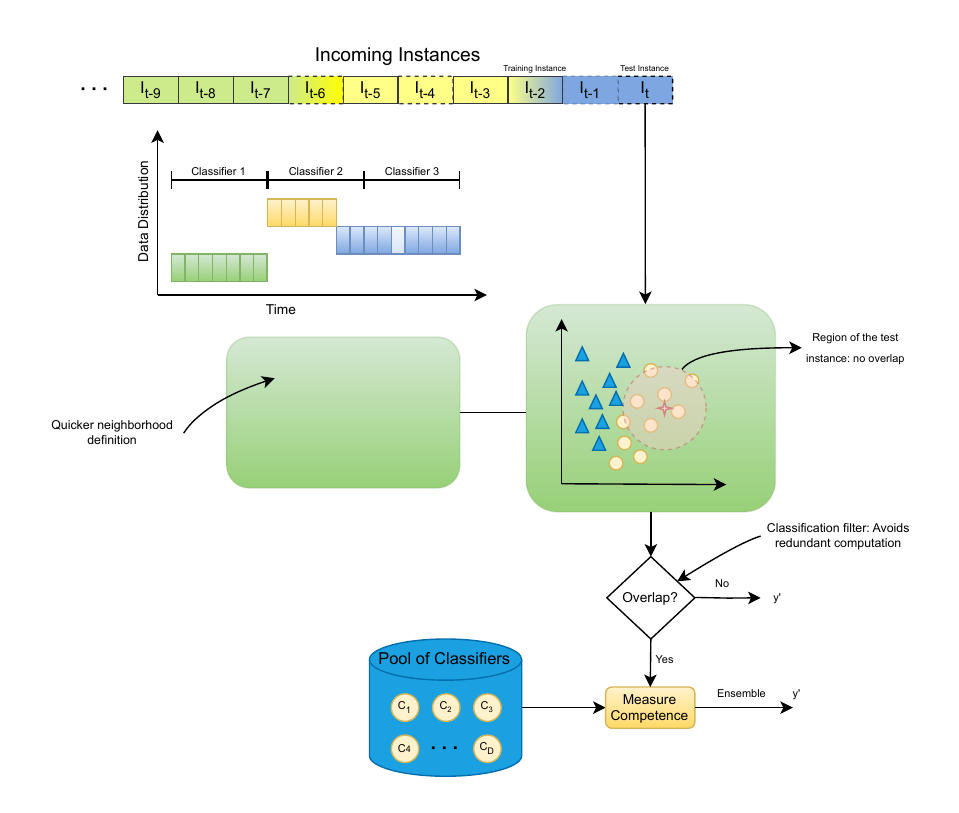}
    \caption{An overview of IncA-DES and its components. Notice that the incremental training policy aims to train classifiers in different perspectives of data and the local region has no class overlap.}
    \label{fig:running-fig}
\end{figure}

Still into the \ac{kNN} algorithm, many \ac{DS} methods employ it to define the \ac{RoC} to measure the competence of the classifiers \citep{almeidaEtAl2018, cavalheiro2021}. A brute force \ac{kNN} can be prohibitive under a streaming scenario, where decisions must be made in real-time due to the fast and possibly infinite number of instances arriving \citep{assis2023}. To address this, we propose an Online K-d tree algorithm for approximate neighborhood search. It can quickly process arriving instances and mitigate some problems found in the original K-d tree, such as deletion complexity, unbalancing with the addition of new nodes, and the need for normalized data. The presented Online K-d tree algorithm has promoted extensibility in the proposed framework, with a negligible loss in accuracy performance, which is expected in approximate neighborhood search. Figure \ref{fig:running-fig} shows an overview of the proposed framework. Notice that the classifiers are trained in different distributions of the data as they arrive over time.

Experimental analysis carried out on 22 datasets, including 11 real-world, 4 with an induced \textit{virtual concept drift}, and 7 synthetic datasets, has shown that the proposed \ac{IncA-DES} presented better accuracy performance in comparison with state-of-the-art methods, including \ac{DS} and ensemble methods for concept drift, in scenarios with different levels of label availability. Further, the fusion of the framework with our proposed Online K-d tree algorithm reduced the overall processing time compared to the brute force \ac{kNN}, with a negligible loss in accuracy. The overlap-based classification filter has enhanced effectiveness, bringing gain in both accuracy and speed.

Our main contributions are as follows:

\begin{itemize}
    \item We propose a novel framework based on \ac{DS} with a training policy that favors the generation of local experts and uses a drift detector to have an Adaptive \ac{DSEW}.
    \item We optimize the neighborhood search with an Online K-d Tree algorithm that can quickly process incoming instances in data streams.
    \item We apply a local overlap-based classification optimization of \ac{DS}, which has promoted effectiveness in the framework by improving the accuracy performance and speeding up the processing of instances.
    \item We perform experiments considering label availability, which better reflects environments found in the real world, comparing several \ac{DS} and ensemble methods for data streams with concept drift.
\end{itemize}

The remainder of this work is structured as follows: in Section \ref{sec:theoreticalFoundation}, we define key concepts used in this work, such as concept drift, batch processing, online processing, and basic concepts from \ac{DS}.  We present the proposed framework in Section \ref{sec:proposal}, in Section \ref{sec:stateoftheart}, we present related works in the literature on concept drift detection and ensemble learning for concept drift, perform experiments comparing with state of the art and an ablation study with the presented components in Section \ref{sec:experiments}, and conclude the paper in Section \ref{sec:conclusion}.

\section{Theoretical Foundation}\label{sec:theoreticalFoundation}

\subsection{Concept Drift}

\begin{figure}[htpb]
    \centering
    \subfloat[Original Data at time $t$.]{
    \centering
        \label{subfig:original-data}
        \includegraphics{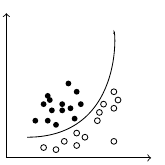}
        }\hspace{5mm}
    \subfloat[Real Concept Drift at time $t+\delta$.]{
        \label{subfig:real-drift}
        \includegraphics{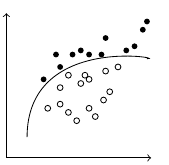}
        }\hspace{5mm}
    \subfloat[Virtual Concept Drift at time $t+\delta$.]{
        \label{subfig:virtual-drift}
        \includegraphics{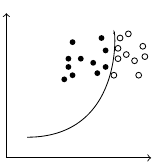}
        }\\
    \caption{Types of Concept Drift -- Probabilistic source. Notice that \textit{real concept drift} changes the best decision boundary, while \textit{virtual concept drift} does not.}
    \label{fig:types_of_drift_prob}
\end{figure}

In this work, we consider that a concept drift can be \textit{real} or \textit{virtual}, shown in Figure \ref{fig:types_of_drift_prob}. Given that $P_t(y|\mathbf{x})$ is the \textit{a posteriori} probability relating a feature vector $\mathbf{x}$ with the target class $y$ at a given time $t$, a \textit{real concept drift} happens when $P_t(y|\mathbf{x}) \neq P_{t+\delta}(y|\mathbf{x})$, for any $\delta > 0$ \citep{gama2014, almeidaEtAl2018, lu2019, bayram2022, hinder2024a}. In a \textit{real concept drift}, the relation between the feature vectors and the target class changes over time, influencing the ideal decision boundary, as shown in Figure \ref{subfig:real-drift}, often bringing the need to retrain models from scratch.

A \textit{virtual concept drift}, on the other hand, does not affect the ideal decision boundaries, as in Figure \ref{subfig:virtual-drift}. It is also described as a stable concept where different regions of the feature space become available over time \citep{kolter07a}, meaning that old data is still relevant. It happens when there is a change in the unconditional or the \textit{a priori} distributions between $t$ and $t+\delta$, while the  \textit{a posteriori} probability remains the same.

A \textit{virtual concept drift} is given by $P_t(\mathbf{x}) \neq P_{t+\delta}(\mathbf{x})$ and/or $P_t(y) \neq P_{t+\delta}(y)$, while $P_t(y|\mathbf{x}) = P_{t+\delta}(y|\mathbf{x})$ \citep{gama2014, lu2019, bayram2022}. Although \textit{virtual concept drift} does not directly change the decision boundary, it is often the case that the models need to be updated, as the decision boundaries learned in the past may not reflect the current distribution, or have a limited view of it.

From an image recognition perspective, the arrival of winter may change the appearance of the space after it snows, but not necessarily invalidate the data learned in spring, summer, or fall, characterizing a \textit{virtual concept drift}. Furthermore, when we started to use masks during the COVID-19 pandemic, face recognition models had to learn the new concept of people wearing masks, but maskless faces are still faces \citep{agate2022}.

In summary, under \textit{real concept drift} we usually need to update decision models to learn the new decision boundary, as part, or the whole information from the past may be conflicting with the current environment. In \textit{virtual concept drift}, old information is still relevant. Thus, the best strategy in this case is to integrate new data into the information already known from the past. In other words, if a classifier knew in advance the ideal decision boundary for the problem at time $t$, there would be no need to do anything at $t+\delta$ under a \textit{virtual concept drift}, while under a \textit{real concept drift} this boundary would be invalidated.

There are still many other concept drift properties in the literature that we do not describe in detail in this work, such as the concept drift severity, speed, and recurrence. For a more in-depth discussion regarding different types of concept drift, refer to \cite{forman2006, minku2010, lu2019, bayram2022, hinder2024b, hinder2024a}.

\subsection{Batch versus Online Processing}
\label{subsec:batch-vs-online}

\begin{figure}[htb]
    \centering

    \subfloat[Stream of batches. Each batch contains $n$ instances.]{
    \label{subfig:stream_batches}
    \includegraphics{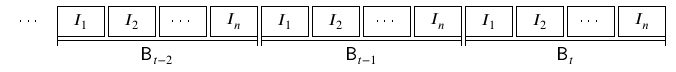}
    }\\
    
    \subfloat[Stream of individual instances.]{
    \label{subfig:stream_instances}
    \includegraphics{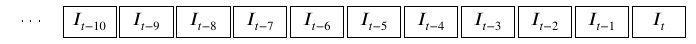}
    }
    \caption{Stream of instances vs stream of batches.}
    \label{fig:streams_instances_batches}
\end{figure}

The concept drift literature often assume that the data used to update the models is made available in batches \citep{almeidaEtAl2018,kozal2021, yang2022} or single instances \citep{Guo2021, liu2021}, shown in Figure \ref{fig:streams_instances_batches}. Batch processing methods (Figure \ref{subfig:stream_batches}) assume that the labeled data used to update the models comes in chunks of a fixed size, containing $n > 1$ instances (e.g., we may receive a new batch of labeled instances every month). In contrast, single instance-based, i.e., online processing methods (Figure \ref{subfig:stream_instances}), update their models with individual instances. In this work, we consider the online processing approach as it can readily process an instance when it is available.

\subsection{Dynamic Selection}

In this section, we describe concepts regarding \ac{DS}. Firstly, let us define some key notations of \ac{DS} techniques that will be used throughout this paper:

\begin{itemize}
    \item $C=\{c_1,\dots c_D\}$: Pool containing $D$ classifiers.
    \item $\mathbf{x}$: an unlabeled test instance.
    \item $I=\{\mathbf{x}, y\}$: a labeled instance containing a feature vector $\mathbf{x}$ and the label $y$.
    \item \ac{DSEL}: The validation set with labeled instances that are used to measure the competence of the classifiers.
    \item $\theta = \{I_1,\dots,I_k\}$: Region of Competence (\ac{RoC}) of the test instance $\mathbf{x}$, such that $\theta \subset DSEL$.
\end{itemize}

\ac{DS} techniques are multiple classifier systems that build the ensemble on the testing phase, depending on the characteristics of the test instance, and usually follow three phases: pool generation, selection and integration \citep{britto2014}. By having a pool of classifiers, the goal is to select the most promising ones in the test instance and integrate them as an ensemble to classify the test instance. Depending on the \ac{DS} method, these steps may change. For example, in \ac{DCS} methods, the single best classifier will classify the test instance. In such a case, the integration part is not needed. 

The goal of the pool generation phase is to generate an accurate and diverse pool of candidate classifiers $C$ \citep{britto2014}. In the case of \ac{DS} for data streams, the online bagging technique \citep{oza} is commonly used, in which all of the classifiers in $C$ have a chance to be trained on every incoming instance. However, by taking into account the need to have a diverse pool of candidate classifiers, there may be better approaches than the online bagging to generate $C$. Considering that the local regions of the feature space become available over time, we use an incremental training approach for training the classifiers in $C$ in our proposed framework.

The selection and integration phases in \ac{DS} can be merged into the classification phase, which usually presents three steps \citep{cruz2018}:

\begin{enumerate}
    \item $\theta$ definition: is the \ac{RoC}, commonly defined through a \ac{kNN} algorithm;
    \item Selection Criteria (Accuracy, Ranking, etc.): dictates how the system will evaluate which classifier(s) will be used to classify the test instance;
    \item Selection Mechanism (\acf{DCS} or \acf{DES}): determines whether only the single best classifier will label the test instance or if an ensemble of the classifiers that meet the selection criteria will be built.
\end{enumerate}

The $\theta$ is where the competence of candidate classifiers is measured according to the selection criteria. It is commonly defined by using a \ac{kNN} algorithm and is denoted by $\theta = \{I_1,\dots, I_k \}$, where $k$ is the number of neighbors, and $I_k$ is a labeled instance. To do so, we need a set of labeled instances named \ac{DSEL}, which can be either the training or a validation set. Besides \ac{kNN}, examples of other approaches explored in literature for \ac{DS} are fuzzy hyperbox \citep{DAVTALAB2024102036}, Graph Neural Networks \citep{SOUZA2024102145}, and meta-features \citep{CRUZ201784}.

\ac{DS} techniques that run under data streams often use a sliding window on the incoming labeled instances as the \ac{DSEL}. In this work, when referring to such validation set for \ac{DS}, we will use the term \ac{DSEW} to differ from the \ac{DSEL} used in static \ac{DS}.

After the set $\theta$ is defined, the competence of the candidate classifiers is measured based on the selection criteria. \ac{DCS} methods select the single best classifier, while \ac{DES} may select one or more classifiers to compose an ensemble. For instance, a \ac{DES} method could select all classifiers that correctly classify all instances in $\theta$. Then, the chosen classifiers are integrated as an ensemble to classify $\mathbf{x}$.

\section{Proposed Method -- Incremental Adaptive Dynamic Ensemble Selection (IncA-DES)}\label{sec:proposal}

\begin{figure}[htbp]
    \centering
    \includegraphics{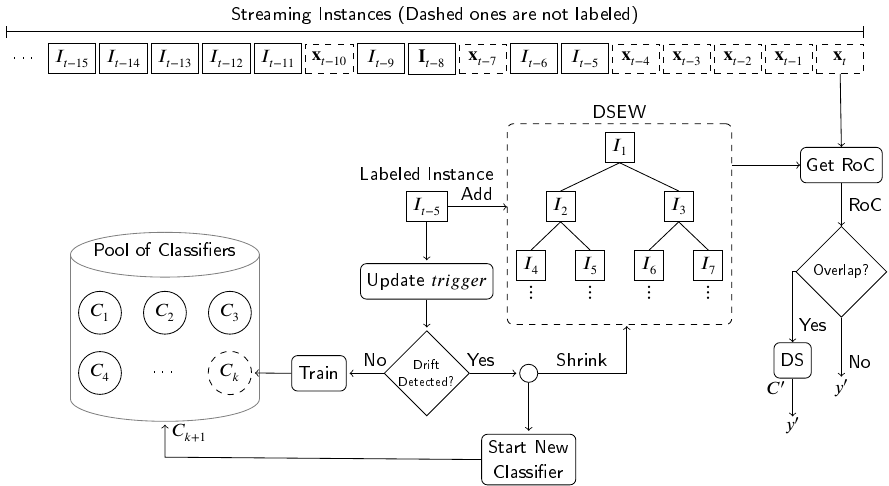}
    \caption{IncA-DES's scheme of training and classification. The instance $I_{t-5}$ is used for updating the $trigger$, and if a concept drift is detected, a new classifier starts to be trained, and the DSEW is shrunken. Otherwise, the last classifier $C_k$ is trained with $I_{t-5}$. For classifying data point $\mathbf{x_{t}}$, the \ac{RoC} is computed from the \ac{DSEW}. If there is a disagreement between most neighbors, the \ac{DS} method is used. Otherwise, the majority class in the \ac{RoC} is Predicted. We represent the \ac{DSEW} as a K-d Tree, but it could be any data structure depending on the neighborhood search method.}
    \label{fig:incades-scheme}
\end{figure}

In this work, we propose \ac{IncA-DES}, a \ac{DS}-based framework for data streams that address concept drift. It can leverage information of stable concepts and adapt to concept drift when needed. Figure \ref{fig:incades-scheme} shows an overview of the proposed framework when a newly labeled training sample is available and when a sample comes for being classified. Notice that the $trigger$ component is responsible for adapting to concept drift. 

Whenever a training instance arrives, a prediction is made by the system and sent to the $trigger$. For instance, drift detectors that track the error-rate can be used, such as \ac{DDM} \citep{gama2004} and \ac{RDDM} \citep{BARROS2017}. If a concept drift is detected, the \ac{DSEW} is shrunk. Thus, we avoid unnecessary adaptation that may happen in blind adaptation strategies while still being aware of concept drift. In addition, incoming training instances are added to the \ac{DSEW}, ensuring it remains up-to-date, and are used for training $C_k$, the last classifier added to $C$. The rationale is that, taking into account the natural progression of data arriving over time, the classifiers are trained considering different perspectives, in different moments, favoring the generation of local experts. Additionally, old classifiers trained in past knowledge are kept as ensemble candidates, promoting both time-dependent learning and robustness.

As a counterpart of the training approach chosen for \ac{IncA-DES}, by training one single classifier one by one, diversity may take longer to develop if compared to resampling strategies \citep{oza, gomes2017, abadifard2023}. However, such an approach may be better if we have limited access to labeled data. The reason is that, when using a resampling training strategy, candidate classifiers may not receive enough instances for training in such cases, increasing the chances of having underfitted classifiers.

When classifying an instance, \ac{IncA-DES} checks for overlap in the \ac{RoC} $\theta$, which is computed through a neighborhood search algorithm. If the rate of the majority class in $\theta$ is above a threshold $\omega$, the class representing the neighborhood is predicted. Otherwise, the \ac{DS} method is utilized. Such classification strategy helps reduce the processing time of \ac{IncA-DES}, as \ac{DS} methods are computationally costly. Still thinking of processing time, brute force \ac{kNN} brings limitations on how big a validation set can be, as due to the continuous data stream, searching for neighbors can become computationally expensive \citep{assis2023}. Since data streams are concerned with processing incoming instances quickly, we propose an Online K-d tree algorithm to reduce the processing time of the \ac{RoC} definition, thereby opening space for having a larger \ac{DSEW}, beneficial in cases of stable concepts and \textit{virtual concept drift}.

\subsection{IncA-DES Training}

The pseudocode of the training procedure is in Algorithm \ref{alg:incades-training}. We divide the training of \ac{IncA-DES} into three steps:

\begin{enumerate}
    \item $Trigger$ update (Line 1);
    \item Adaptive \ac{DSEW} management (Lines 2-7);
    \item Time-dependent Pool Management (Lines 8-17).
\end{enumerate}

In the following sections, we explain these specific topics and link them to the pseudocode.

\begin{algorithm}[htpb]
\caption{IncA-DES Training ($Instance$)}
\label{alg:incades-training}

\SetKwInOut{Input}{input}\SetKwInOut{Output}{output}

\Input{Labeled Instance $I$\\}
Update trigger with $I$ \Comment*{First update the drift detector} 

$DSEW \gets DSEW \cup I$ \Comment*{Add $I$ to the DSEW}
\If{$|DSEW| > W$} {
$removeOldestInstance(DSEW)$ \Comment*{Drop the last instance if needed}
}

\If{concept drift was detected}{
\Comment{Shrink the DSEW and start a new classifier if a concept drift was detected}
shrink $DSEW$\\
$C_k \gets $ a new classifier \\
$prune(C, DSEL, C_{k-1}, D)$ \Comment*{Prune a classifier based on the pruning method chosen}
$C \gets C \cup C_k$ 
}
\If{$C_k$ was already trained on $F$ instances}{
\Comment{Start a new classifier if the last one was already trained on $F$ instances}
$C_k \gets $ a new classifier \\
$prune(C, DSEW, C_{k-1}, D)$\\
$C \gets C \cup C_k$\\
}
$C_k \gets  latestClassifierAvailable(C)$ \Comment*{Get $C_k$ from $C$}
$train(C_k, I)$ \Comment*{Train $C_k$ on the training instance}
\end{algorithm}

\subsubsection{Trigger Update and Adaptive DSEW Management}

When a labeled instance $I$ arrives for training, the $trigger$ component is updated, as in Line 1 of the pseudocode. This is done by getting a prediction from the system and comparing the result to the true label of $I$. The result is sent to the $trigger$. After that, the \ac{DSEW} management is performed. The new instance is added to the \ac{DSEW} (Line 2), and the oldest one is removed if it has surpassed its maximum size (Line 4). If a concept drift is detected, the \ac{DSEW} is shrunk, leaving only the instances that arrived after the warning level that was given by the $trigger$ (Line 7).

One may want to limit the size of the \ac{DSEW} because of memory and processing time. Since the \ac{RoC} definition is done through a \ac{kNN} algorithm, its computational cost can increase with the growth of the search space. If there is a stable concept drift, it is preferable that as many instances as possible compose the \ac{DSEW}, limited solely by the available memory. However, the processing time of the \ac{kNN} algorithm becomes a concern. To mitigate this, we propose an Online K-d tree approximate neighborhood search algorithm, which is detailed in Section \ref{sec:kdtree}.

\subsubsection{Time-dependent Pool Management}

Lastly, the update of the pool $C$ is performed by training the classifier $C_k$ with the training instance $I$. $C_k$ is always the last classifier added to the pool, which can receive a maximum of $F$ instances for training. Once past that limit, the training of that classifier is interrupted, and the next time an instance arrives for training, a new classifier is added to $C$ (Lines 12-16), which will be trained on the incoming instances. Moreover, the pool of \ac{IncA-DES} has a limit $D$ of how many classifiers it can have in order to prevent the infinite generation of classifiers that would lead to high memory consumption. When adding a new classifier, if $C$ is full, the oldest one is pruned (Lines 10 and 15). We call this an age-based pruning engine, which, in contrast to dropping classifiers considering their performances, requires less processing time and does not have much difference in accuracy, as shown by \cite{almeidaEtAl2018}.

Furthermore, the training of $C_k$ is interrupted not only when it has received $F$ instances for training, but when a concept drift is detected as well (Lines 9-11). The rationale is that we guarantee that $C_k$ is continually trained on the newest concept. The classifier's training is done after deciding whether a new classifier will be added through a concept drift or if the limit $F$ of training was surpassed (Line 17). The $F$ parameter brings a trade-off regarding diversity and representativeness, i.e., the size of the local region in which a classifier is trained. On the one hand, if a classifier has few instances for training, it may not have enough information to learn properly. Still, it can lead to quicker adaptation and generation of diversity. On the other hand, a bigger value for $F$ may postpone the generation of diversity in $C$ since we train the classifiers incrementally. 

However, the optimal value for $F$ may change according to the dataset since a more complex decision boundary may require one classifier to have more training information. At the same time, smaller groups in the feature space may get better results with a more diverse pool of candidate classifiers.

We argue that our incremental training strategy favors the generation of local experts, considering that different regions of the feature space become available in different timestamps, and we limit the region where classifiers are trained. This assumption comes from the fact that the pattern of events or people's habits is expected to change over time. As a counterpart, the diversity may take longer to be created, as one classifier is added to $C$ at a time. 

As \ac{IncA-DES} trains its classifiers incrementally, it needs an incremental learner, such as the \ac{HT} classifier \citep{domingos2000}, which is widely used in the concept drift literature. 

\subsection{Classification}\label{subsec:classification}

To classify an unlabeled test instance $\mathbf{x}$, the \ac{IncA-DES} first computes the neighborhood of the test instance $\mathbf{x}$ in the \ac{DSEW} to define its $\theta \subseteq DSEW$. The $\theta$ is defined as the $k$-nearest neighbors of $\mathbf{x}$ in the \ac{DSEW}, where the distance between $\mathbf{x}$ and the instances in the \ac{DSEW} can be computed using any distance function, such as the Euclidean or Canberra distances. Then, the class overlap of $\theta$ is measured. If the rate of the majority class in $\theta$ is greater than $\omega$, the system predicts that class directly. Formally, let $s_i\in \theta$ be a subset associated with a class label $y_i \in c$, such that $c$ is the set of possible class labels. Denote $y_{maj}$ the majority class in $\theta$. The decision rule is as follows:

\begin{equation}
    \frac{|\{s_i\in \theta\  | \ y_i = y_{maj}\}|}{|\theta|} \geq \omega \implies return\ \  y_{maj}
\end{equation}

If the above condition is not met, the competence of the classifiers in $C$ is measured in the \ac{RoC}, and the ensemble $C' \subseteq C$ is built based on a \ac{DS} method, which can be any \ac{DS} method based on \ac{RoC} \citep{cruz2018}. Then, $C'$ classifies the instance $\mathbf{x}$ through majority voting, as presented in Figure \ref{fig:incades-scheme}.

The most straightforward benefit of the overlap-based classification is the processing time, as the \ac{DS} method is not used if most neighbors, as dictated by the $\omega$ parameter, agree with the class label. The \ac{kNN} usually struggles in local regions with high class overlap, and limiting its action to regions with low or no overlap can lead to best performance. In regions with class overlap, \ac{DS} usually have the best results over \ac{kNN} \citep{cruz2017}.

\subsection{Online K-d Tree}
\label{sec:kdtree}

When using a brute force \ac{kNN} search to define the \ac{RoC}, as the search space (i.e., number of instances) $n$ grows, the processing time grows in $O(Kn)$, where $K$ is the dimensionality of each sample (i.e., number of features) and $n$ the search space. The K-d tree algorithm \citep{bentley1975} can be used to decrease the search space, which works by building a K-dimensional binary tree data structure through recursively partitioning the feature space, and then can be used for neighborhood search. 

However, K-d trees were designed assuming we initially have access to the whole training set. As in data streams we have a potentially infinite flow of data, a K-d tree is required to be updated at every moment, with many insert and delete operations, which will make the data structure unbalanced. Other limitations include the need to normalize data, as the difference in the magnitude of the features may vary and harm the decrease of the search space \citep{Wu2008}. However, data normalization is not a trivial task in data streams since the range of the features can change a lot over time, and many times using the original values of features is preferred in data streams over applying a min-max normalization, for example \citep{barboza2023}. To address these challenges, we propose an Online K-d tree that tackles the concerns mentioned in data streams.

In order to deal with the feature magnitude issue and the need of normalization, we have used the Canberra Distance function (Equation \ref{eq:canberra-distance}) \citep{perlibakas2004} to perform the neighborhood search. The reasons why we chose it include: 1) it normalizes the distances with the sum of the absolute values of the features; 2) it has shown to have better results than the Euclidean Distance in data streams \citep{barboza2023}.

\begin{equation}\label{eq:canberra-distance}
    d(\mathbf{x}_1,\mathbf{x}_2) = \sum_{i=1}^{K}\frac{|\mathbf{x}_1[i]-\mathbf{x}_2[i]|}{|\mathbf{x}_1[i]|+|\mathbf{x}_2[i]|}
\end{equation}

\begin{figure}
    \centering
    \subfloat[An example of a balanced K-d tree.]{
    \centering
        \label{subfig:kdtree-balanced}
        \includegraphics{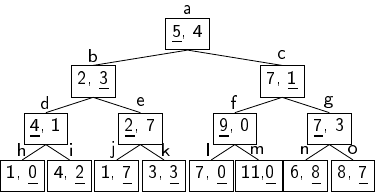}
        }
    \subfloat[Split of the nodes in a two-dimensional feature space. The blue vertical line refers to the split of the root node. The green data points are the 5 nearest neighbors defined by the K-d tree with the Canberra Distance.]{
        \label{subfig:kdtree-plot}
        \includegraphics[width=0.3\linewidth]{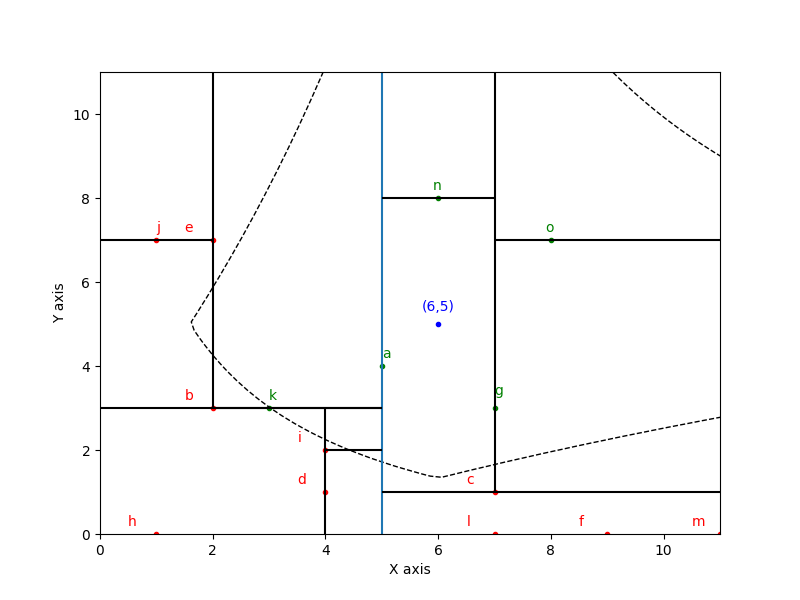}
        }
    \caption{Example of a balanced K-d tree.}
    \label{fig:balanced-kdtree}
\end{figure}

Regarding the balancing concern in K-d trees, the initial splits of the nodes in the data structure may not be efficient when more instances arrive. Consider that one initial training data gives us the K-d tree shown in Figure \ref{subfig:kdtree-balanced}, and its divisions in two-dimensional feature space are like in Figure \ref{subfig:kdtree-plot}. The green points are the 5-nearest neighbors of the data point $(6,5)$ defined by the K-d tree with the Canberra distance, and the dashed contour is the domain of the Canberra Distance to the farthest neighbor.

\begin{figure}
    \centering
    \subfloat[An example of an unbalanced K-d tree.]{
    \centering
        \label{subfig:kdtree-unbalanced}
        \includegraphics{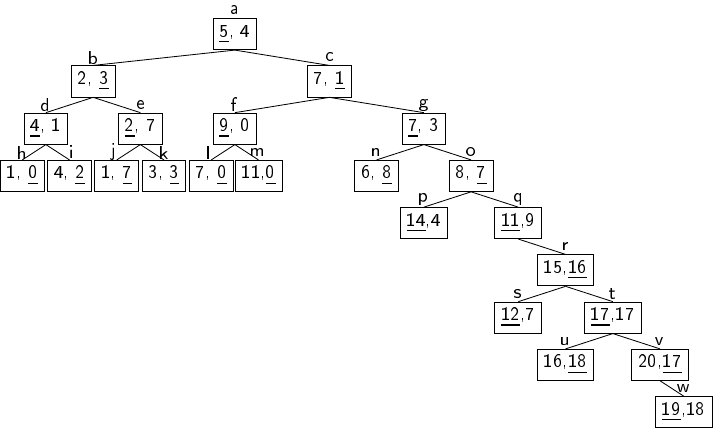}
        }\hspace{5mm}
    \subfloat[Split of the nodes of an unbalanced K-d tree in a two-dimensional feature space. The green data points are the 5 nearest neighbors defined by the K-d tree with the Canberra Distance.]{
        \label{subfig:kdtree-plot-unbalanced}
        \includegraphics[width=0.5\linewidth]{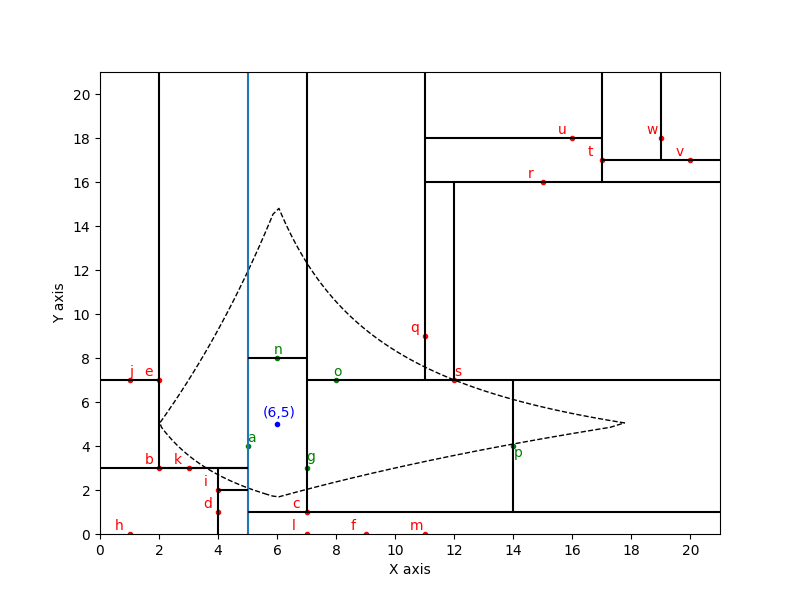}
        }\hspace{5mm}
    \caption{Example of an unbalanced K-d tree.}
    \label{fig:unbalanced-kdtree}
\end{figure}

After the arrival of more instances in a data stream, the data distribution may change, e.g., the range that features arrive is different than the training data we had initially, as exemplified in Figure \ref{subfig:kdtree-unbalanced}. This could happen, for instance, if we have built the K-d tree with data from the winter. Later, when the spring or summer arrives, the range of the features (e.g., the temperature) will be different.

See that after the tree becomes unbalanced, there is not an efficient split of the nodes anymore, as shown in Figure \ref{subfig:kdtree-plot-unbalanced}. If we use this structure to perform neighborhood searches, the decrease in search space may not be as efficient, depending on the depth of the structure and the data distribution. In this example, the point $s$ is a nearest neighbor than $p$ to the point $(6,5)$ by the Canberra Distance, but the K-d tree algorithm pruned it from the search space.

For that reason, a common strategy found in the literature is to rebuild the K-d tree when it is two times bigger than when it was built \citep{muja2009, jo2017}. This strategy, which we also adopt, ensures that the tree remains balanced even when the data distribution changes with the arrival of new instances and helps keep the neighborhood search efficient.

Next, let us define the structure of a K-d tree node. Be $t$ a K-dimensional tree node such that $t=(I, s)$, where $I=(\mathbf{x},y)$ is a data point, $\mathbf{x}$ is a K-dimensional feature vector, $y$ the instance label and $s$ is the split dimension of the node, such that $s\leq K$. The split dimension $s$ says what is the current node's key, in which $key=\mathbf{x}[s]$. Each node $t$ can have a left and right subtree, denoted by $lt$ and $rt$, respectively. In order to maintain a K-d tree consistent, and the search of a node can be done in $O(\log{n})$, as in a binary search tree, the following properties must be fulfilled:

\begin{itemize}
    \item $\forall$ Instance $I \in lt, I[s] < I_t[s]$
    \item $\forall$ Instance $I \in rt, I[s] \geq I_t[s]$,
\end{itemize}
\noindent where $I_t$ is the instance that belongs to the K-d tree node $t$.

\subsubsection{Building a KDTree}

A balanced K-d tree brings a more efficient neighborhood search by enabling better search space pruning. If a subtree holds most instances in the dataset, excluding it from the search space may speed up neighborhood search. However, discarding too many instances in an unbalanced K-d tree can reduce the effectiveness of the neighborhood search since helpful neighbors are more likely to be pruned. The pseudocode for building a balanced K-d tree is in Algorithm \ref{alg:build-kdtree}.

Firstly, we receive $N$ instances in which the median of the split dimension ($s=0$ in the root node) is gathered in Line 5 of the algorithm. The first instance that carries the median value will be the root node (Line 9); the instances with a smaller value than the median in the dimension $s$ will be inserted to the left side of the root node (Line 11), and instances with higher or equal values will be to the right of the root (Line 13). This process is repeated with every set of instances recursively (Lines 15 and 16). 

\begin{algorithm}[!htb]

\SetKwInOut{Input}{input}\SetKwInOut{Output}{output}

\Input{Set of Instances $instances \gets$ $\{I_1, I_2,...,I_N\}$,\\
$splitDimension$\\}
\Output{K-d tree Node}


\If{sizeOf($instances$) == 0} {
    return null
}

$instancesToTheLeft \gets \emptyset$\\
$instancesToTheRight \gets \emptyset$

$median \gets$ getMedian($instances$, $splitDimension$) \Comment*{Get the median of the current split dimension}

\ForEach{$I \in$ $instances$} {
\Comment{Values lower than the median go to the left subtree, and values greater or equal to the right}
    \If{$I[splitDimension]$ == $median$} {
        $medianInstance \gets I$
    }\ElseIf{$I[splitDimension] < median$} {
        $instancesToTheLeft \gets instancesToTheLeft \cup I$\\
    }\Else{
        $instancesToTheRight \gets instancesToTheRigth \cup I$\\
    }
}

$node \gets KDTreeNode(medianInstance, splitDimension)$

$node.left \gets $ Build K-d Tree($instancesToTheLeft$, $(splitDimesion+1)\mod{K}$)\\
$node.right \gets $ Build K-d Tree($instancesToTheRight$, $(splitDimension+1)\mod{K}$)\\

return $node$

\caption{Build K-d Tree(Set of Instances, Split Dimension)}
\label{alg:build-kdtree}
\end{algorithm}

\subsubsection{Insertion}

The insertion algorithm in our Online K-d tree is the same as a regular one, as described by \cite{drozdek2016}, and its pseudocode is shown in Algorithm \ref{alg:insertion-alg}. Starting with the root node, the algorithm iteratively traverses the tree by comparing the instances' values in the nodes' split dimension until it finds a leaf node (Lines 4-11). If the instance to be added has a smaller value than the current node on the split dimension, the search continues on the subtree to the left, or to the right otherwise. When the leaf node is found, if the tree is empty, it is added to the root node (Line 13). Otherwise, the value on the split dimension of its parent node is compared (Line 14). If the instance to be added is smaller than the split value of its parent node, it will be added to the left (Line 15). It is added to the right if its value is greater or equal to its parent node (Line 17).

\begin{algorithm}

\SetKwInOut{Input}{input}\SetKwInOut{Output}{output}

\Input{Instance $I$ \\}

$p \gets KDTree.root$ \\
$prev \gets \emptyset$ \\

$depth \gets 0$

\While {$p$ is not a leaf node} {
    \Comment{Iterate the tree until it finds a leaf node}
    $prev \gets p$\\
    \If{$I[depth] < p.instance[depth]$} {
        $p \gets p.left$
    }\Else {
        $p \gets p.right$
    }
    $depth \gets (depth+1)\mod{K}$
}

\If{$KDTree.root == null$} {
    $KDTree.root \gets Node(I, depth)$ \Comment*{Assign to the root if the tree is empty}
}\ElseIf{$I[(depth-1)\mod{K}] < p.instance[(depth-1)\mod{K}]$} 
{
    $prev.left \gets Node(I, depth)$ \Comment*{Assign to the left child if the value is smaller than the leaf node}
}\Else {
    $prev.right \gets Node(I, depth)$ \Comment*{Assign to the right child otherwise}
}

\caption{K-d Tree insertion algorithm ($Instance$)}
\label{alg:insertion-alg}
\end{algorithm}

\subsubsection{Deletion}

When a node is deleted, the K-d tree may become inconsistent, and the search for a node based on the key values of the split dimensions, as in a binary search tree, can not be done anymore. A deletion algorithm in the K-d tree, in order to maintain its consistency, requires a lot of comparisons and arrangements of the nodes \citep{drozdek2016} or to a subtree to be entirely rebuilt. In data streams, the large flow of data brings the need to process incoming instances quickly, and a fast deletion is mandatory. 

\begin{figure}
    \centering
    \subfloat[An example of a 2-dimensional K-d tree. The node (7,1) is to be deleted.]{
    \centering
        \label{subfig:kdtree-ex1}
        \includegraphics{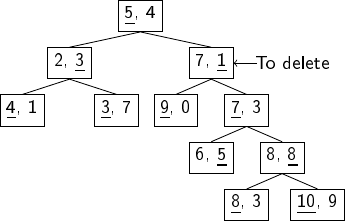}
        }\hspace{5mm}
    \subfloat[The nodes (6,5) and (8,8) have the same split dimension as the node to be deleted. Thus, they are candidates for replacement.]{
        \label{subfig:kdtree-ex2}
        \includegraphics{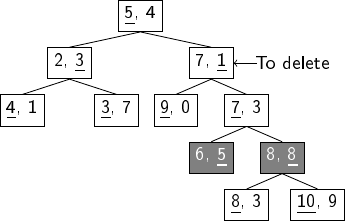}
        }\hspace{5mm}
    \subfloat[As the Node (6,5) had the smallest value in the split dimension, it replaces (7,1).]{
    \centering
        \label{subfig:kdtree-ex3}
        \includegraphics{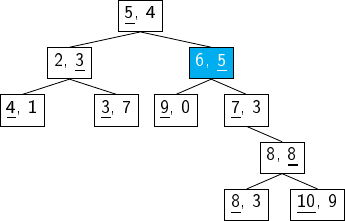}
        }\hspace{5mm}
    \subfloat[The nodes (7,3) and (8,3) now make the K-d tree inconsistent since their values in the split dimension are lower than (6,5). They should be on the left subtree of (6,5), not on the right.]{
        \label{subfig:kdtree-ex4}
        \includegraphics{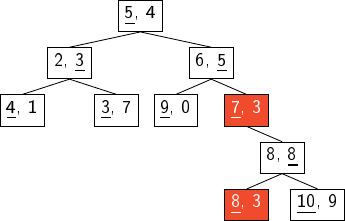}
        }\hspace{5mm}
    \caption{How inconsistencies can be created in a K-d tree with the deletion of a node.}
    \label{fig:deletion-kd-tree}
\end{figure}

In Figure \ref{fig:deletion-kd-tree}, we present a step-by-step of the deletion algorithm presented by \cite{drozdek2016}. In Figure \ref{subfig:kdtree-ex1}, there is a consistent K-d tree, and we want to delete the node (7,1). To do so, one must choose a node to replace it on the right subtree. The candidates must be split on the same dimension as (7,1), i.e., the second dimension. The candidates are the nodes (6,5) and (8,8), highlighted in Figure \ref{subfig:kdtree-ex2}. Since (6,5) has a lower value than (8,8) in the split dimension, it is chosen to replace the deleted node. The new K-d tree after the replacement is in Figure \ref{subfig:kdtree-ex3}. However, after replacing the deleted node with (6,5), one property of the subtree now is not satisfied: $\forall$ Instance $I \in rt, I[s] \geq I_t[s]$. In Figure \ref{subfig:kdtree-ex4}, we can see that the nodes (7,3) and (8,3) are in the right subtree of (6,5), but their values in the split dimension $s$ are smaller than 5, contradicting the above property. In this data structure, the search for a node would not be done in $O(\log{n})$. It would need either a depth-first or a breadth-first search.

One solution would be to replace it with the node with the smallest value in the right subtree in the split dimension. However, inconsistencies could still be generated in different dimensions. Further, a more complex deletion should be performed when the node replacing the deleted one is not a leaf.

\begin{figure}
    \centering
    \includegraphics{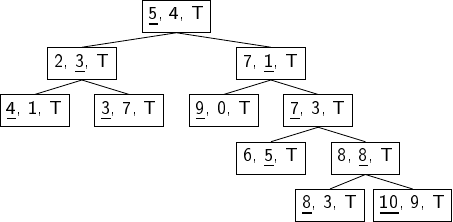}
    \caption{2-dimensional Tree with Flagged Nodes.}
    \label{fig:k-dtree-flagged}
\end{figure}

\begin{figure}
    \centering
    \includegraphics{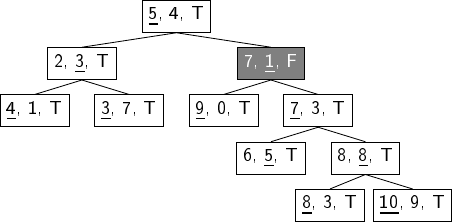}
    \caption{2-dimensional Tree with Flagged Nodes: A Node was Deleted.}
    \label{fig:k-dtree-false}
\end{figure}

Therefore, deleting a node in a K-d tree is not a simple task. In the Online K-d tree proposed in this work, we employ a lazy deletion, which "deactivates" the deleted nodes without explicitly removing them from the data structure. The structure of the K-d tree is maintained, and the distance is not calculated on the deleted nodes. This strategy was also used by \cite{cai2021}. To do so, a flag is added to the nodes in the data structure, which tells whether the node is active, as in Figure \ref{fig:k-dtree-flagged}, where T stands for True, i.e., the node is active. If the node (7,1) is to be deleted, its flag will be set to False, and the resulting tree will be like in Figure \ref{fig:k-dtree-false}. Thus, we can deactivate nodes while keeping the K-d tree consistent. In this approach, the time complexity of the deletion remains solely on the search algorithm.

As a counterpart, this approach may lead to excessive memory usage since nodes that are not being used anymore are still maintained solely to keep a consistent structure. As we remove many instances, searching for one may become costly as the number of inactive nodes increase. To mitigate this issue, we set a limit of $\beta$ for the proportion of inactive nodes. Once this limit is exceeded, the tree is rebuilt.

\subsubsection{Neighborhood Search}

Once we have a K-d tree data structure, we can search for the $k$ nearest neighbors. Since we use the Canberra Distance for the neighborhood search, a question arises: What criterion should be used to prune subtrees? The approximate neighborhood search K-d tree usually relies on the squared difference of the split dimension of the current node to do so, or to the absolute difference between features, but it uses the Euclidean Distance. With another distance function, the same split criterion may not be adequate for pruning subtrees from the neighborhood search. Instead, we use a Canberra Distance segment on the current node's split dimension, as shown in Equation \ref{eq:seg-canberra}.

\begin{equation}\label{eq:seg-canberra}
    seg=\frac{|I[s]-x[s]|}{|I[s]| + |x[s]|}
\end{equation}

Thus, if the calculated $seg$ is smaller than the current maximum distance (i.e., the largest distance among the $k$ nearest neighbors), that subtree will be searched for neighbors. Otherwise, it is pruned from the search space. For the cases where both $I[s]$ and $\mathbf{x}[s]$ are equal to zero, which would lead to an undefined value due to a division by zero, $seg$ will be equal to zero instead.

Algorithm \ref{alg:getNeighbors} exposes the pseudo-code of the neighborhood search of the K-d tree. In our version of the K-d tree, we start to calculate the distances from the root node, as done by \cite{chen2019}. Line 3 checks whether the node is active, and its distance is calculated to the target test instance in Line 4. If there are fewer instances than $k$ in the distances and neighbors list, it is added to both distances and neighborhood lists in Lines 7 and 8. If there are already $k$ neighbors calculated, we get the maximum current distance in Line 11 and compare the calculated distance from the current node to the target instance in Line 12. If it is smaller than the current maximum neighbor, it replaces it on both distances list and neighborhood in Lines 13 and 14.

Regardless of whether a node is active, the best subtree to continue the neighborhood search is chosen by comparing the target instance's value to the current node's split dimension. If both left and right subtrees exist for the current node, the decision of which is the best one is made in Line 17. If the test instance has a split value greater or equal to the current node, the best subtree is the right one, as in Line 18. Otherwise, the best is the left subtree (Line 21). If only one subtree exists, be the right or the left, they are set as the best (Lines 23-26). If we have found a leaf node, the algorithm returns the neighbors found so far (Line 28). 

Then the decision of whether to search or not the subtree is done in Line 30, which is where the segment of the canberra distance in the split dimension $s$ is calculated by using Equation \ref{eq:seg-canberra}, and is compared to the current maximum distance. If the algorithm says that the best subtree should be search for neighbors, the search continues recursively in Line 31. This is done on the subtree other than the best one also (Lines 33-35). In the end, the $k$ neighbors are returned in Line 36.

\begin{algorithm}

\SetKwInOut{Input}{input}\SetKwInOut{Output}{output}
\SetKwProg{Fn}{Function}{}{end}

\Input{KDTreeNode $Node$, 
Target Instance $\mathbf{x}$, 
List $distances$, List $currentNeighbors$, Number of Neighbors $k$}
\Output{Neighbors, Distances}

$I \gets Node.instance$\\
$s \gets Node.splitDimension$\\

\If{$Node\  is\  Active $ } 
{ 
    $distanceToNode \gets distanceFunction(I, \mathbf{x})$ \Comment{Calculate distance if the node is active}
    \If {Size of $distances$ is smaller than $k$ neighbors} {
        \Comment{Add distance and instance to the lists if its size is smaller than $k$}
        $distances.add(distanceToNode)$\\
        $neighbors.add(Node.instance)$
    }\Else {
        \Comment{Otherwise, tests whether the current maximum distance is greater than the distance to the current node}
        $maximum, maxIndex \gets getMaximum(distances)$\\
        \If{$distanceToNode \leq maximum$} {
            $distances[maxIndex] \gets distanceToNode$\\
            $neighbors[maxIndex] \gets Node.instance$
        }
    }
}

\If{$Node.left \neq null$ and $Node.right \neq null$}{
    \Comment{Checks what is the best subtree to continue the search}
    \If{$\mathbf{x}[s] \geq I[s]$}{
        $best = Node.right$\\
        $other = Node.left$
    } \Else{
        $best = Node.left$\\
        $other = Node.right$
    }
} \ElseIf{$Node.left \neq null$} {
    $best = Node.left$
} \ElseIf{$Node.right \neq null$} {
    $best = Node.right$
} \Else{
    return $neighbors$
}

$maximum \gets getMaxDist(distances)$\\

\If{$Should \ Search \ SubTree(Node, \mathbf{x}, s, maximum)$}{
    $neighbors, distances \gets Neighborhood Search(best, target, distances, neighbors, $k$)$ \Comment{Continue neighborhood search on the best subtree}
}

$maximum \gets getMaxDist(distances)$\\

\If{$other \neq \emptyset$} {
    \If{$Should \ Search \ SubTree(Node,\mathbf{x}, s, maximum)$} {
        $neighbors, distances \gets Neighborhood Search(other, target, distances, neighbors, $k$)$ \Comment{Search the other subtree too if needed}
    }
}

return $neighbors$

\caption{Neighborhood Search ($Node, \mathbf{x}, distances, neighbors, k$)}
\label{alg:getNeighbors}
\end{algorithm}

\section{Related Works}\label{sec:stateoftheart}

We may find various approaches for dealing with concept drift in the literature, such as blind adaptation \citep{widmer1994, Pfahringer2007, almeidaEtAl2018, Guo2021, yang2022}, use of drift detectors \citep{BARROS2017, yu2022, Gulcan2023, Cerqueira2023}, ensembles \citep{Guo2021, liu2021, yang2022}, and \acf{DS} \citep{almeidaEtAl2018, abadifard2023}.

We focus on describing methods based on Triggers (Section \ref{sec:triggers}) and ensemble (Section \ref{sec:ensemble}) since the proposed \ac{IncA-DES} combines these two approaches. For an analysis that includes other strategies to deal with concept drifts, refer to \cite{lu2019, Han2022}.

\subsection{Concept Drift Detection}\label{sec:triggers}

Concept drift detectors, in the majority, monitor the classification error. An increase in the classification error may be evidence of concept drift, and different detectors employ different tests in order to trigger a concept drift. The \ac{PH} continuously performs statistical tests on the error rate. When it rises over a predefined threshold, it is told that a change has happened \citep{page1954}. The \ac{DDM} \citep{gama2004}, one of the most well-known drift detectors, also monitors the error rate of classification and considers different thresholds for warning and drift levels. The warning level is where the concept drift started, which is given when the error rate rises above a certain level. If the error rate keeps growing and gets to the drift level, a concept drift has happened. The \ac{EDDM} \citep{garcia2006} extends \ac{DDM} by taking into account the distance between two misclassifications through the usage of distance functions. \ac{RDDM} \citep{BARROS2017} also extends the \ac{DDM} by triggering a concept drift when the warning level stays for too long.

\ac{STEPD} \citep{nishida2007} comes from the principle that accuracy remains similar as new instances come if the concept is stable, and a decrease in accuracy may indicate a concept drift. The difference in accuracy is checked through a statistical test, in which when the p-value is smaller than a threshold, a concept drift is detected.

\ac{HDDM} \citep{frias-blanco2015} applies either the Hoeffding's Inequality \citep{hoeffding1963} (\ac{HDDM}A) or the McDiarmid's Inequality \citep{mcdiarmid1989} (\ac{HDDM}W) for detecting changes in the moving average of streaming data. \ac{ACDDM} \citep{YAN2020} also uses Hoeffding's Inequality, monitoring the prequential error. \ac{ECDD} tracks the classification error by using Exponentially Weighted Moving Average \citep{Ross_2012}. \ac{ADWIN} maintains a sliding window $W$ and continuously compares statistics of two different data portions inside $W$. When the statistics of the subwindows differ from each other, a concept drift is triggered \citep{Bifet2007}.

\ac{FHDDM} \citep{Pesaranghader2016} uses the assumption that, in stable concepts, the accuracy may either remain unchanged or increase. A concept drift is triggered if the accuracy decreases above a certain level. Hoeffding's Inequality compares the maximum accuracy gathered to the current one. \ac{FHDDMS} \citep{Pesaranghader2018} maintains a short and a long window by considering that a short window is best for abrupt concept drift and the larger one for gradual concept drift. \ac{FHDDMS$_{add}$}, from the same work, replaces the binary calculation by the summation of the recent points.

SeqDrift1 performs hypothesis tests on two different batches of the data \citep{sakthithasan2013}. If the hypothesis is rejected, the batches are combined, and the hypothesis test is repeated when a new batch is available. Authors have used the Bernstein bound \citep{peel2010} on the hypothesis test. SeqDrift2 extends SeqDrift1 by applying reservoir sampling, which the authors argue to result in a tighter threshold for drift detection \citep{Pears2014}.

\ac{MDDM} uses McDiarmid's Inequality on statistical tests applied on a sliding window in which earlier instances have a higher weight than the oldest. There are three different versions of the \ac{MDDM}, where each one uses a different weight function \ac{MDDM}A uses the arithmetic weighting, \ac{MDDM}G the geometric weighting function, and \ac{MDDM}E an exponential function.

Recent works have also been focusing on unsupervised drift detection. \ac{UDD} uses the uncertainty of predictions of neural networks as the error rate \citep{baier2021}. \ac{LD3} considers the changing correlation of the predicted labels overtime \citep{Gulcan2023}. \ac{STUDD} uses two classification models: a student and a teacher \citep{Cerqueira2023}. The teacher is trained on the true labels of the dataset, while the student is trained using the teacher's predictions on the same dataset. The concept drift is tracked by taking into account the difference between the predictions of both models, and when this difference between the two models rises above a threshold, a concept drift is triggered. Thus, there is no need for the true label of the data, as the error rate is calculated through the difference of previously trained models (some access to labels is still needed initially).

\subsection{Ensemble Learning for Concept Drift}
\label{sec:ensemble}

Ensemble learning algorithms can be divided into two categories: static and dynamic. Static ensembles are those that the ensemble of classifiers is defined during the training phase. All of the classifiers are trained by following a training strategy, such as bagging \citep{Breiman1996}. Dynamic ensembles, differently from static, define the ensemble during the test phase. This is done by gathering the most promising classifiers according to the characteristics of the test instance. In this section, we explore methods of both types found in the literature.

\subsubsection{Static Ensemble}

Many of the ensemble methods for data streams (static or dynamic) use online bagging, or OzaBag \citep{oza} to train the classifiers. Online bagging simulates the classic bagging algorithm used in static \ac{ML} by using the Poisson distribution with $\lambda=1$, which the authors show to be an approximation of the sampling with replacement procedure from the static bagging algorithm \citep{Breiman1996}. \cite{oza} also presented Online Boosting, in which if an instance is misclassified by a model, its weight is increased when shown to other classifiers to be trained. Otherwise, it is decreased. This process focus training on instances that are hard to be classified, as on the classic boosting algorithm \citep{FREUND1995}.

Later on, authors have used the online bagging algorithm with different values for the $\lambda$ parameter. Leveraging Bagging \citep{bifet2010} have extended the Online Bagging by employing $\lambda=6$, and by the addition of the \ac{ADWIN} drift detector to adapt to concept drift, aiming to replace outdated classifiers. Most of the ensemble methods for concept drift that use online bagging seems to prefer to use $\lambda=6$. By taking into account the Poisson distribution, if $\lambda=6$, we have that $Poisson(X\geq1)=99.75\%$, i.e., there is a high chance that all of the classifiers will be trained on the incoming instances at least once. This may lead to low-diversity ensemble, given that most of the classifiers are trained on all of the instances. This $\lambda$ value was also adopted by \ac{ARF}, which trains \ac{HT} classifiers through online bagging, and each classifier carries its own drift detector \citep{gomes2017}. In \ac{ARF}, drifted classifiers are replaced by new ones trained in recent data.

\cite{minku2012}, coming from the assumption that high-diversity ensemble is better after concept drift, and low-diversity is preferable under stable concepts \citep{minku2010}, proposed \ac{DDD}, which maintains two ensembles trained with different $\lambda$ values. The low-diversity ensemble is trained by using $\lambda=1$. High diversity is induced by employing lower values for $\lambda$, which in \ac{DDD} is chosen by empirically testing different values for different datasets. When concept drift is detected, the high diversity ensemble is used to perform the predictions. When the concept is stable, the low diversity ensemble is used.

\ac{SEA} \citep{street2001} trains classifiers on the incoming batches of data. The ensemble is updated only if the new classifier trained on the latest labeled batch is able to improve the ensemble's performance. \ac{AWE} \citep{wang2003} weights the classifiers based on their performance to the most recent batches of data, i.e., classifiers that better classify the last arrived batch are given higher weights when voting. \ac{AUE} \citep{brzezinski2011} includes incremental learners and updates the classifiers' weights based on the current data distribution. The inclusion of incremental learners, different from \ac{AWE}, makes the ensemble members of \ac{AUE} able to be incrementally updated. Thus, it is able to update the classifiers on newest labeled instances.

\ac{OAUE} \citep{BRZEZINSKI2014}, as the name suggest, is an online version of the \ac{AUE} algorithm. \ac{OAUE} updates the ensemble one instance at a time, characterizing an online processing of data (see Section \ref{subsec:batch-vs-online}). The weights of the models are immediately updated on the latest instance, ensuring a more effective real-time adaptation to concept drift.

\ac{CALMID} utilizes a hybrid labeling strategy, combining randomness, uncertainty, and a selective sampling strategy to choose samples to be requested for labels \citep{liu2021}. In \ac{CALMID}, arrived labeled samples are weighted according to the difficulty and imbalance rate of its class. The ensemble is then trained on the weighted sample by using online bagging. \ac{CALMID} also keeps two different sliding windows: 1) label sliding window, to be aware of the class distributions; 2) sample sliding window, which is used to store the most recent instances of each class. Concept drift is addressed through the \ac{ADWIN} drift detector. When \ac{ADWIN} detects a concept drift, a new classifier is trained on the sample sliding window and is added to the ensemble. Then the less accurate classifier is pruned from the ensemble.

\ac{EOCD} \citep{feitosa2021} dynamically updates the weight of classifiers in an ensemble based on their performance over time. Initially, classifiers are trained on the initial data, with the most accurate ones set as active and the rest as inactive. Classifiers whose weights fall below a threshold are replaced by inactive classifiers, and a new inactive classifier is trained on the most recent instances. An optimization technique is employed to select new active classifiers and their weights, while the remaining classifiers are moved to an inactive set and trained offline. Concept drift is detected using \ac{RDDM} \citep{BARROS2017}, and adaptation involves choosing new ensemble members and training new classifiers.

\ac{HE-CDTL} aims at the transfer learning domain, using an adaptive weighted correlation alignment to improve the transfer of knowledge between domains \citep{yang2022}. Like \ac{SEA}, classifiers are trained on the incoming batches of data, and are combined through a class-wise ensemble strategy. \ac{SEOA} \citep{Guo2021} uses two different types of neural networks: deep and shallow. The rationale is that shallow neural networks are able to learn a new concept more quickly if compared to deep neural networks. The choice of which one to use is done based on the error-rate. A high error-rate indicates a changing distribution, thus the shallow neural network is used. Otherwise, the deep neural network is used for the predictions. In other words, the deep neural network is used in stable concepts, and shallow neural networks under concept drift.

\ac{SCHCDOE} maintains the model untouched under stable concepts. When data distribution is changing (warning level is given), \ac{SCHCDOE} updates its model by using random subspaces \citep{liu2023}. Under concept drift, abnormal data are discarded, and the training is performed by combining ensemble learning and incremental learning. \ac{CCEKL} aims at the class imbalance issue \citep{CHEN2024b}. The authors present a misclassification cost to adapt to changing class distribution. The adaptation to concept drift is done trough a continuous kernel learning method.

\subsubsection{Dynamic Ensemble}\label{sec:dynamicSelection}

\ac{DS} approaches for dealing with data streams with concept drift usually train classifiers on the incoming instances by adopting a batch-based processing or some strategy of online processing. The idea is to maintain an updated pool of classifiers $C$ and to keep the earliest instances as the \ac{DSEL}. \ac{AO-DCS} constructs the subsets in which the competence of the classifiers is measured through the values of the features instead of using the most commonly used neighborhood-based approach \citep{zhu2004}. Incoming instances are divided into data chunks, in which each one is used for training a base classifier and also constructing an evaluation set $Z$, which can be considered as the \ac{DSEL}. Numerical values are discretized, and instances that share the same values in the same features belong to the same subset. The competence of each classifier is measured on the subset of the test instance, and the single classifier with the best average accuracy in the subset is chosen for classification.

The Dynse Framework \citep{almeidaEtAl2018} trains classifiers in batches of data arrived in time and uses a sliding window as a \ac{DSEL} in order to always maintain recent instances to measure the competence of the classifiers. \ac{DDCS} \citep{cavalheiro2021} applies resampling to update the classifiers in $C$, while still training classifiers in new batches of data. \ac{DynEd} \citep{abadifard2023} uses diversity-based ranking in order to select the classifiers, and when a drift is triggered by an \ac{ADWIN} detector, a new classifier is trained on the last instances from a sliding window, and is added to the pool $C$. Classifiers in $C$ are updated with resampling, in which the $\lambda$ value changes depending on the pool diversity and fluctuations in the accuracy. A K-Means algorithm is also applied on $C$ to divide the classifiers into groups before performing the \ac{DES} itself, in which the most accurate classifiers on a sliding window are selected.

\cite{ZYBLEWSKI2021138} uses stratified bagging and preprocessing strategies to deal with the imbalanced class problem in data streams. Incoming data chunks are subjected to sampling with replacement, preserving the number of instances for both minority and majority classes. A classifier is trained on the resulting data. The arrived data chunks are the \ac{DSEL} that will be used to perform the \ac{DES}. \ac{DES-ICD} also aims at the imbalanced class issue by using oversampling strategies \citep{jiao2022}. They compare the class ratio from the most recent batch of data to the previous one and generate new samples from the minority class by considering the degree of change in the distribution. \ac{DESW-ID} tries to overcome the imbalanced data problem by adopting resampling strategies by training more classifiers with the minority class. The \ac{DS} strategies use a reverse search algorithm to pick the best number of classifiers for an ensemble. The error rate sorts the classifiers, and the authors have used \ac{ADWIN} as a concept drift detector in order to shrink the adaptive window in which the classifiers are ranked \citep{Han2023}. \ac{AB-DES} is a chunk-based algorithm that employs a dual adaptive sampling technique for training the classifiers \citep{wei2024}. \ac{AB-DES} stores data chunks obtained from an undersampling method and applies oversampling to minority classes. In addition, newer data chunks have a higher weight than the oldest ones, and a sliding window-based drift detector is used to adapt to concept drift.

\cite{assis2023} argue on the unfeasibility of using neighborhood-based \ac{DS} methods in data streams and propose \ac{MSTS}, in which the \ac{RoC} are the ten last instances in the data stream. Thus, there is no need to do a neighborhood search.

\ac{ARE} \cite{paim2024} has induced diversity by training \ac{HT}s in different random subspaces and used only incorrectly classified instances for the training. Classifier selection is performed by choosing the classifiers that are able to correctly classify the last arrived instances, as done in \ac{MSTS} \citep{assis2023}. \ac{ARTE} \citep{PAIM2025112830} has used the same structure of \ac{ARF} for training, maintaining an ensemble with the members carrying each a drift detector. In \ac{ARTE}, each ensemble member is trained on a different random feature subspace. Additionally, the cut points for splitting the trees are chosen randomly. The ensemble members are also chosen based on the performance to the most recent instances.

In Table \ref{tab:concept-matrix-ds-drift}, we present a concept matrix of the described \ac{DS} methods for concept drift. Many of the cited methods rely on the brute force \ac{kNN} to define the \ac{RoC}. The proposed framework uses the Online K-d tree to define the \ac{RoC} aiming to mitigate the limitation of the validation set and adopt a different training strategy than other methods.


\begin{table}[htb]
    \scriptsize
    \centering
    \caption{Concept matrix of DS methods for concept drift.}
    \label{tab:concept-matrix-ds-drift}
    \begin{tabular}{l l l l l}
        \toprule
        Method & Type & Drift Detector & \ac{RoC} Definition & Training \\
        \midrule
        \ac{AO-DCS} \citep{zhu2006} & Batch & -- & Attribute-Oriented & Streaming Batches \\
        Dynse \citep{almeidaEtAl2018} & Batch & -- & brute force \ac{kNN} & Streaming Batches \\
        DDCS \citep{cavalheiro2021} & Online & -- & brute force \ac{kNN} & Online Bagging \\
        \citep{ZYBLEWSKI2021138} & Batch &  & brute force kNN & Streaming Batches \\
        DES-ICD \citep{jiao2022} & Online & \ac{ADWIN} & brute force \ac{kNN} & Streaming Batches \\
        DynEd \citep{abadifard2023} & Online & \ac{ADWIN} & -- & Online Bagging \\
        DESW-ID \citep{Han2023} & Online & \ac{ADWIN} & Sliding Window & Streaming Batches \\
        MSTS \citep{assis2023} & Online & -- & Sliding Window & Online Boosting \\
        \ac{AB-DES} \citep{wei2024} & Batch & Sliding Window-based & brute force \ac{kNN} & Adaptive Bagging \\
        \ac{ARE} \citep{paim2024} & Online & \ac{ADWIN} & Sliding Window & Online Bagging \\
        \ac{ARTE} \citep{PAIM2025112830} & Online & \ac{ADWIN} & Sliding Window & Online Bagging \\\hline
        \textbf{IncA-DES} & \textbf{Online} & \textbf{Any} & \textbf{Online K-d tree} & \textbf{Incremental} \\
        \bottomrule
    \end{tabular}
\end{table}

\section{Experiments}\label{sec:experiments}

In this section, we define the datasets used in the experiments, present the experimental protocol, and compare the proposed framework to the state-of-the-art methods. Additionally, we do an Ablation Study comparing different components of \ac{IncA-DES} and a Case Study where we assess the scalability of the Online K-d tree when the search space increases.

\subsection{Datasets}

In this work, we divide datasets into real-world datasets and synthetic datasets. The real-world datasets offer more realistic scenarios in which we do not know about the presence or nature of concept drift. Synthetic datasets contain induced concept drift, making them applicable to evaluate the behavior of a \ac{ML} model to known changing distributions. 

Table \ref{tab:datasets} exposes the datasets used in the experiments and their characteristics. They were gathered from the USP DS Repository \citep{Souza_2020}, the MOA Repository \citep{moa}, and the UCI \citep{uci}. The synthetic datasets were generated using the \ac{MOA} framework \citep{moa}. To the SEA, Sine STAGGER and Agrawal datasets, 1,000,000 instances were generated, and the concept has changed abruptly every 10,000 instances, characterizing a recurrent concept drift. On the Hyperplane and LED datasets, we have a gradual concept drift starting at the 50,000th instance, and the new concept stabilizes at the 50,500th instance. On the RandomRBF dataset, there is an incremental concept drift throughout the whole stream. For the Hyperplane, LED, and RandomRBF datasets, 100,000 instances are generated. Other details about these datasets are the same as the default in the \ac{MOA} framework.

\begin{figure}
    \centering
    \includegraphics[width=0.8\linewidth]{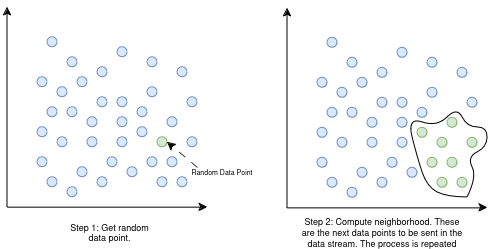}
    \caption{Steps for simulating \textit{Virtual Concept Drift}}
    \label{fig:steps-simulate}
\end{figure}

Furthermore, on the Letters, Digits, Pen Digits, Dry Bean, and Rice datasets, a \textit{virtual concept drift} was induced by dividing the data stream into feature spaces, as done by \cite{almeidaEtAl2018}. To do so, in each step, a single random instance is chosen from the dataset, and the 199 nearest neighbors to it according to the Euclidean Distance are sent to the data stream. Thus, we use a chunk of 200 instances, as it is our training size. This process is repeated until the dataset is empty. This way, we simulate an environment where different regions of the feature space become available with time. Figure \ref{fig:steps-simulate} shows the steps of this process. All of the datasets used in the experiments of this work are available in our GitHub repository\footnote{https://github.com/eduardovlb/IncA-DES}.

\begin{table}[!htb]
    \centering
    \caption{Datasets used in the experiments. A: Abrupt, G: Gradual, I: Incremental, R: Recurrent, V: Virtual.}
    \small
    \begin{tabular}{l c c c cc }
        \toprule
        Dataset & Source & Type & \# Instances & \# Features & \# Classes \\
        \midrule
        \textit{Real-world} \\
        Electricity & \cite{harries1999} & Unknown & 45,312 & 8 & 2 \\
        Nursery & \cite{rajkovic1997} & Unknown & 12,960 & 8 & 5 \\
        Ozone & \cite{zhang2008} & Unknown & 2,536 & 72 & 2 \\
        Gas Sensor & \cite{vergara2012} & Unknown & 13,910 & 128 & 6 \\
        Adult & \cite{becker1996} & Unknown & 45,222 & 14 & 2 \\
        Rialto & \cite{losing2016} & Unknown & 82,250 & 27 & 10 \\
        NOAA & \cite{ditzlerEtAl2012} & Unknown & 18,159 & 8 & 2 \\
        Keystroke & \cite{souza2015} & Unknown & 1,600 & 10 & 4 \\
        Covertype & \cite{jock1998} & Unknown & 581,012 & 54 & 7 \\
        Yeast & \cite{yeast_110} & Unknown & 1,484 & 8 & 10 \\
        Asfault & \cite{SOUZA2018} & Unknown & 8,066 & 62 & 5 \\
        Insects-AB & \cite{Souza_2020} & A & 52,848 & 33 & 6 \\
        Insects-GB & \cite{Souza_2020} & G & 24,150 & 33 & 6 \\
        Insects-IB & \cite{Souza_2020} & I & 57,018 & 33 & 6 \\
        Letter & \cite{slate1991} & V & 20,000 & 16 & 26 \\
        Digits & \cite{alpaydin1998} & V & 5,620 & 64 & 10 \\
        Pen Digits & \cite{alpaydin1998} & V & 10,992 & 16 & 10 \\
        Dry Bean & \cite{koklu2020} & V & 13,611 & 16 & 7 \\
        Rice & \cite{cinar2019} & V & 3,810 & 7 & 2 \\
        \hline
        \textit{Synthetic} \\
        SEA & \cite{street2001} & A, R & 1,000,000 & 3 & 2 \\
        STAGGER & \cite{schlimmer1986} & A, R & 1,000,000 & 3 & 2 \\
        Sine & \cite{gama2004} & A, R & 1,000,000 & 2 & 2 \\
        Agrawal & \cite{agrawal1993} & A & 1,000,000 & 9 & 2 \\
        Hyperplane & \cite{hulten2001} & G & 100,000 & 10 & 2 \\
        LED & \cite{Breiman1996} & G & 100,000 & 24 & 10 \\
        RandomRBF & \cite{bifet2009a} & I & 100,000 & 10 & 2 \\
        \bottomrule
    \end{tabular}
    \label{tab:datasets}
\end{table}

\subsection{Experimental Protocol}
\label{sec:exp-protocol}

The experiments in this work were designed to compare different methods and approaches in different scenarios, i.e., different dimensionality, sample size, and label availability. To do so, various real-world and synthetic datasets were used and two different training policies were adopted. The first one is the test-then-train, where every instance is used for testing and then for training \citep{gama2014}. Such a policy considers that all of the labels are available for the dataset without delay.

The second training strategy introduces two limitations in the training policy to make testing closer to real-world environments: 1) delayed arrival of labels and 2) partial availability of the labels. The first limitation is straightforward: a delay is applied to when a label becomes available after it was used to test the model's prediction performance. Since the delay of labels in the real world varies according to the problem, it is difficult to define how long a label takes to arrive. For instance, in weather forecasts, information about whether it has rained a day is available the next day. As a counterpart, other datasets require a longer delay for the labels and often require human action. \cite{gomes2017} employs a delay of 1,000 instances, which we use in this work, in addition to two other levels: equal to our training size (200) and 1/4 of the training size (50).

On the second limitation, we considered that, for every two instances in the data stream, only one is labeled, which we call partial labeling. These two limitations were introduced together with the aim of better simulating an environment in the real world. The chosen evaluation metrics were the average accuracy, the prequential accuracy on a sliding window of 1,000 instances \citep{gama2014}, and the average number of Instances per second (I/s).

The baseline state-of-the-art methods used for comparison include three \ac{DS} methods for concept drift and four ensemble methods for concept drift, which are listed below:

\begin{itemize}
    \item ARE: A \ac{DS} method that trains classifiers only on the instances that the ensemble is not able to classify correctly. Each classifier is trained in a different random region of the subspace to induce diversity, and the classifiers to compose the ensemble are chosen according to their performance in the last arrived labeled instances \citep{paim2024}.
    \item Dynse: A batch-based \ac{DS} method that trains classifiers whenever a batch of labeled data arrives in the timestamp \citep{almeidaEtAl2018}. 
    \item DynED: An online \ac{DS} method for concept drift that divides the classifiers into different groups through a K-means algorithm and chooses the best ones to compose an ensemble based on diversity. The training is done through an online bagging algorithm that updates the $\lambda$ parameter with time \citep{abadifard2023}. 
    \item \ac{ARF}: A well-known ensemble-based method for concept drift where each inner classifier carries a concept drift detector. Classifiers are trained through online bagging, and whenever a single classifier is considered outdated by the drift detector, it is replaced \citep{gomes2017}.
    \item OzaBag: This is the original online bagging, which adapts the batch-based bagging \citep{Breiman1996} to online learning \citep{oza}.
    \item LevBag: An extension to OzaBag, which increases the $\lambda$ parameter and includes output detection codes \citep{bifet2010}.
    \item \ac{OAUE}: An online variation of \ac{AUE}, which trains its classifiers on incoming instances, weights them based on the prediction error and frequently replaces non-accurate classifiers with new ones \citep{BRZEZINSKI2014}.
\end{itemize}

Before the comparison, a hyperparameter tuning for \ac{IncA-DES} was performed, where the impact of each hyperparameter ($trigger$, $D$, $F$, $k$ and $\omega$) is measured individually. More details can be found in Appendix \ref{app:hyp}. The default configuration after hyperparameter tuning is exposed in Table \ref{tab:set-hyperparameters}. Datasets used for the hyperparameter tuning (Electricity, NOAA, Nursery, and Digits) were excluded from the comparison with the state-of-the-art methods. Details about the state-of-the-art hyperparameters are in Appendix \ref{app:art-hyp}.

\begin{table}[htb]
    \centering
    \caption{Set of Hyperparameters of IncA-DES. $D$ is the pool size, $F$ is the maximum number of instances a classifier receives for training, $k$ is the number of neighbors, $\omega$ the rate of the majority class in the \ac{RoC}, and $\beta$ is the proportion of inactive nodes in the Online K-d tree.}
    \begin{tabular}{l c}
        \toprule
        Hyperparameter & Value \\
        \midrule
        $W$ & Limited by Memory \\
        $Trigger$ & \ac{RDDM} \\
        $D$ & 75 \\
        $F$ & 200 \\
        Pruning Engine & Age-based \\
        DS method & KNORAE \\
        $k$ & 5 \\
        $\omega$ & 0.8 \\
        Distance Function & Canberra Distance \\
        $\beta$ & 0.3 \\
        Base Classifier & Hoeffding Tree \\
        \bottomrule
    \end{tabular}
    \label{tab:set-hyperparameters}
\end{table}

Since the \ac{DSEW} has its size limited solely by the memory, none of the datasets tested needed a deletion operation in \ac{IncA-DES}. Therefore, the $\beta$ parameter defined for the K-d tree did not influence \ac{IncA-DES}. Despite that, after some empirical study, we recommend setting $\beta=0.3$, i.e., if 30\% of the nodes are inactive, the tree is rebuilt. This should give a good balance between binary search time and an effective neighborhood definition.

Later, we assess how the components settled for \ac{IncA-DES} (K-d tree and Overlap-based classification) influence accuracy performance and processing time. In these experiments, seven real-world datasets with different dimensionalities were used to cover different scenarios equally. In the end, we have used three synthetic datasets to evaluate how the Online K-d tree differs from the brute force \ac{kNN} in different search spaces. Since with the synthetic datasets we can control the environment and generate as many instances as needed, different sizes for the \ac{DSEW}, i.e., the search space $n$, were considered to compare how the K-d tree influences the accuracy, processing time and I/s. Only one concept of the datasets was generated, and each search space was used to label 100,000 instances.

The experiments were performed using Java 17 running the \ac{MOA} framework \citep{moa}, except for the DynED \citep{abadifard2023}, which was built in Python 3.8 on the scikit-multiflow library \citep{skmultiflow}. All of the results reported are an average of 10 runs.

\subsection{Comparing With State of the Art}

\subsubsection{Experiments With Test-then-train}

Table \ref{tab:acc-art} shows the average accuracy of the tested methods and the number of wins (W), ties (T), and losses (L) of \ac{IncA-DES} to the respective methods. Notice that \ac{IncA-DES} got the best average accuracy, followed by DynED and ARE, and it also had more wins than losses compared to all of the state-of-the-art methods, except DynED, to which there was a tie. \ac{IncA-DES} and \ac{ARE} were tied in the first place in the average rank.

\begin{figure}
    \centering
    \includegraphics[width=0.8\linewidth]{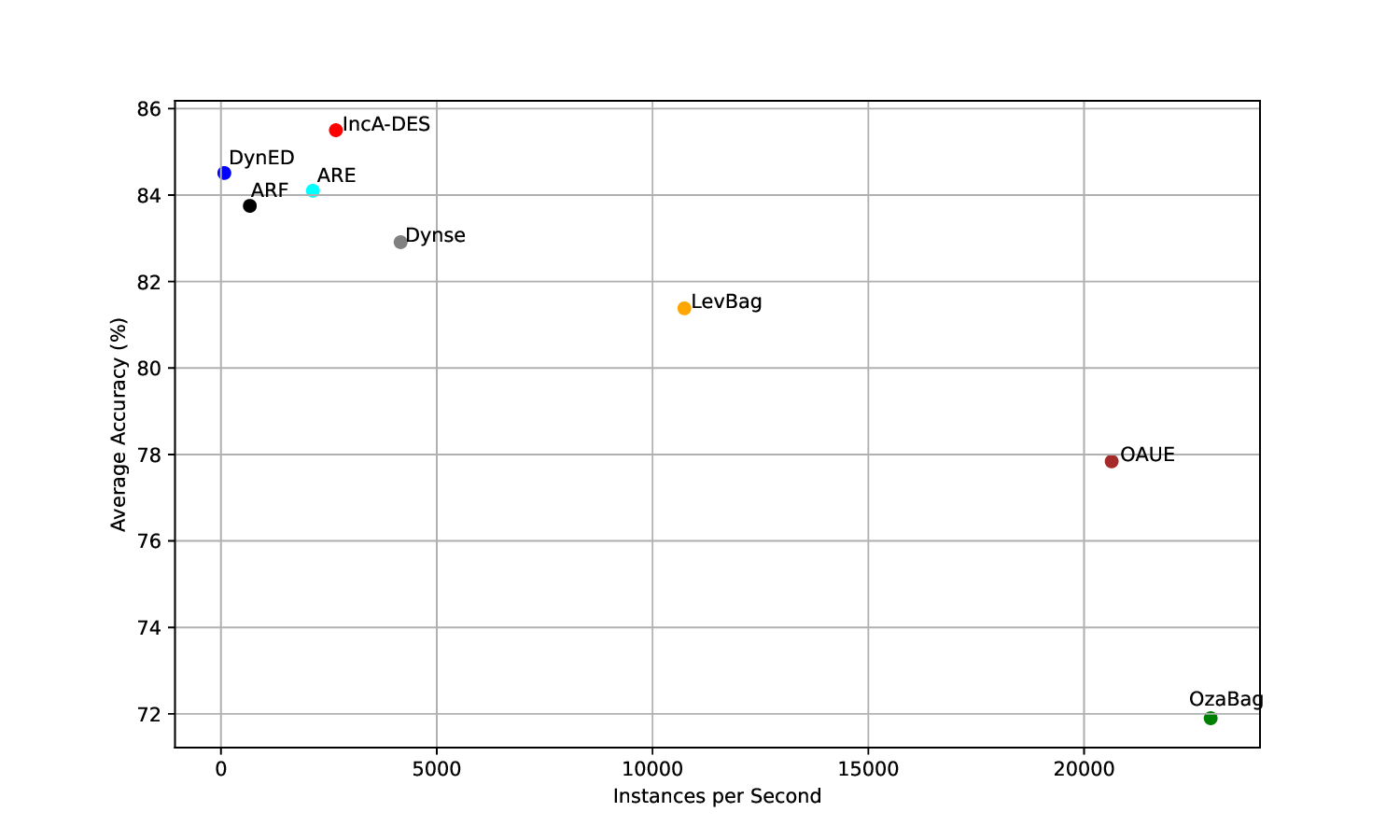}
    \caption{Plot of average accuracy versus average instances per second of the state-of-the-art methods on all tested datasets.}
    \label{fig:plot-acc-time}
\end{figure}

In Figure \ref{fig:plot-acc-time}, we see a plot of accuracy versus I/s of the state-of-the-art methods, where we can see that \ac{IncA-DES}, when compared to \ac{ARE}, DynED and \ac{ARF}, was not only the most accurate one but also the one that was able to process more instances in one second, although it uses neighborhood search for defining the \ac{RoC}. More specifically, it processed 1.25 times more instances than \ac{ARE} in one second, 34.57 times more instances than DynED, and 3.98 times more instances than \ac{ARF}. \ac{IncA-DES} also relies on a simpler training policy, where classifiers are trained incrementally one by one. It looks a lot like the training approach used in Dynse \citep{almeidaEtAl2018}, but due to the higher \ac{DSEW} and the need to compute the neighborhood in the training phase as well (to update the drift detector), it was not faster than Dynse.

Although other methods like OzaBag and OAUE were the fastest, they were the least accurate -- OzaBag had a difference of 13.60 percentage points to IncA-DES and OAUE of 7.67 percentage points. LevBag and Dynse got good accuracy with more I/s, but still not as accurate as \ac{IncA-DES}, \ac{ARE}, or DynED. For the interested reader, Table \ref{tab:time-art}, found in Appendix \ref{app:time-art}, shows the average I/s for each method.

\begin{table}[!htb]
\centering
\small
\caption{Accuracies of IncA-DES and methods of the state of the art. The W-T-L row counts the mumber of wins, ties, and losses of the methods compared to the proposed framework.}
\begin{tabular}{l c c c c c c c c}
    \toprule
        Dataset & IncADES & ARE & Dynse & DynED & ARF & OzaBag & LevBag & OAUE \\ \midrule
        Ozone & 93.84 & \textbf{94.50} & 93.62 & 93.25 & 94.32 & 93.96 & 93.86 & 93.96 \\ 
        Gas Sensor & \textbf{96.76} & 91.01 & 91.19 & 94.92 & 91.12 & 55.45 & 82.51 & 71.93 \\ 
        Adult & 83.26 & 83.05 & 82.18 & 83.49 & 84.14 & \textbf{84.43} & 84.25 & 84.35 \\ 
        Rialto & \textbf{82.43} & 81.83 & 79.71 & 74.02 & 72.82 & 31.01 & 60.40 & 52.90 \\ 
        Keystroke & \textbf{96.49} & 94.98 & 84.81 & 88.90 & 94.20 & 82.91 & 88.81 & 74.86 \\ 
        Insects-AB & \textbf{75.73} & 74.50 & 72.32 & 74.60 & 75.17 & 57.61 & 69.13 & 64.64 \\ 
        Insects-GB & \textbf{79.38} & 77.69 & 75.61 & 76.16 & 78.03 & 61.86 & 71.99 & 59.69 \\ 
        Insects-IB & 61.70 & \textbf{65.93} & 60.60 & 64.11 & 65.70 & 54.33 & 61.35 & 56.92 \\ 
        Yeast & 53.27 & 55.91 & 49.07 & \textbf{56.40} & 51.17 & 49.67 & 49.77 & 44.24 \\ 
        Asfault & 88.25 & 87.81 & 85.66 & 88.38 & \textbf{88.67} & 71.49 & 87.06 & 77.53 \\ 
        Covertype & 93.00 & \textbf{95.47} & 92.82 & 94.69 & 94.88 & 83.66 & 91.77 & 90.03 \\ 
        Pen Digits & \textbf{95.55} & 94.49 & 93.23 & 90.02 & 92.61 & 86.51 & 89.53 & 86.97 \\ 
        Dry Bean & 89.66 & \textbf{90.62} & 89.86 & 90.12 & 88.47 & 89.50 & 89.23 & 88.00 \\ 
        Rice & 91.43 & 92.02 & 90.13 & 91.79 & \textbf{92.42} & 91.61 & 91.76 & 88.57 \\ 
        Letters & \textbf{85.34} & 50.82 & 82.66 & 65.30 & 58.70 & 61.94 & 64.33 & 63.81 \\ 
        SineRec & 93.98 & 96.76 & 89.80 & \textbf{99.14} & 96.98 & 57.06 & 93.36 & 90.62 \\ 
        SeaRec & 87.05 & 87.98 & 85.87 & \textbf{89.05} & 87.89 & 84.32 & 87.16 & 86.77 \\ 
        StaggerRec & \textbf{99.91} & 99.71 & 99.70 & 99.73 & 99.79 & 72.36 & 98.02 & 97.87 \\ 
        Agrawal & 83.60 & 80.99 & 79.00 & \textbf{94.07} & 80.36 & 61.97 & 81.17 & 85.03 \\ 
        Hyperplane & 87.48 & 88.55 & \textbf{89.83} & 84.76 & 86.91 & 87.35 & 87.74 & 88.43 \\ 
        LED & 69.52 & 73.72 & 70.00 & 73.22 & \textbf{73.83} & \textbf{73.83} & 73.78 & 73.72 \\ 
        RandomRBF & 93.47 & \textbf{94.23} & 86.30 & 93.05 & \textbf{94.23} & 89.04 & 93.40 & 91.58 \\ \bottomrule
        Average & \textbf{85.50} & 84.10 & 82.91 & 84.51 & 83.75 & 71.90 & 81.38 & 77.84 \\ 
        Av. Rank & \textbf{3.23} & \textbf{3.23} & 5.32 & 3.45 & 3.27 & 6.41 & 4.86 & 6.05 \\
        W-T-L & - & 12-0-10 & 19-0-3 & 11-0-11 & 12-0-10 & 18-0-4 & 16-0-6 & 17-0-5 \\ 
        \bottomrule
\end{tabular}
\label{tab:acc-art}
\end{table}

Now let us evaluate the windowed prequential accuracy of the methods, shown in Figure \ref{fig:preqAcc-art}. We excluded the OAUE and OzaBag methods from the graphs for visualization purposes since they are the least accurate. Considering the Ozone dataset (Figure \ref{subfig:ozone-art}), we see that \ac{ARE} outperformed the other methods in most of the timestamps. On the Gas Sensor dataset, in Figure \ref{subfig:gassensor-art}, \ac{IncA-DES} had the highest accuracy on most of the timestamps and did not present drops in accuracy at many moments where other methods did. Although it did not have the best accuracy on the Adult dataset, in Figure \ref{subfig:adult-art}, we can notice that the prequential accuracy of the methods was quite the same throughout the timestamp.

\begin{figure}[!htb]
\centering
    \subfloat[Ozone.]{
    \label{subfig:ozone-art}
    \includegraphics[width=0.3\linewidth]{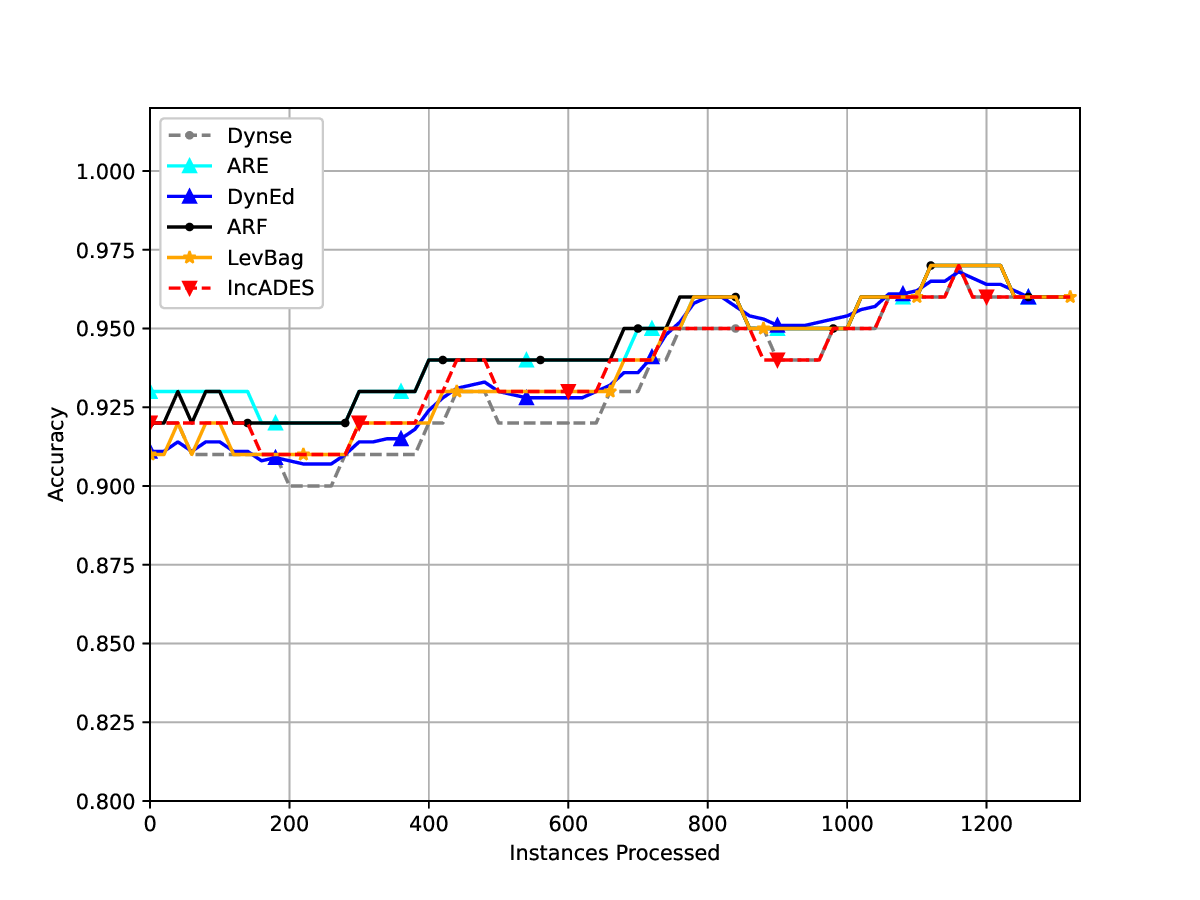}
    }
    \subfloat[GasSensor.]{
    \label{subfig:gassensor-art}
    \includegraphics[width=0.3\linewidth]{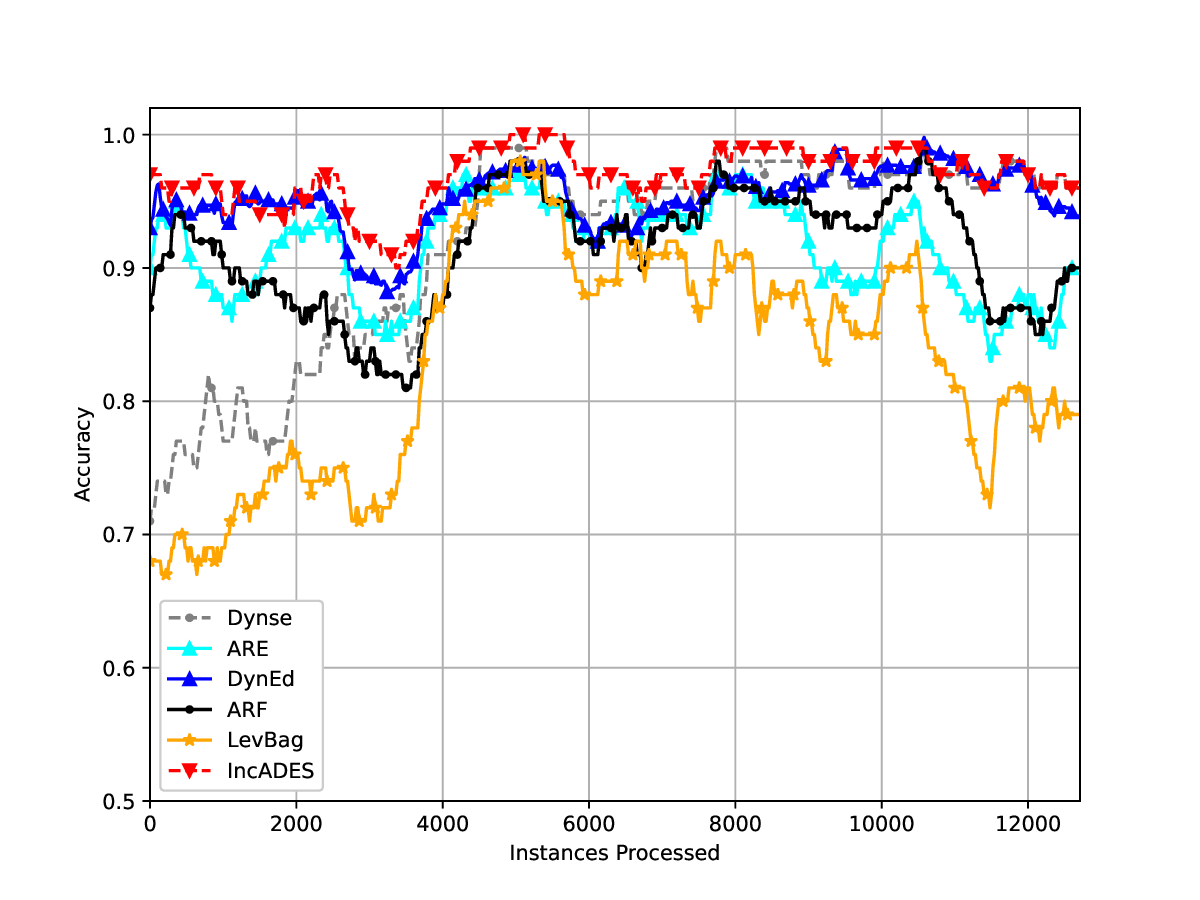}
    }
    \subfloat[Adult.]{
    \label{subfig:adult-art}
    \includegraphics[width=0.3\linewidth]{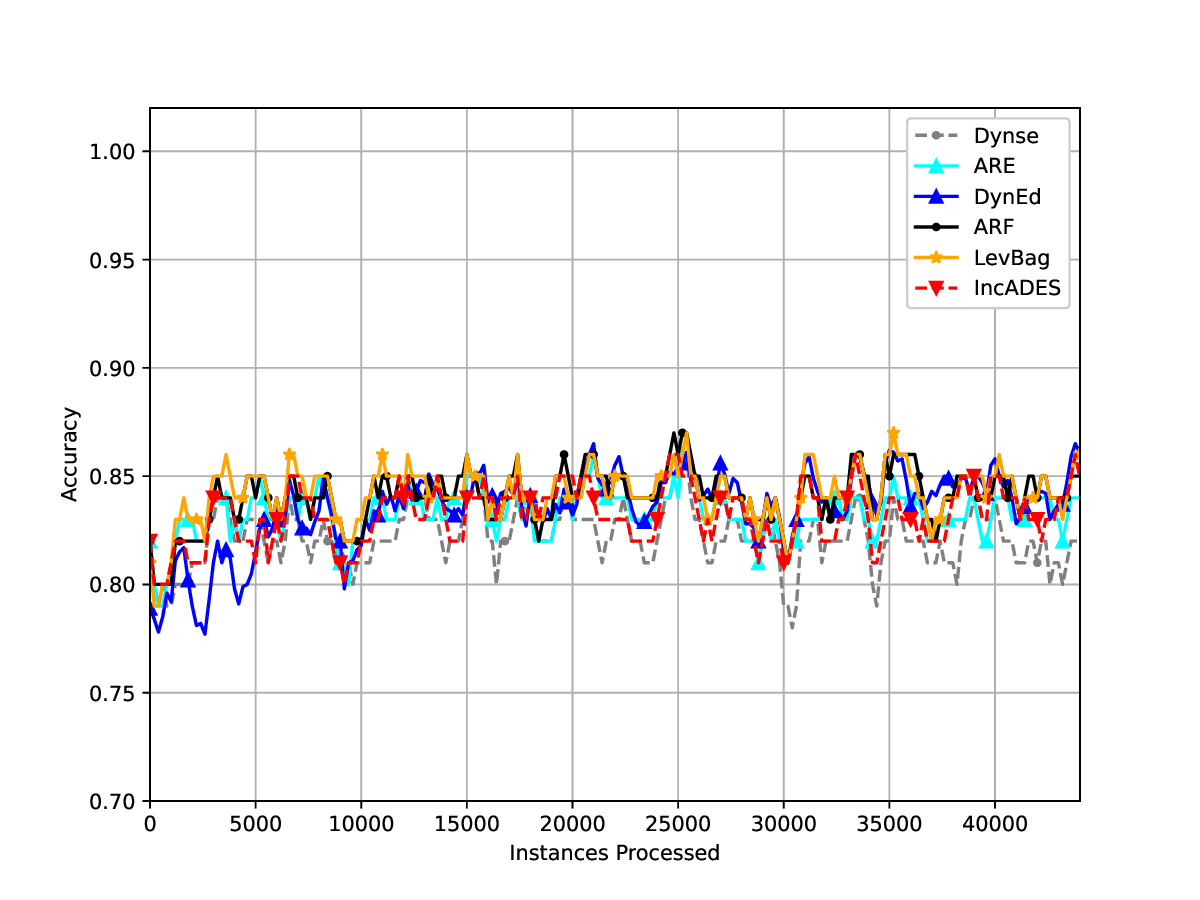}
    }\\
    \subfloat[Rialto.]{
    \label{subfig:rialto-art}
    \includegraphics[width=0.3\linewidth]{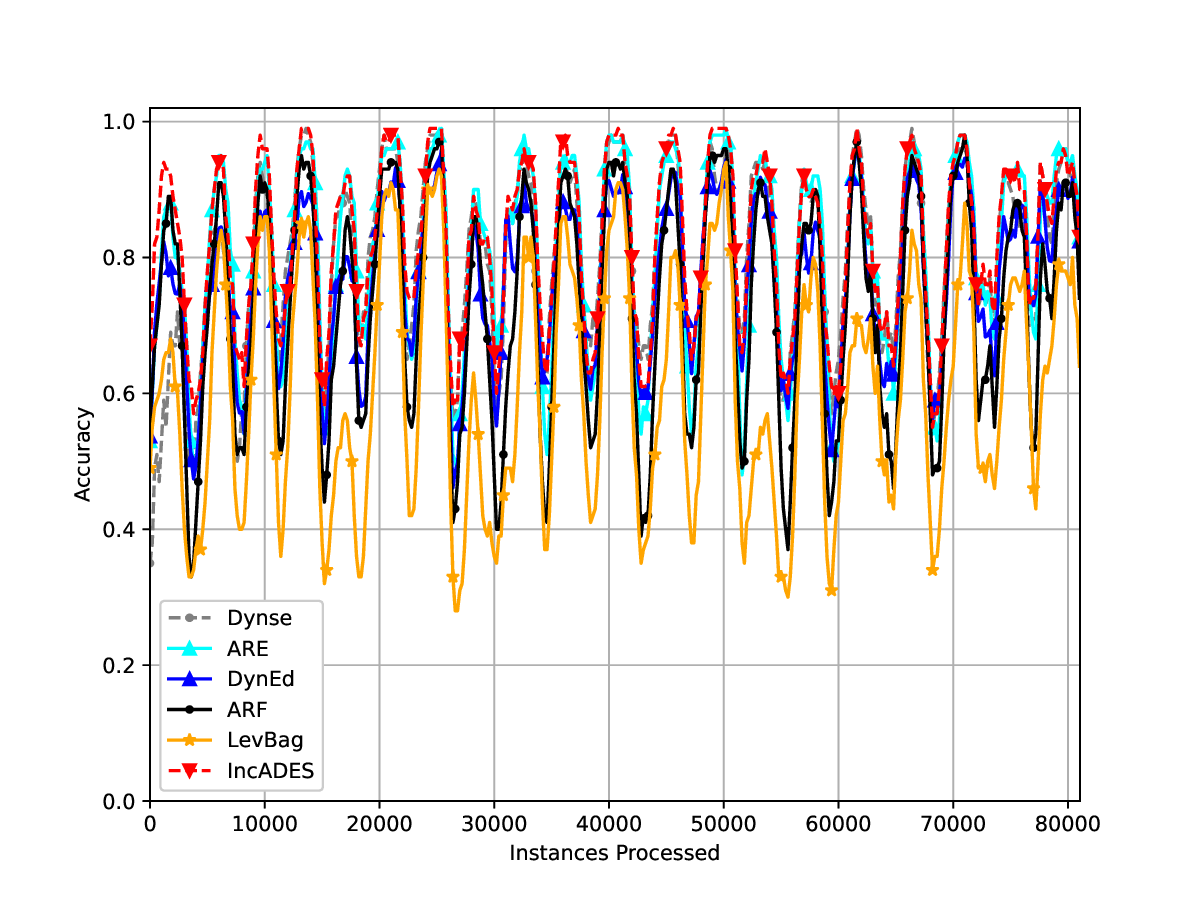}
    }
    \subfloat[Keystroke.]{
    \label{subfig:keystroke-art}
    \includegraphics[width=0.3\linewidth]{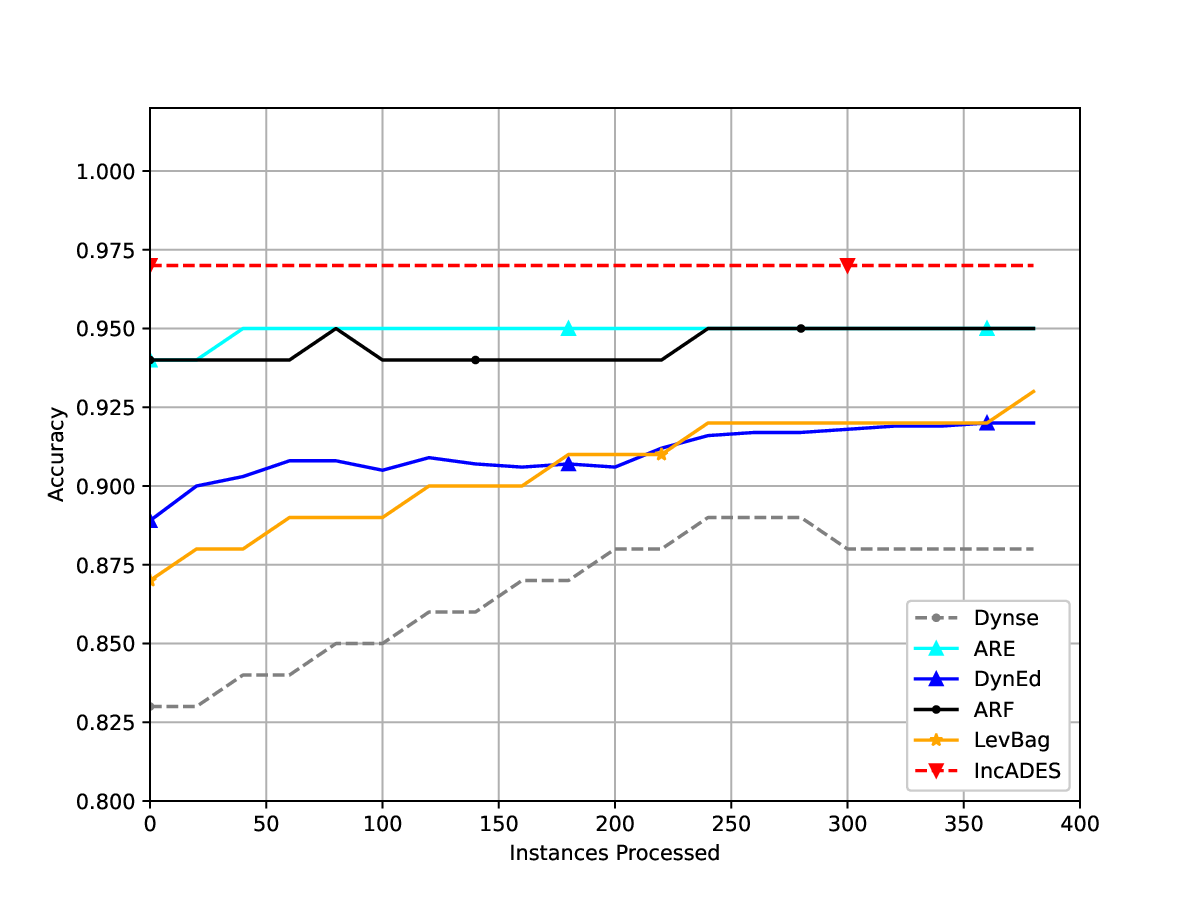}
    }
    \subfloat[Insects-AB.]{
    \label{subfig:insects-ab-art}
    \includegraphics[width=0.3\linewidth]{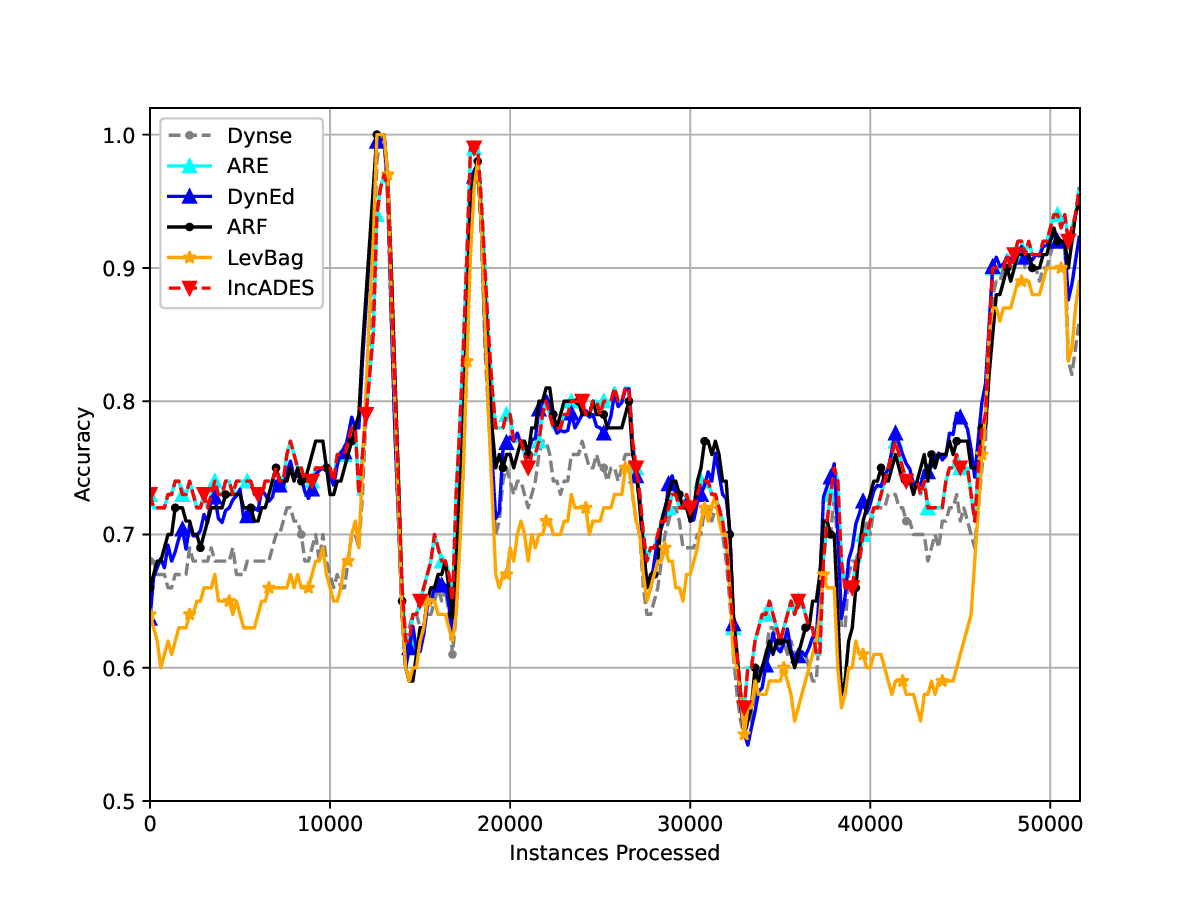}
    }\\
    \subfloat[Insects-GB.]{
    \label{subfig:insects-gb-art}
    \includegraphics[width=0.3\linewidth]{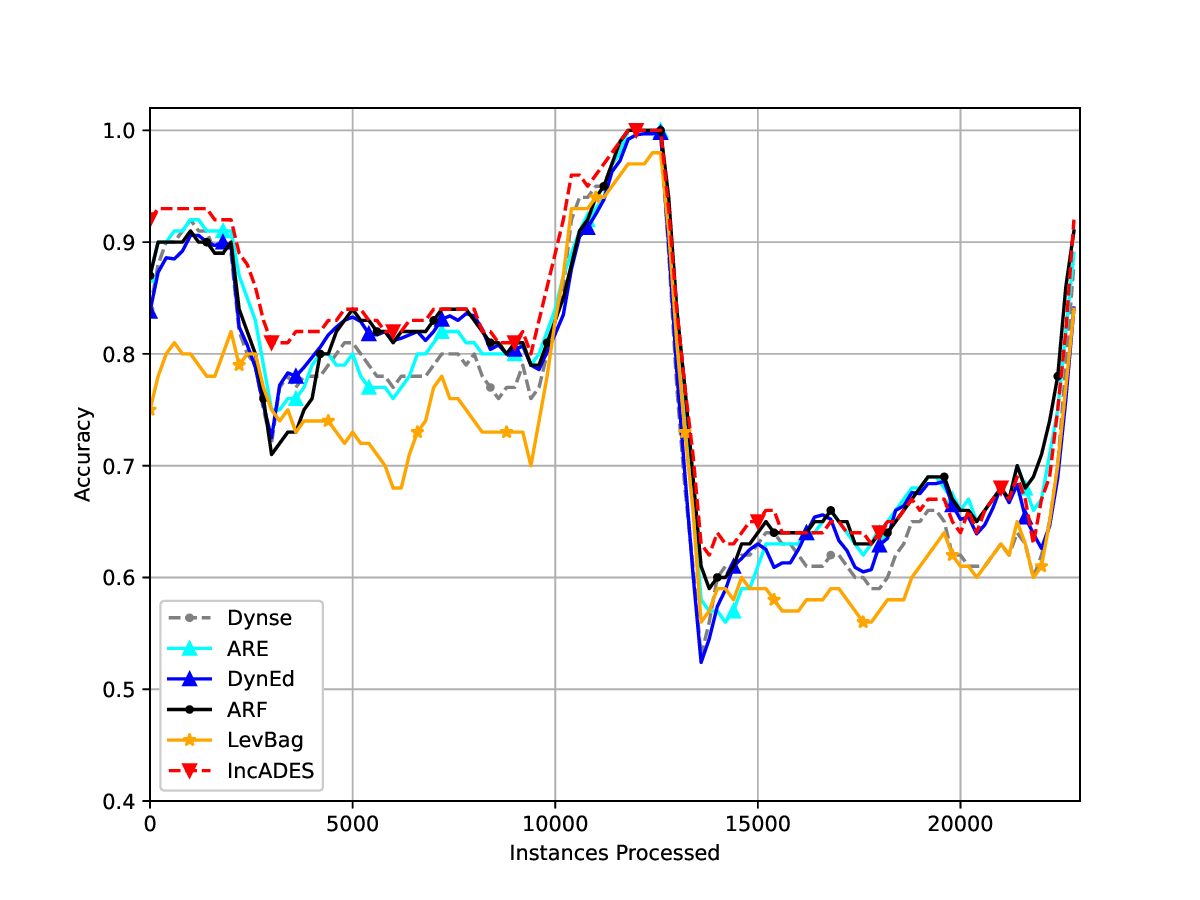}
    }
    \subfloat[Insects-IB.]{
    \label{subfig:insects-ib-art}
    \includegraphics[width=0.3\linewidth]{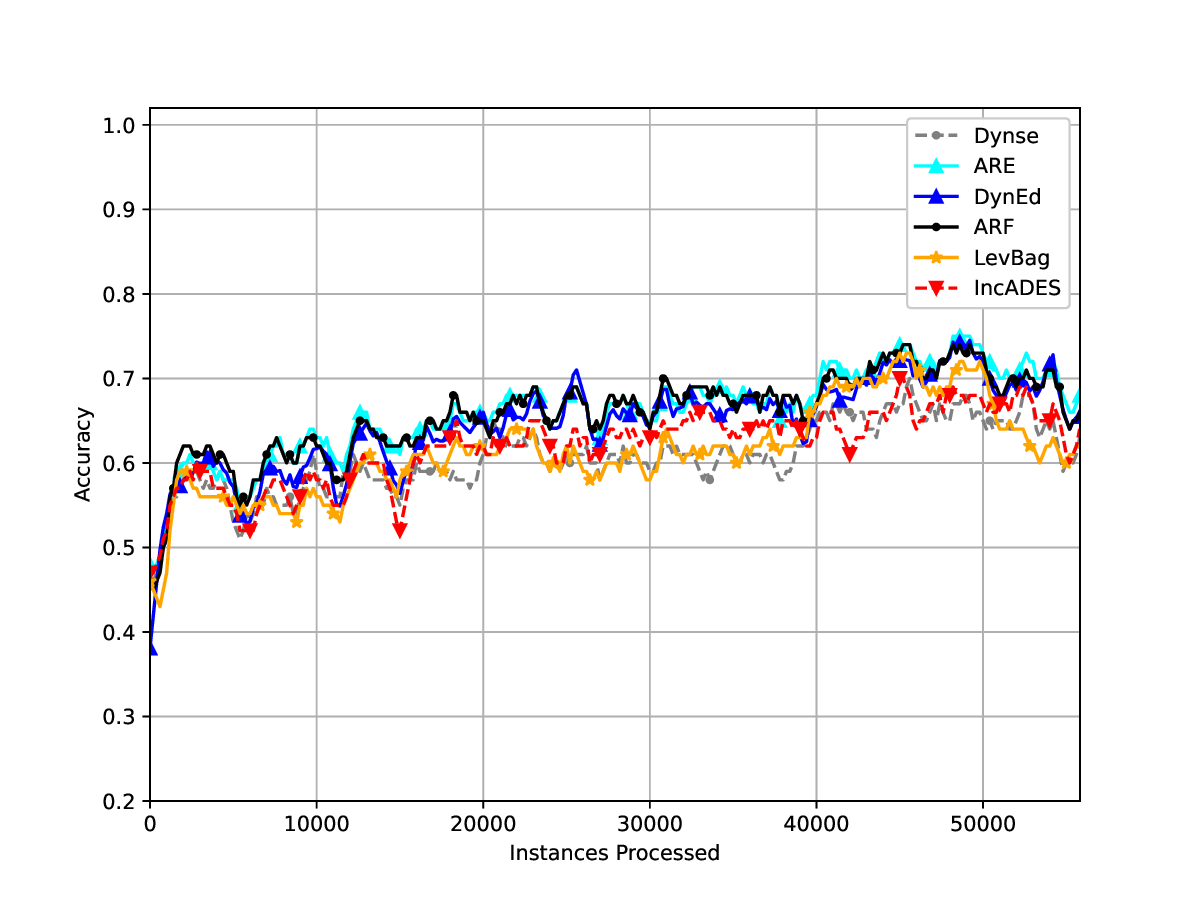}
    }
    \subfloat[Yeast.]{
    \label{subfig:yeast-art}
    \includegraphics[width=0.3\linewidth]{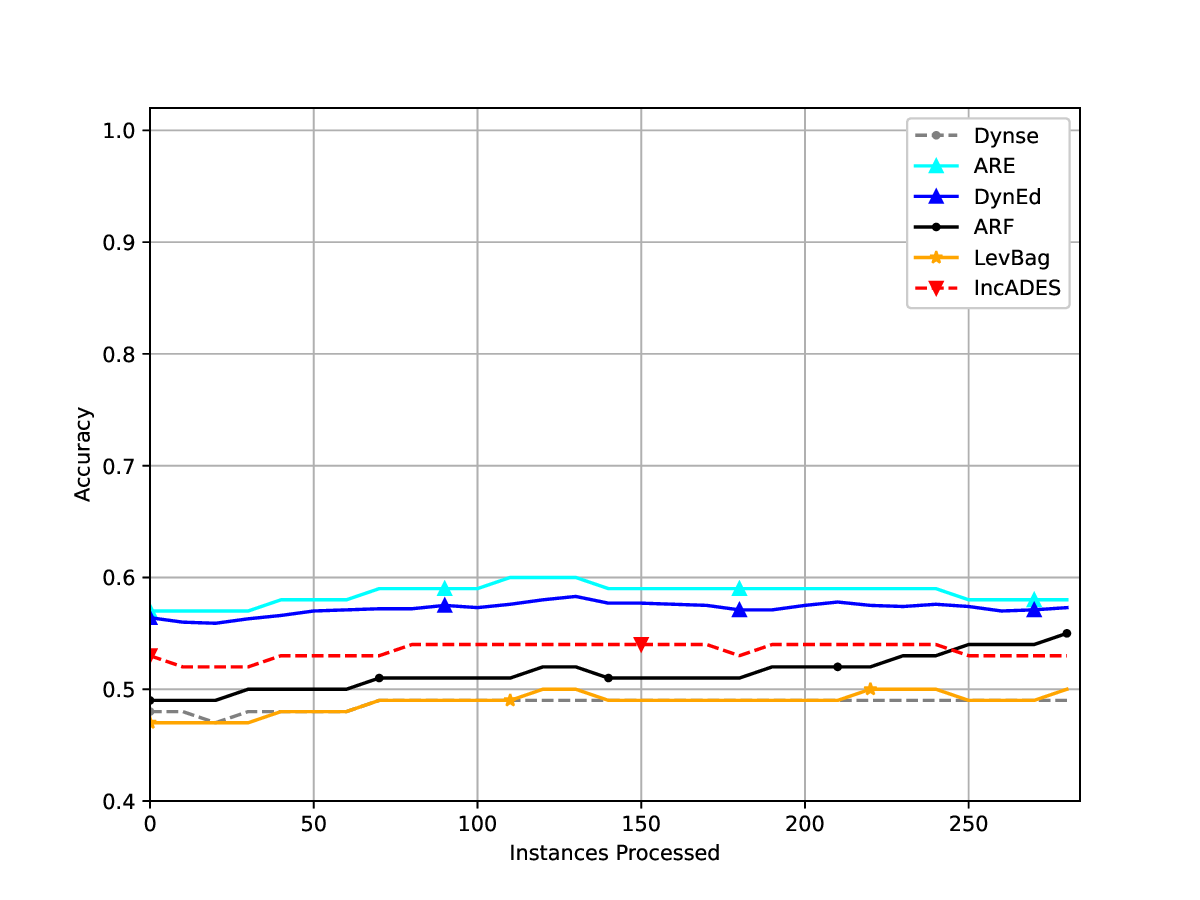}
    }\\
    \subfloat[Asfault.]{
    \label{subfig:asfault-art}
    \includegraphics[width=0.3\linewidth]{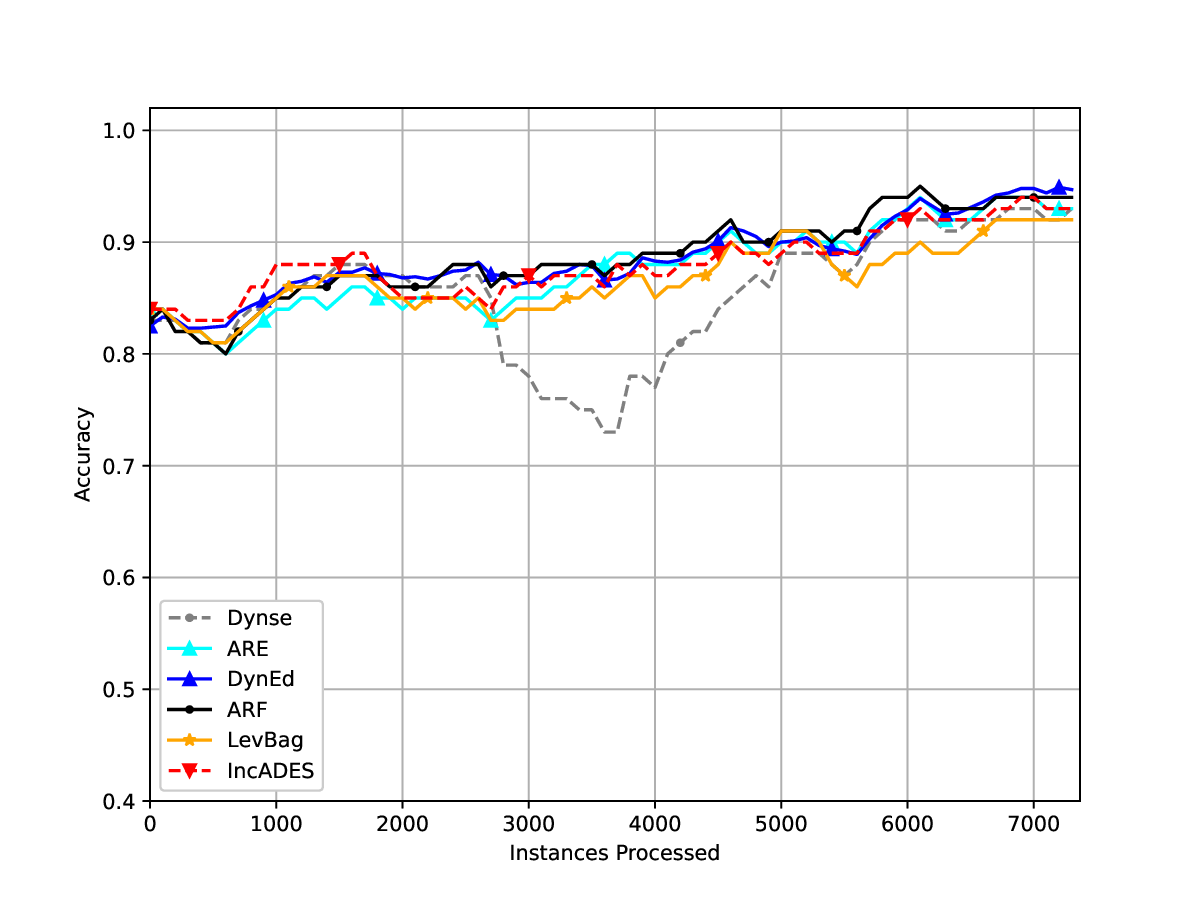}
    }
    \subfloat[Covertype.]{
    \label{subfig:covertype-art}
    \includegraphics[width=0.3\linewidth]{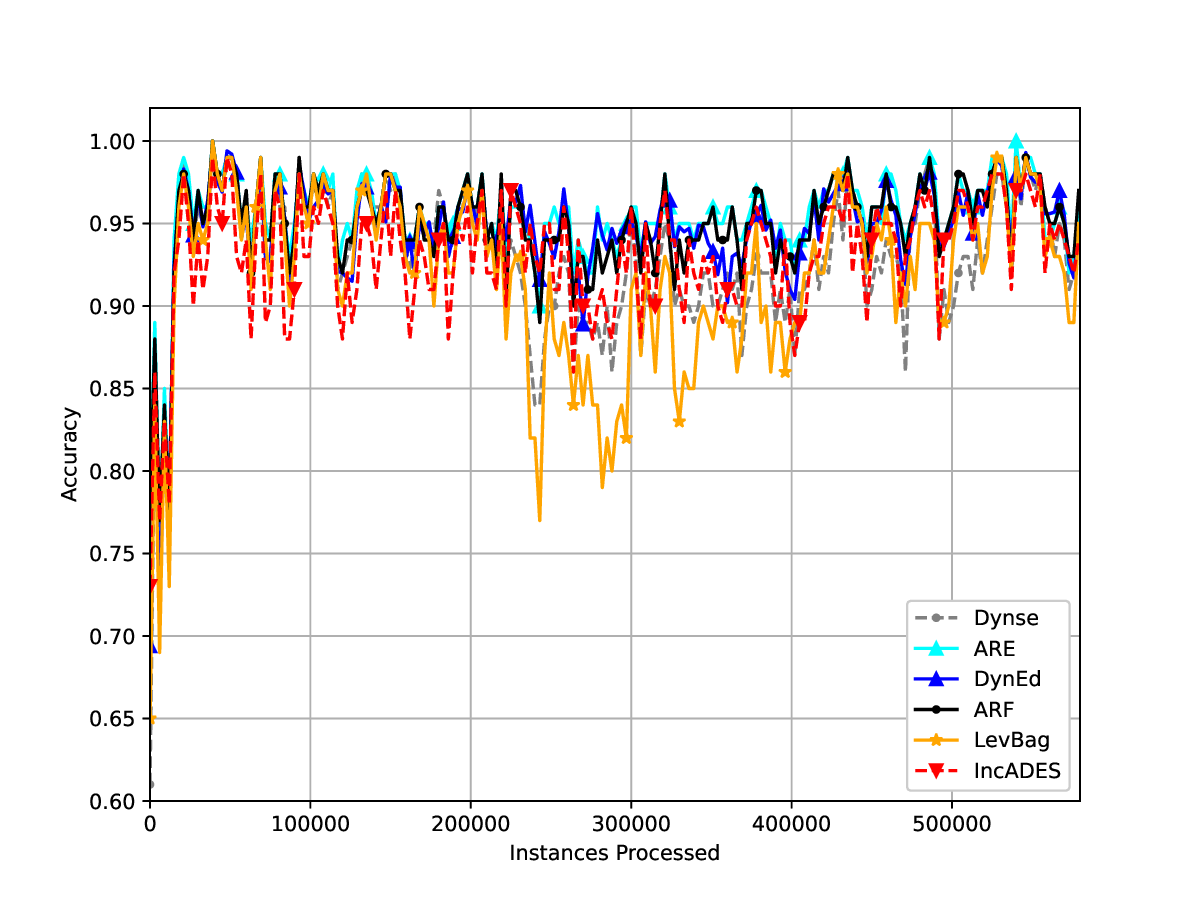}
    }
    \subfloat[Pen Digits.]{
    \label{subfig:pen-digits-art}
    \includegraphics[width=0.3\linewidth]{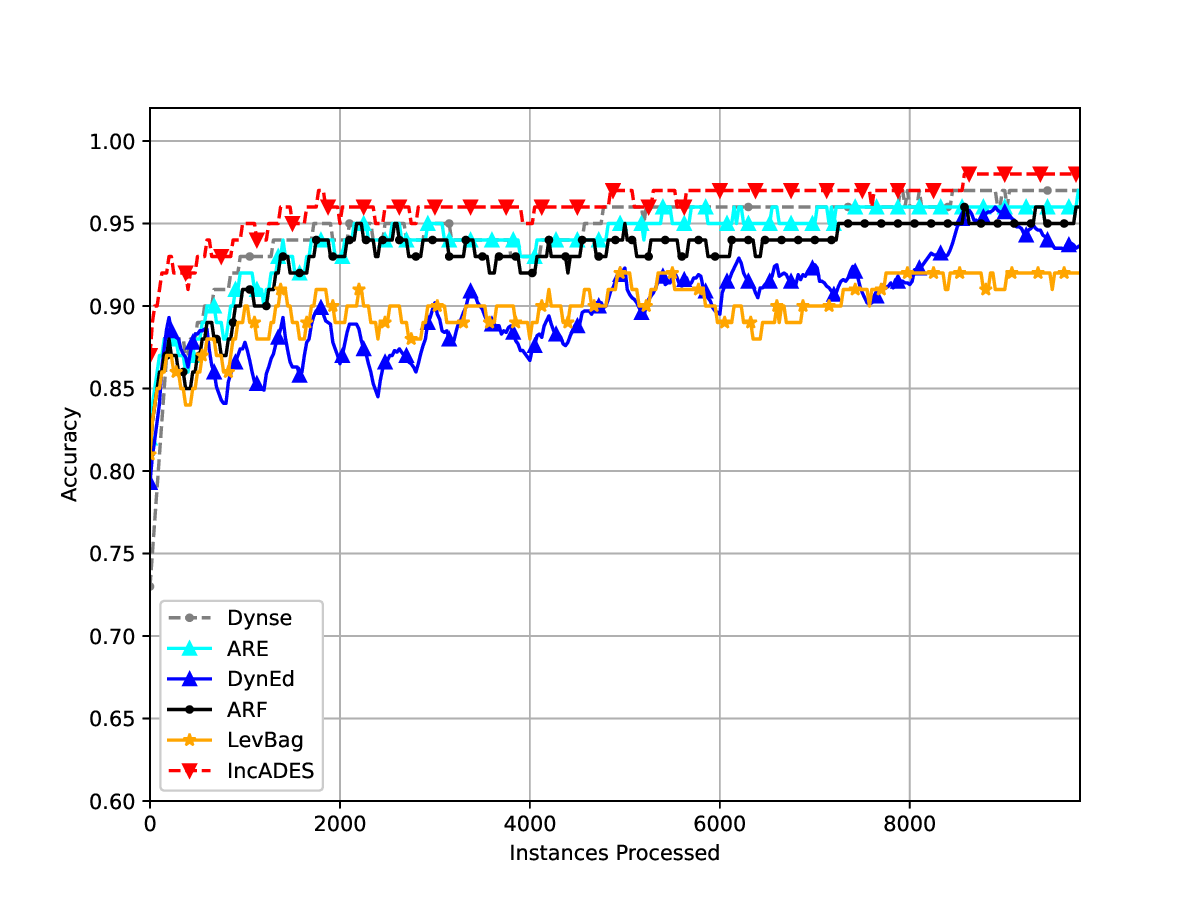}
    }\\
    \caption{Prequential accuracies of IncA-DES and other state-of-the-art methods on test-then-train.}
    \label{fig:preqAcc-art}
\end{figure}

\begin{figure}[!htb]
\centering
\ContinuedFloat
    \subfloat[DryBean.]{
    \label{subfig:dry-bean-art}
    \includegraphics[width=0.3\linewidth]{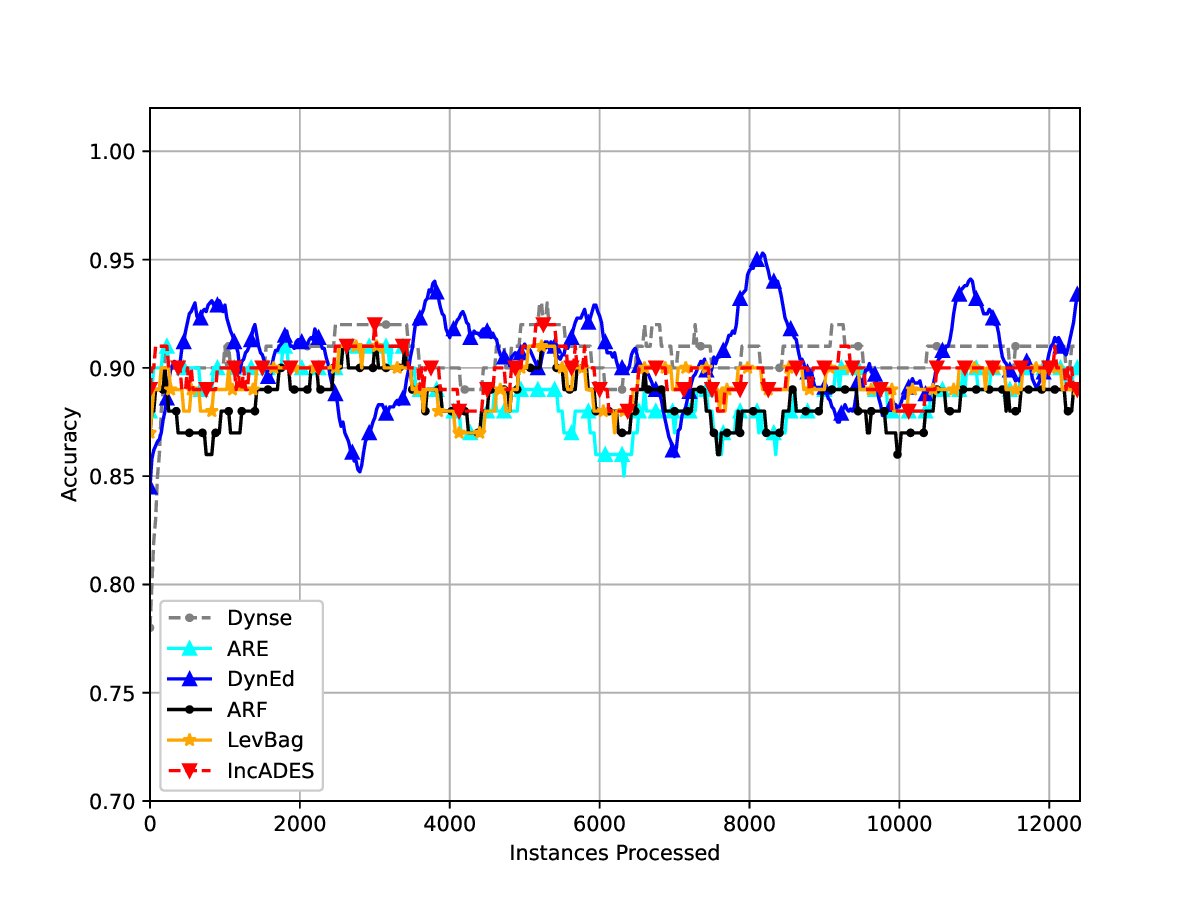}
    }
    \subfloat[Rice.]{
    \label{subfig:rice-art}
    \includegraphics[width=0.3\linewidth]{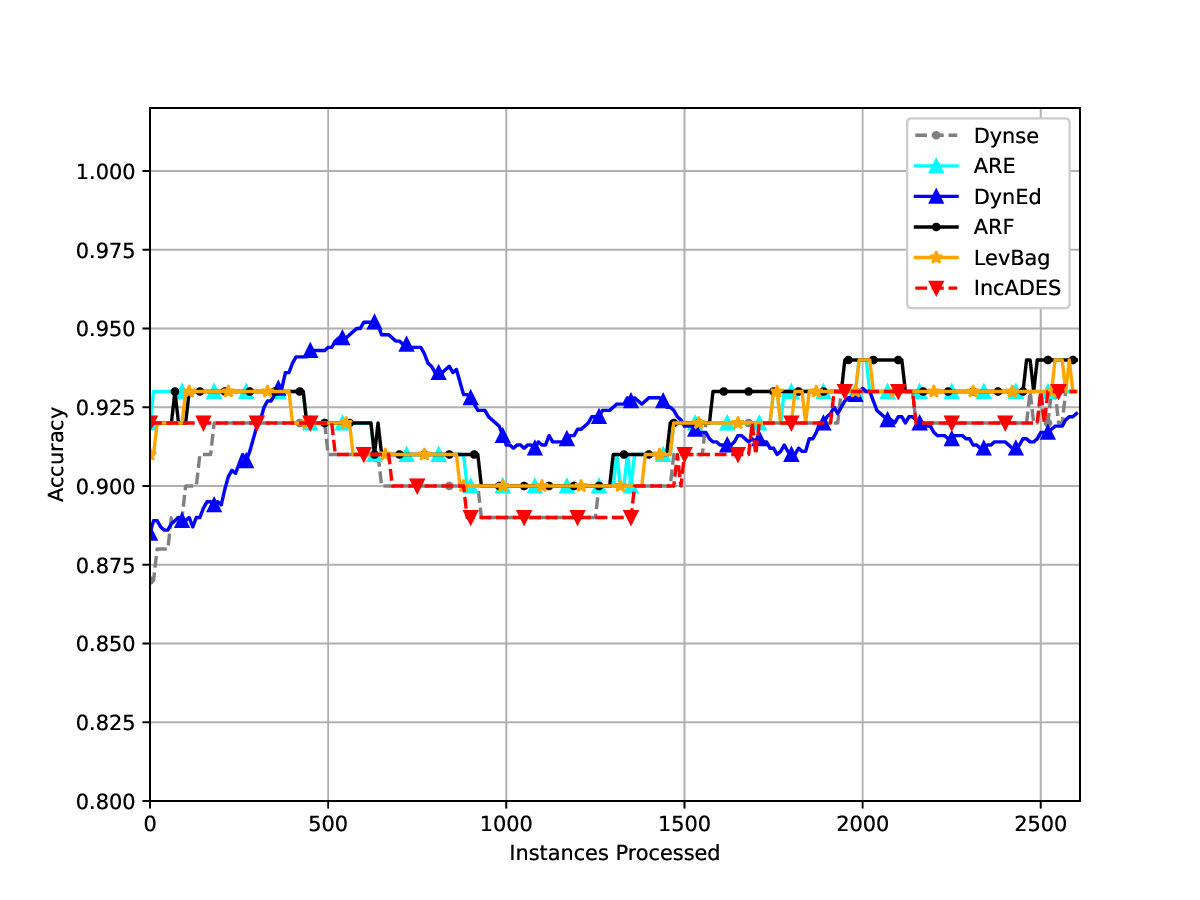}
    }
    \subfloat[Letter.]{
    \label{subfig:letter-art}
    \includegraphics[width=0.3\linewidth]{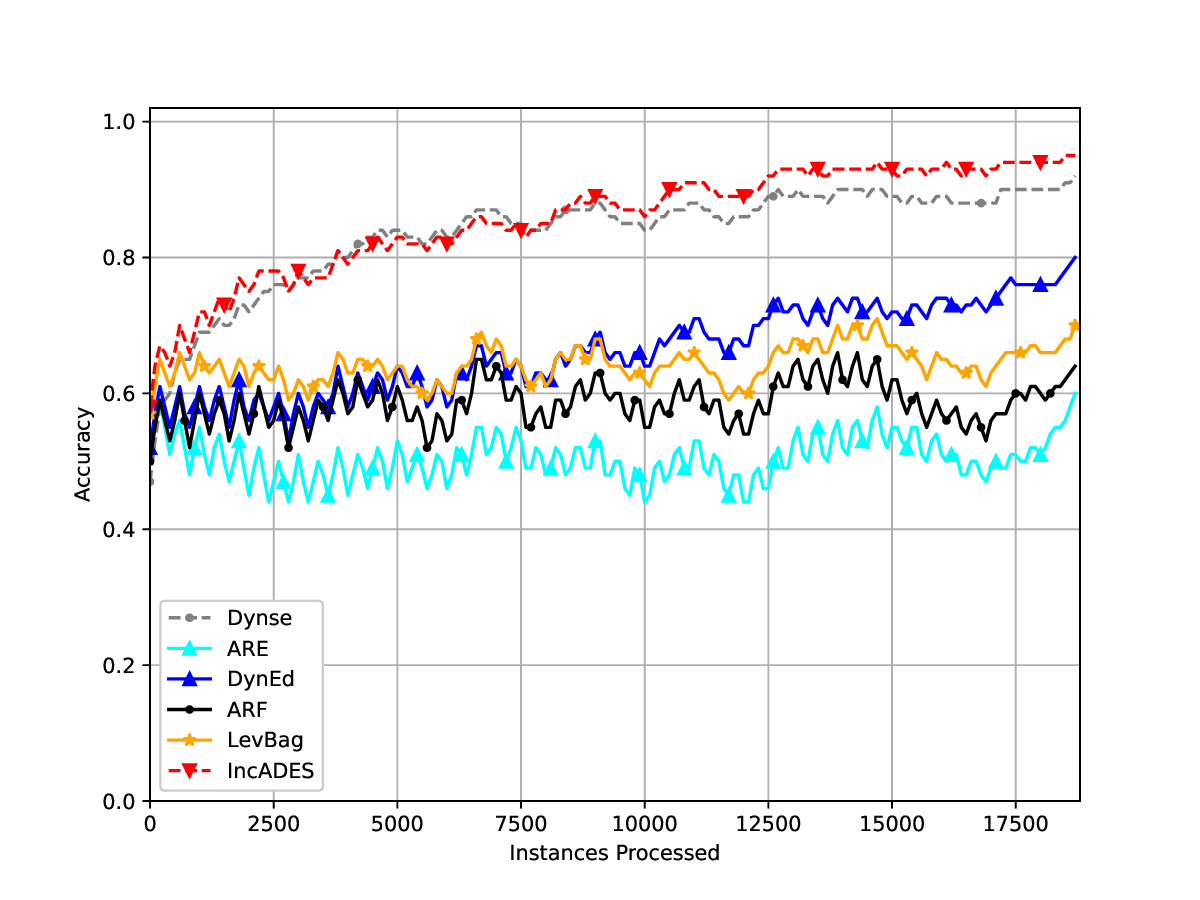}
    }\\
    \subfloat[Sine-Rec.]{
    \label{subfig:sine-rec-art}
    \includegraphics[width=0.3\linewidth]{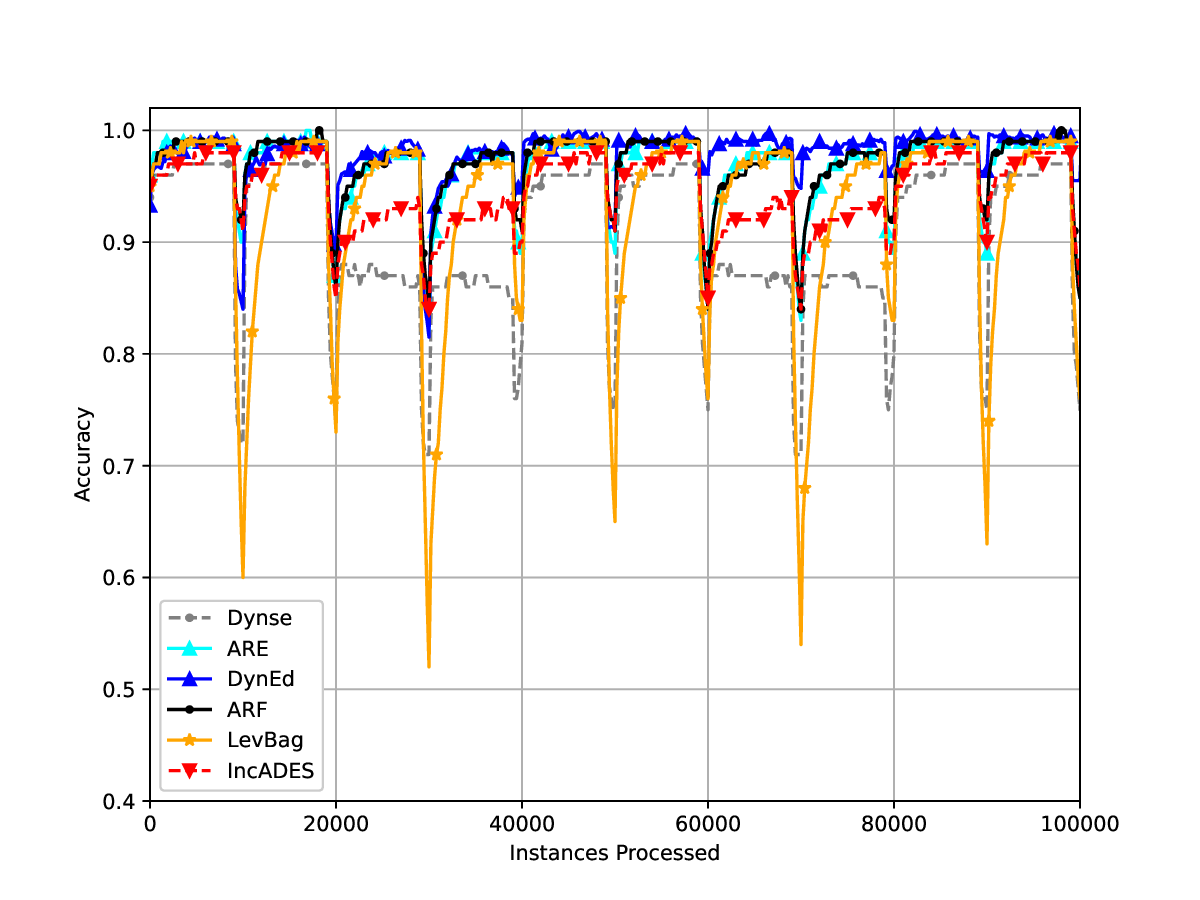}
    }
    \subfloat[SEA-Rec.]{
    \label{subfig:sea-rec-art}
    \includegraphics[width=0.3\linewidth]{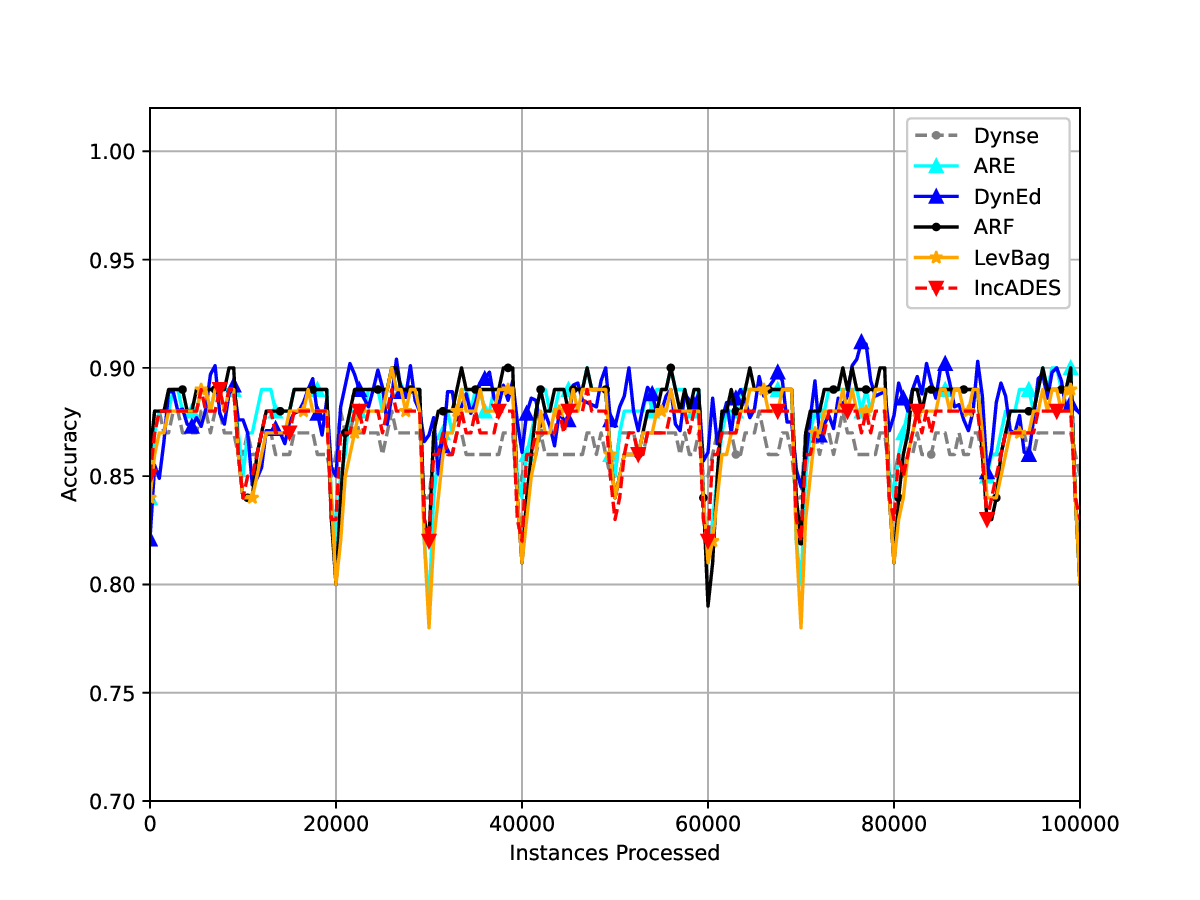}
    }
    \subfloat[STAGGER-Rec.]{
    \label{subfig:stagger-rec-art}
    \includegraphics[width=0.3\linewidth]{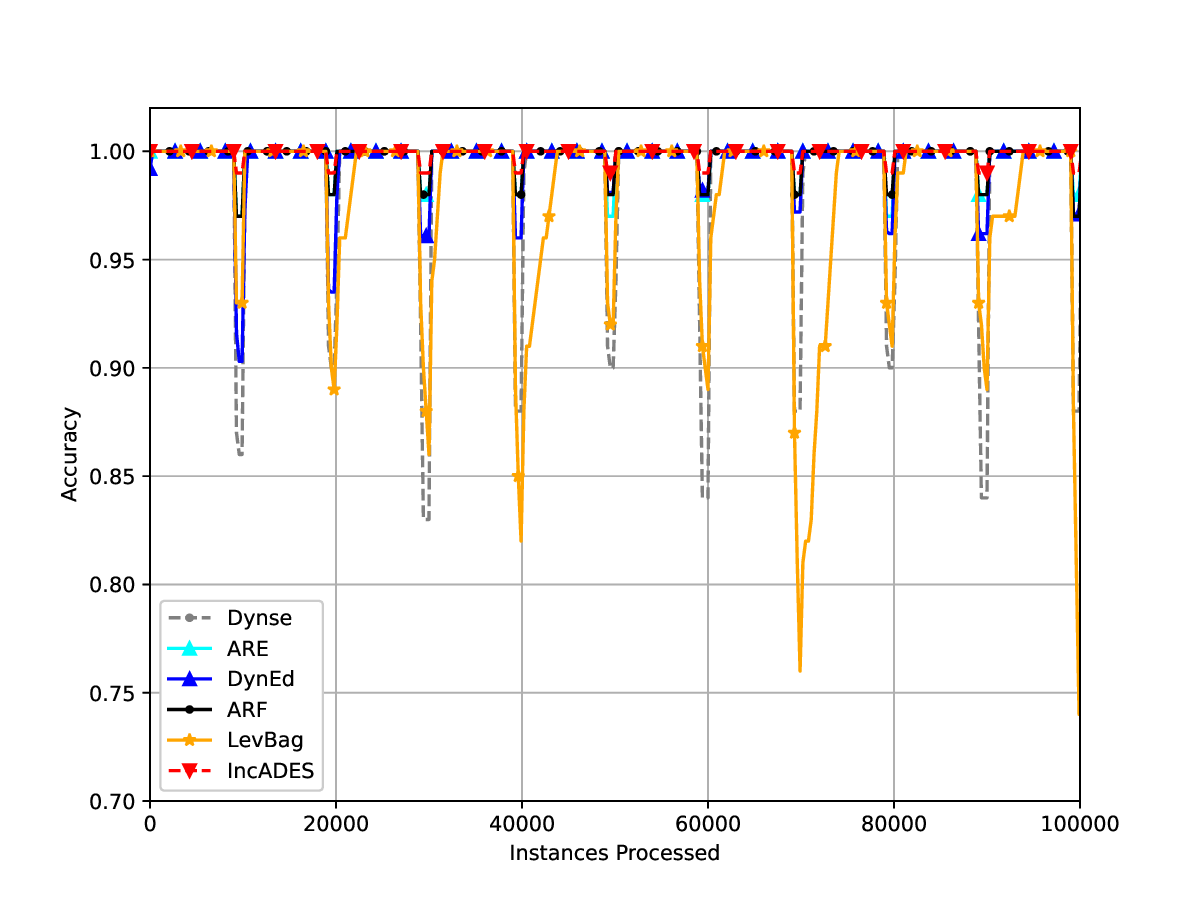}
    }\\
    \subfloat[Agrawal.]{
    \label{subfig:agrawal-art}
    \includegraphics[width=0.3\linewidth]{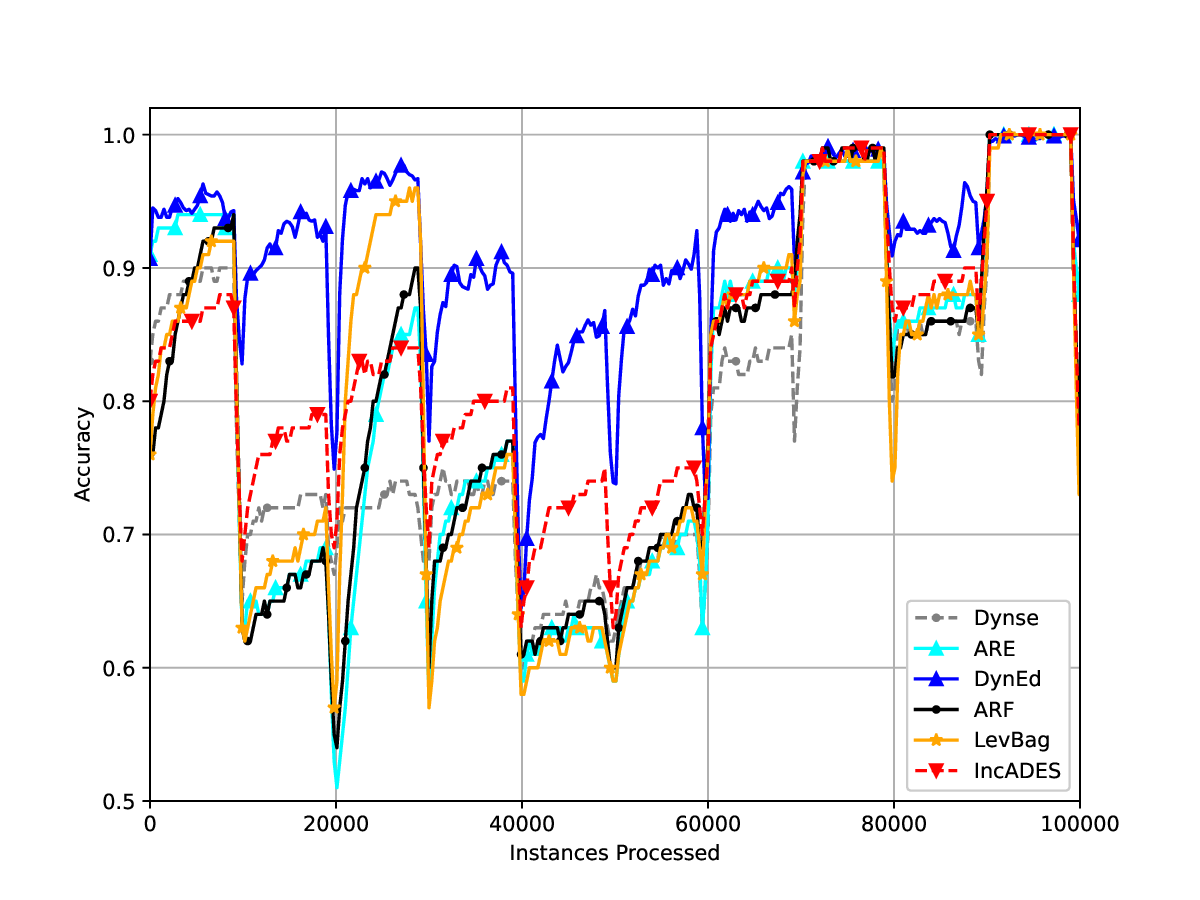}
    }
    \subfloat[Hyperplane.]{
    \label{subfig:hyperplane-art}
    \includegraphics[width=0.3\linewidth]{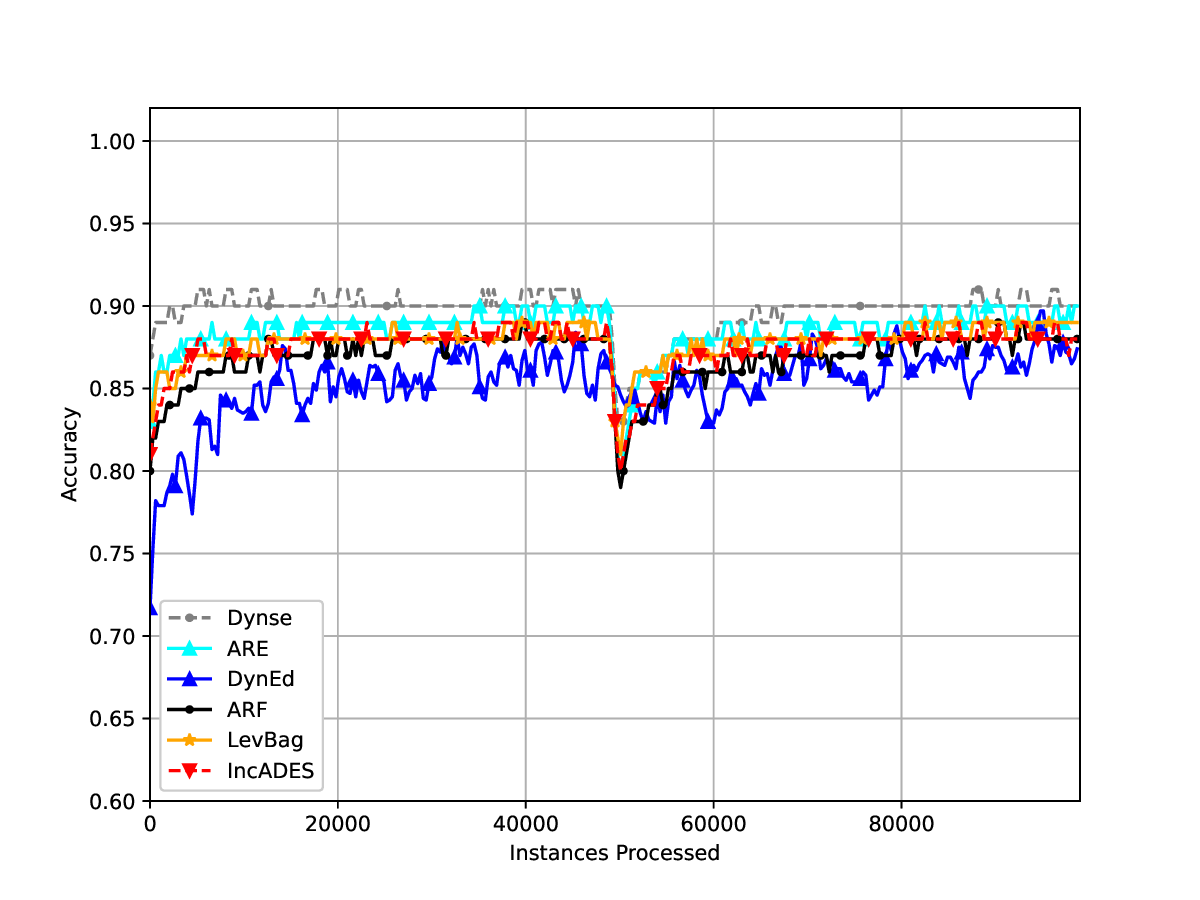}
    }
    \subfloat[LED.]{
    \label{subfig:led-art}
    \includegraphics[width=0.3\linewidth]{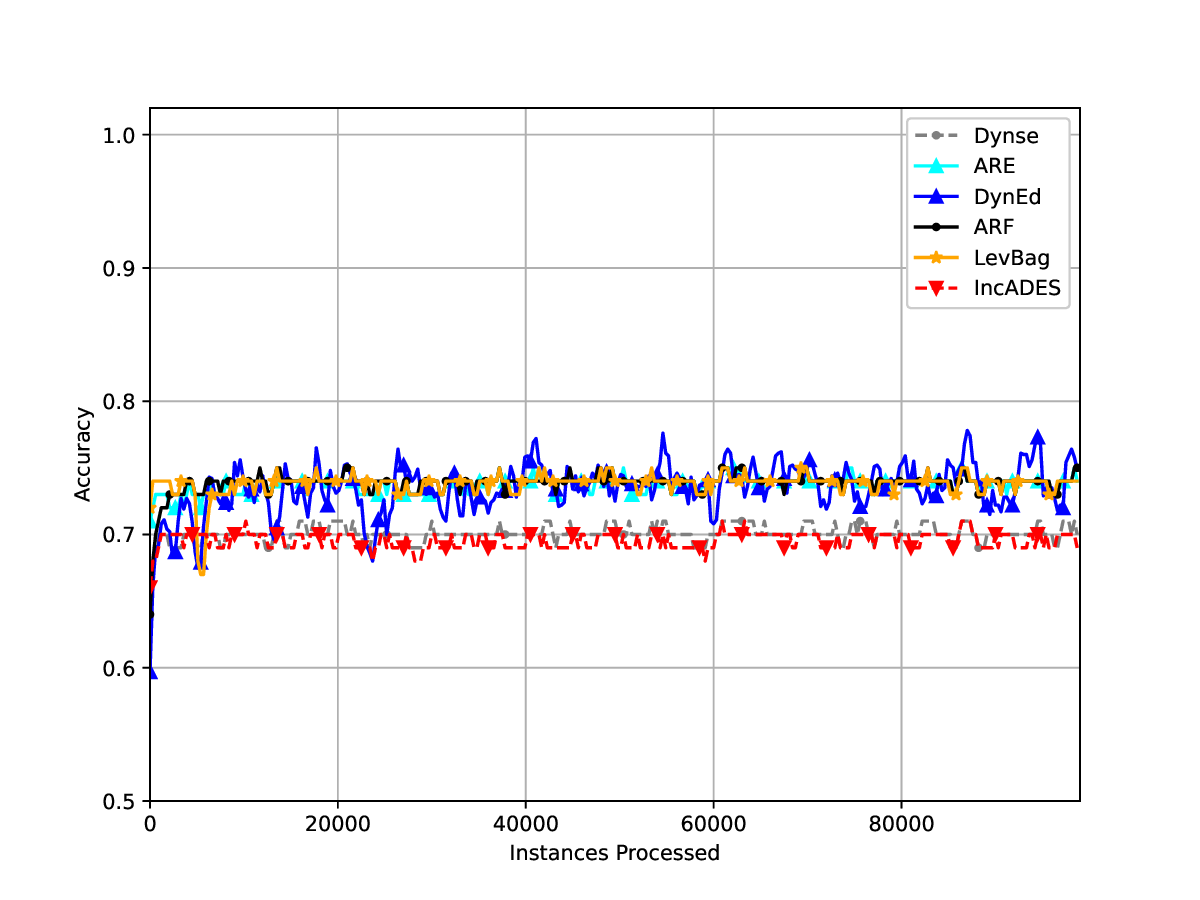}
    }\\
    \subfloat[RandomRBF.]{
    \label{subfig:randomrbf-art}
    \includegraphics[width=0.3\linewidth]{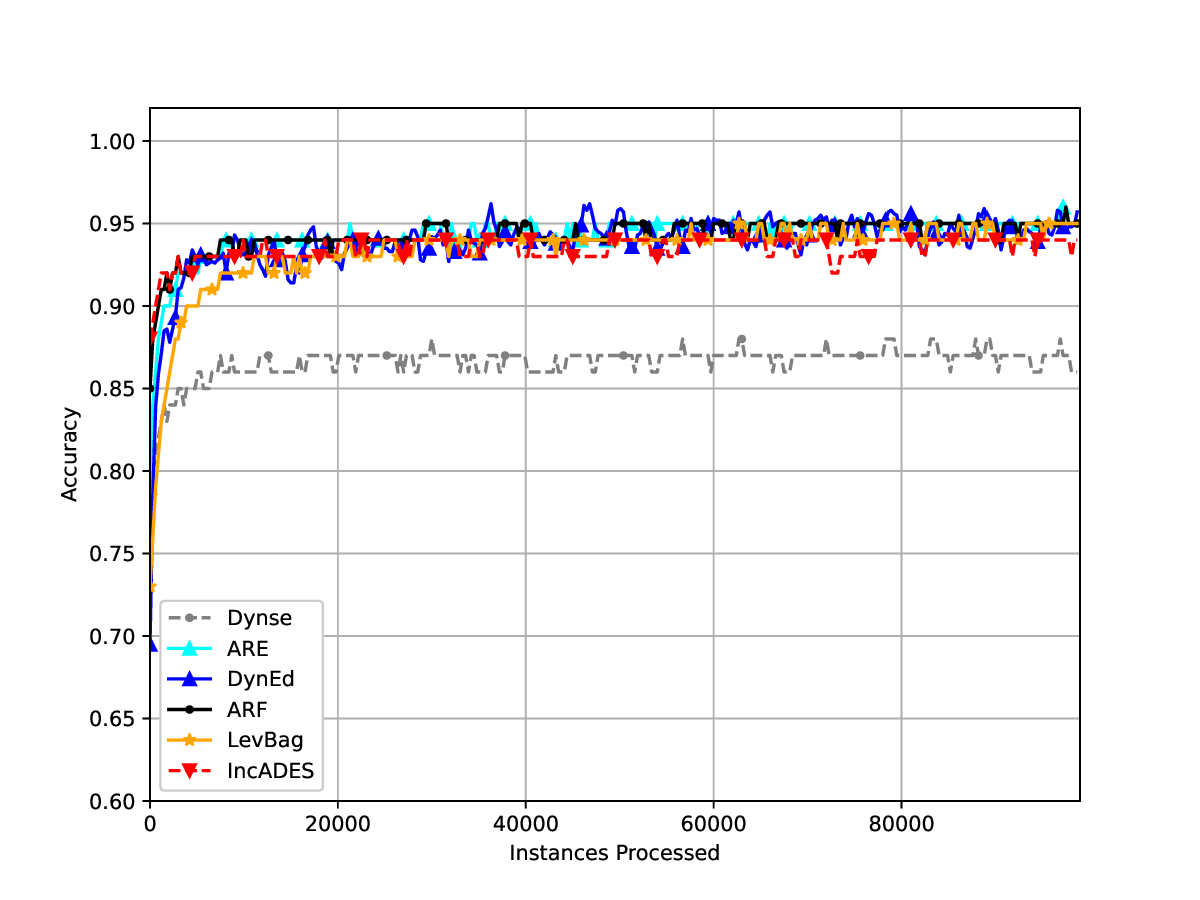}
    }
    \caption{Prequential accuracies of IncA-DES and other state-of-the-art methods on test-then-train (Continued).}
    \label{fig:preqAcc-art2}
\end{figure}

On the Rialto dataset (Figure \ref{subfig:rialto-art}), \ac{IncA-DES}, at the moments where all of the methods present drops in accuracy, was the one that struggled less and had higher accuracy on most of the moments. \ac{ARE} was competitive on this one. On the Keystroke dataset in Figure \ref{subfig:keystroke-art}, \ac{IncA-DES} is seen as the most accurate at every moment. On the Insects benchmark, \ac{IncA-DES} was competitive on the datasets with abrupt and gradual concept drifts (Figures \ref{subfig:insects-ab-art} and \ref{subfig:insects-gb-art}). However, on the Insects-IB dataset (Figure \ref{subfig:insects-ib-art}), which carries an incremental concept drift, \ac{IncA-DES} was not able to be competitive with \ac{ARE}, \ac{ARF} or DynED. \ac{ARE} presented the best prequential accuracy in every moment in the Yeast dataset (Figure \ref{subfig:yeast-art}), and many methods kept a similar accuracy in the timestamp of the Asfault dataset (Figure \ref{subfig:asfault-art}). \ac{IncA-DES} has presented some drops in accuracy on the Covertype dataset in Figure \ref{subfig:covertype-art}, probably due to the presence of categorical features, that neighborhood search can not handle well, as also suggests Dynse's performance.

Getting into the datasets with an induced \textit{virtual concept drift}, \ac{IncA-DES} was the best performing on the Pen Digits and Letters datasets, in Figures \ref{subfig:pen-digits-art} and \ref{subfig:letter-art}, presenting the best accuracy on all of the timestamps and a significative dominance on the Letters. On the Dry Bean and Rice datasets (Figures \ref{subfig:dry-bean-art} and \ref{subfig:rice-art}), DynED appears as the most accurate in most of the timestamps.

On the synthetic datasets, we see \ac{IncA-DES} struggling less on the Sine dataset (Figure \ref{subfig:sine-rec-art}) after the first concept drift. Still, on the subsequent ones, it seems that the decision boundaries are not adequately learned, presenting a smaller accuracy on the 3rd and 4th concepts. This is perceived in the remaining of the stream. On the STAGGER-Rec dataset, in Figure \ref{subfig:stagger-rec-art}, it was the one who struggled the less with the drifting concepts and did not present the drops in accuracy that the other methods did after concept drift. We see dominance from DynED on the Agrawal dataset (Figure \ref{subfig:agrawal-art}), and a greater accuracy from Dynse and \ac{ARE} on the Hyperplane dataset (Figure \ref{subfig:hyperplane-art}).

On the LED dataset (Figure \ref{subfig:led-art}, the neighborhood-based methods (\ac{IncA-DES} and Dynse) seem to struggle, also likely due to the presence of categorical features, as in the Covertype dataset. On the RandomRBF dataset (Figure \ref{subfig:randomrbf-art}), we see that \ac{IncA-DES}, \ac{ARE} and \ac{ARF} got a similar accuracy at many moments, but yet \ac{ARF} and \ac{ARE} were the best performing.

Finally, we have performed a Wilcoxon Signed-Rank Test, where we can directly compare two different methods. The p-values rounded to two decimals are in Table \ref{wilcoxon-test-then-train}. Notice that \ac{IncA-DES} only did not present a statistically significant difference to \ac{ARE}, DynED and \ac{ARF}, but still with a higher average accuracy.

\begin{table}[!ht]
    \centering
    \caption{P-values of the Wilcoxon Signed-Rank Test. P-values smaller than 0.05 are set in bold, and the ($\pm$) sign indicates whether the column-wise method has surpassed the row-wise method.}
    \label{wilcoxon-test-then-train}
    \begin{tabular}{c c c c c c c c c}
    \toprule
         & ARE & Dynse & DynEd & ARF & OzaBag & LevBag & OAUE & IncA-DES \\ \midrule
        ARE & - & \textbf{0.00 (-)} & 0.73 & 0.50 & \textbf{0.00 (-)} & \textbf{0.00 (-)} & \textbf{0.00 (-)} & 0.98 \\
        Dynse &  & - & 0.06 & \textbf{0.04(+)} & \textbf{0.00 (-)} & 0.81 & \textbf{0.02 (-)} & \textbf{0.00 (+)} \\ 
        DynED & ~ &  & - & 0.94 & \textbf{0.00 (-)} & \textbf{0.00 (-)} & \textbf{0.00 (-)} & 0.68 \\ 
        ARF & ~ & ~ &  & - & \textbf{0.00 (-)} & \textbf{0.00 (-)} & \textbf{0.00 (-)} & 0.39 \\ 
        OzaBag & ~ & ~ & ~ &  & - & \textbf{0.00 (+)} & \textbf{0.00 (+)} & \textbf{0.00 (+)} \\ 
        LevBag & ~ & ~ & ~ & ~ &  & - & \textbf{0.00 (-)} & \textbf{0.00 (+)} \\ 
        OAUE & ~ & ~ & ~ & ~ & ~ &  & - & \textbf{0.00 (+)} \\ 
        IncA-DES & ~ & ~ & ~ & ~ & ~ & ~ &  & - \\ \bottomrule
    \end{tabular}
\end{table}

In conclusion, we can say that \ac{IncA-DES} was the method with the best average accuracy, the fastest between the most accurate methods, and obtained a good predictive accuracy in most of the scenarios tested, while presenting more wins than losses compared to every tested method, even with a simpler training approach. However, it seems to struggle on some datasets, such as Ozone, Insects-IB, Covertype, Sine, and LED, where lower accuracy was perceived in many moments of the timestamps. On the Sine dataset, the struggle is perceived mainly on the 3rd and 4th concepts. It also seems to struggle under slow changing or long stable concepts, as suggests the performance in the Insects-IB dataset. The adopted training strategy may also harm performance in some scenarios, although it can lead to better performance in others. If the data is sent in a meaningful and natural order over time, the classifiers can cover each a different local region, favoring the usage of local-based \ac{DS}. Otherwise, the performance can be compromised.


\subsubsection{Experiments With Delayed and Partial Labels}

\begin{figure}
    \centering
    \includegraphics[width=1\linewidth]{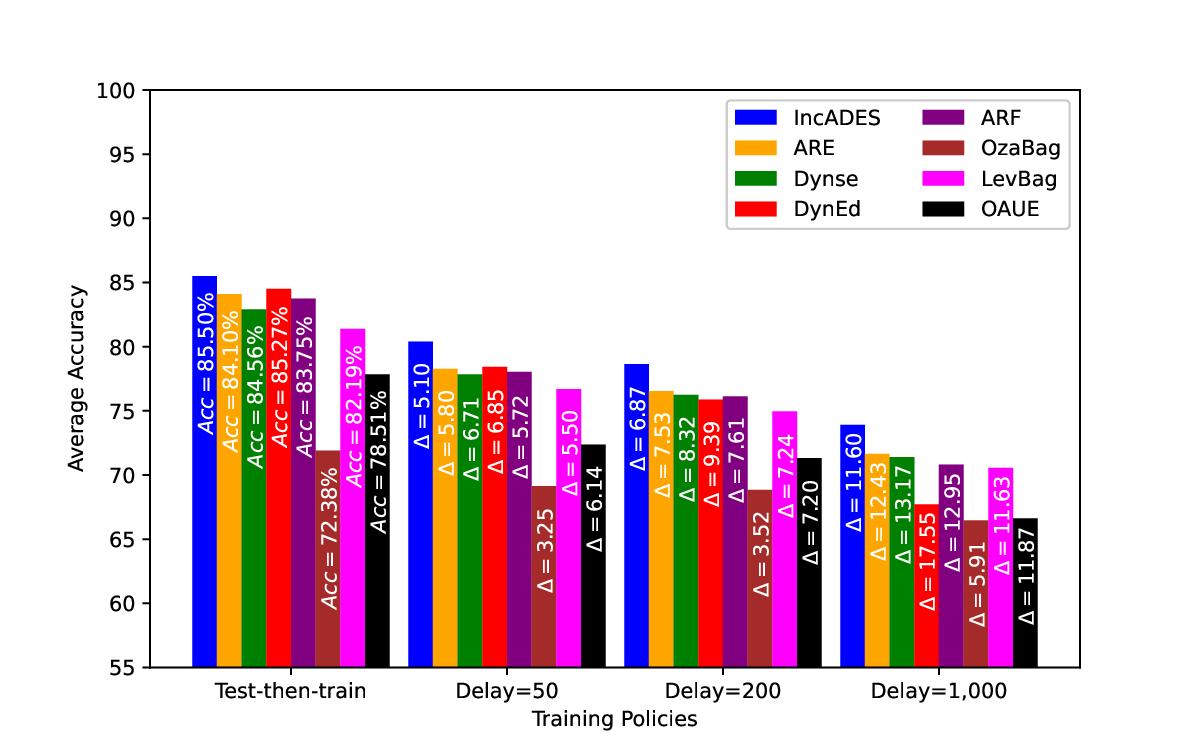}
    \caption{Variations of accuracy for each value of delay of labels. The average accuracy of the test-then-train policy is shown, and the variation $\Delta$ compared to the test-then-train is shown for the other training policies. Notice that \ac{IncA-DES} always had the highest average accuracy for every training policy.}
    \label{fig:var-delay}
\end{figure}

Now, let us analyze the results with a limited label availability. In Figure \ref{fig:var-delay}, we show the accuracies of the test-then-train policy and the variations of each method with each delay. \ac{IncA-DES} still got the highest average accuracy in all of the scenarios. Notice that out of the most accurate methods, it was the one that had the smallest variation in accuracy as the availability of the labels became limited. Although OzaBag presented the smaller variation, its accuracy performance was always poor compared to the other methods. Thus, the proposed framework presented resilience with less access to training data when compared to other methods. We hypothesize that the adopted training strategy of \ac{IncA-DES} helps in scenarios with limited labeled data since the training is focused on one classifier.

Now, we present the wilcoxon signed-rank tests with different delays in Table \ref{wilcoxon-delay}. See that for delays equal to 50 and 200, \ac{IncA-DES} only did not have a statistically significant difference to \ac{ARE}, DynED and \ac{ARF}. When the delay was equal to 1,000, \ac{ARE} has presented statistically significant difference to five methods, while \ac{IncA-DES} to two. \ac{ARE} seems not to lose much performance os most of the datasets when label is limited, while \ac{IncA-DES} presented some subtantial loss in some cases (see Tables \ref{tab:acc-art-delay50}, \ref{tab:acc-art-delay200}, and \ref{tab:acc-art-delay1000} in Appendix \ref{app:acc-delay}), making the difference to other methods not be significant according to the Wilcoxon test.

\begin{table}[!ht]
    \centering
    \caption{P-values of the Wilcoxon Signed-Rank Test with delayed and partial labeling. P-values minor than 0.05 are set in bold, and the ($\pm$) sign indicates whether the column-wise method has surpassed the row-wise method.}
    \label{wilcoxon-delay}
    \begin{tabular}{c c c c c c c c c}
    \toprule
        & & & & \textit{Delay = 50}\\
         & ARE & Dynse & DynEd & ARF & OzaBag & LevBag & OAUE & IncADES \\ \midrule
        ARE & - & \textbf{0.04 (-)} & 0.34 & 0.35 & \textbf{0.00 (-)} & \textbf{0.01 (-)} & \textbf{0.00 (-)} & 0.24 \\ 
        Dynse & ~ & - & 0.39 & 0.34 & \textbf{0.00 (-)} & 0.31 & \textbf{0.00 (-)} & \textbf{0.00 (+)} \\ 
        DynEd & ~ & ~ & - & 0.96 & \textbf{0.01 (-)} & 0.22 & \textbf{0.00 (-)} & 0.27 \\ 
        ARF & ~ & ~ & ~ & - & 0.01 & 0.08 & \textbf{0.01 (-)} & 0.28 \\ 
        OzaBag & ~ & ~ & ~ & ~ & - & \textbf{0.00 (+)} & 0.23 & \textbf{0.00 (+)} \\ 
        LevBag & ~ & ~ & ~ & ~ & ~ & - & \textbf{0.00 (-)} & \textbf{0.00 (+)} \\ 
        OAUE & ~ & ~ & ~ & ~ & ~ & ~ & - & \textbf{0.00 (+)} \\ 
        IncADES & ~ & ~ & ~ & ~ & ~ & ~ & ~ & - \\ \hline
        
        & & & & \textit{Delay = 200}\\
         & ARE & Dynse & DynEd & ARF & OzaBag & LevBag & OAUE & IncADES \\ \midrule
        ARE & - & \textbf{0.02 (-)} & 0.19 & 0.36 & \textbf{0.01 (-)} & \textbf{0.00 (-)} & \textbf{0.01 (-)} & 0.50 \\ 
        Dynse & ~ & - & 0.57 & 0.32 & \textbf{0.00 (-)} & 0.35 & \textbf{0.00 (-)} & \textbf{0.01 (+)} \\ 
        DynEd & ~ & ~ & - & 0.57 & 0.05 & 0.39 & \textbf{0.01 (-)} & 0.31 \\ 
        ARF & ~ & ~ & ~ & - & \textbf{0.02 (-)} & 0.1 & \textbf{0.03 (-)} & 0.25 \\ 
        OzaBag & ~ & ~ & ~ & ~ & - & \textbf{0.03 (+)} & 0.28 & \textbf{0.00 (+)} \\ 
        LevBag & ~ & ~ & ~ & ~ & ~ & - & \textbf{0.01 (-)} & \textbf{0.00 (+)} \\ 
        OAUE & ~ & ~ & ~ & ~ & ~ & ~ & - & \textbf{0.00 (+)} \\ 
        IncADES & ~ & ~ & ~ & ~ & ~ & ~ & ~ & - \\ \hline

        & & & & \textit{Delay = 1,000}\\
         & ARE & Dynse & DynEd & ARF & OzaBag & LevBag & OAUE & IncADES \\ \midrule
        ARE & - & \textbf{0.04 (-)} & \textbf{0.04 (-)} & 0.24 & \textbf{0.03 (-)} & \textbf{0.04 (-)} & \textbf{0.01 (-)} & 0.76 \\ 
        Dynse & ~ & - & 0.32 & 0.64 & 0.06 & 0.35 & \textbf{0.01 (-)} & 0.11 \\ 
        DynEd & ~ & ~ & - & 0.07 & 0.73 & 0.26 & 0.4 & 0.09 \\ 
        ARF & ~ & ~ & ~ & - & 0.16 & 0.34 & \textbf{0.05 (-)} & 0.24 \\ 
        OzaBag & ~ & ~ & ~ & ~ & - & 0.15 & 0.64 & \textbf{0.02 (+)} \\ 
        LevBag & ~ & ~ & ~ & ~ & ~ & - & \textbf{0.02 (-)} & 0.05 \\ 
        OAUE & ~ & ~ & ~ & ~ & ~ & ~ & - & \textbf{0.01 (+)} \\ 
        IncADES & ~ & ~ & ~ & ~ & ~ & ~ & ~ & - \\ \bottomrule
    \end{tabular}
\end{table}

In conclusion, limiting the label availability has interfered with the results of all the tested state-of-the-art methods, and some struggled more than others. DynED had the highest variation in accuracy with a high delay, with a difference of 17.55 percentage points from the delay of 1,000 instances to the test-then-train policy.

Considering \ac{IncA-DES}, \ac{ARE}, \ac{ARF}, and DynED, the most accurate methods in the test-then-train policy, \ac{IncA-DES} had the smallest difference in accuracy on every delay, thus being the most resilient under limited training data considering average accuracy. However, \ac{ARE} was the one that got a significant difference from most of the methods with the longest label delay (1,000).


\subsection{Ablation Study: Evaluating Components of IncA-DES}

Now that we have compared the proposed framework to the state-of-the-art methods let us assess how the proposed Online K-d tree algorithm and the overlap-based classification influence both accuracy performance and processing time of \ac{IncA-DES}. For these experiments, the datasets were selected to cover different dimensionality scenarios: Electricity, NOAA, Adult, Gas Sensor, Covertype, Insects-AB, and Rialto. In these tests, the fact that the K-d tree is rebuilt sometimes influences the processing time. The $\omega$ parameter has been set to 1 in order to use the classification filter with no overlap only.

In Table \ref{tab:acc-ablation-study}, we first present the average accuracy, processing time, and instances per second of a baseline version of \ac{IncA-DES}, which uses the brute force \ac{kNN} for defining the \ac{RoC} and does not consider an overlap-based classification. The subsequent columns expose the results when different components are added: the overlap-based classification, the Online K-d tree, and, finally, both components combined.

\begin{table}[htb]
    \centering
    \caption{Average accuracy, processing time and instances per second of IncA-DES with a baseline configuration (brute force kNN without overlap-based classification) comparing with the addition of the components separately (K-d Tree and overlap-based classification).}
    \label{tab:acc-ablation-study}
    \scriptsize
    \setlength{\tabcolsep}{4pt}
    \begin{tabular}{l | c c c | c c c | c c c | c c c }
    \toprule
        Dataset & \multicolumn{3}{c|}{kNN No Overlap}  & \multicolumn{3}{c|}{kNN Overlap} & \multicolumn{3}{c|}{K-d Tree No Overlap} & \multicolumn{3}{c}{K-d Tree Overlap} \\
        & Accuracy & Time (s) & I/s & Accuracy & Time (s) & I/s & Accuracy & Time (s) & I/s & Accuracy & Time (s) & I/s \\
        \midrule
        Electricity & 88.85 & 18.37 & 2456 & \textbf{88.89} & 8.76 & 5,150 & \textbf{88.88} & 17.49 & 2,579 & 88.65 & \textbf{8.32} & \textbf{5,422} \\
        NOAA & 75.11 & 8.79 & 2,043 & \textbf{75.32} & 6.47 & 2,776 & 75.30 & 8.40 & 2,138 & 75.18 & \textbf{6.43} & \textbf{2,793} \\
        Adult & 83.11 & 318.08 & 142 & \textbf{83.31} & 241.05 & 187 & 83.06 & 341.89 & 132 & 83.25 & \textbf{192.09} & \textbf{234} \\
        Gas Sensor & 94.44 & 83.72 & 164 & 96.68 & \textbf{15.03} & \textbf{912} & 94.40 & 94.09 & 146 & \textbf{96.70} & 24.56 & 558 \\
        Rialto & 80.62 & 693.36 & 118 & \textbf{82.35} & 473.53 & 173 & 79.90 & 386.87 & 212 & 81.66 & \textbf{193.57} & \textbf{424} \\
        Insects-AB & 74.46 & 279.3 & 188 & 75.07 & \textbf{171.01} & \textbf{308} & 74.43 & 505.67 & 104 & \textbf{75.12} & 287.64 & 183 \\
        Covertype & 93.59 & 2,719.04 & 214 & \textbf{93.68} & 1,512.46 & 384 & 93.43 & 2,233.85 & 260 & 93.44 & \textbf{1,046.72} & \textbf{555} \\\hline
        Average & 84.31 & 588.67 & 760 & \textbf{85.04} & 346.90 & 1,413 & 84.20 & 512.61 & 796 & 84.86 & \textbf{251.33} & \textbf{1,453} \\
        \bottomrule
    \end{tabular}
\end{table}

We can see that the framework got more benefits with the overlap-based classification, which improved accuracy and processing time in all tested datasets. Regarding the K-d tree, we can see a drop in average accuracy but an improvement in processing time on some datasets. Its benefits seem more problem-specific, a constraint already known for K-d trees. The processing time may be harmed if the algorithm can not effectively split the subtrees from the search space. This can happen due to, for instance, a data distribution that does not favor the partition through the Canberra Distance.

In conclusion, we can say that the Online K-d tree sped up the \ac{RoC} definition on some datasets. However, due to the non-optimal \ac{RoC} given by the K-d tree, the accuracy performance decreased but at a negligible level. The overlap-based classification improved both accuracy and processing time. Therefore, we recommend that authors think of using this type of classification filter when dealing with \ac{DS} methods based on \ac{RoC} and choose carefully when to use the Online K-d tree algorithm over the brute force \ac{kNN}. If one is more concerned with the accuracy performance, it may prefer to rely on the brute force \ac{kNN}. However, in some cases, the Online K-d tree algorithm can decrease the processing time substantially without much impact on accuracy.

\subsection{Case Study: Varying the Search Space -- Size of DSEW}

In this section, we extend the experiments for the Online K-d tree algorithm by performing tests varying the size of the \ac{DSEW} in the \ac{IncA-DES} framework. The Sine, SEA, and Agrawal datasets were considered for these experiments and only instances of the first concept of these datasets were generated. For each value of $n$, \ac{IncA-DES} classified 100,000 instances. The idea is to compare the difference in performance and processing time as the search space $n$ increases. In these experiments, the K-d tree is not rebuilt. Thus, the analysis is focused on classification. In Table \ref{tab:knn-vs-kdtree-synth}, we show the accuracies, processing time, and how many instances each approach can classify per second.

\begin{table}[htb]
    \centering
    \caption{Accuracies, processing time, and instances per second of brute force kNN and our proposed Online K-d tree algorithm. $n$ stands for the \ac{DSEW}'s size. On the I/s column, values between parenthesis refer to how many times more instances the Online K-d tree could label compared to the brute force kNN in one second. Each search space was used to label 100,000 instances with IncA-DES.}
    \label{tab:knn-vs-kdtree-synth}
    \small
    \begin{tabular}{l | c c c | c c c }
    \toprule
        Dataset & \multicolumn{3}{c|}{kNN}  & \multicolumn{3}{c}{K-d tree} \\ 
         & Accuracy & Time (s) & I/s & Accuracy & Time (s) & I/s \\ 
         \midrule
        \textit{$n$=1,000} & & & & \\
        Sine & 95.94 & 5.51 & 18,149 & \textbf{95.98} & \textbf{3.31} & \textbf{30,211} (1.66x) \\ 
        SEA & 87.27 & 5.05 & 19,802 & \textbf{87.36} & \textbf{1.52} & \textbf{65,789} (3.32x)\\ 
        Agrawal & \textbf{85.56} & \textbf{8.75} & \textbf{11,429} & 85.49 & 14.59 & 6,854 (0.60x) \\ 
        Average & 89.59 & \textbf{6.44} & 16,460 & \textbf{89.61} & 6.47 & \textbf{34,285} (2.08x) \\ \hline
        \textit{$n$=10,000} &  & & & \\
        Sine & \textbf{98.17} & 50.94 & 1,963 & 98.15 & \textbf{6.30} & \textbf{15,873} (8.09x) \\ 
        SEA & \textbf{88.42} & 48.93 & 2,044 & 88.40 & \textbf{5.87} & \textbf{17,036} (8.33x) \\ 
        Agrawal & \textbf{91.17} & \textbf{87.48} & \textbf{1,143} & 90.95 & 126.81 & 789 (0.69x) \\ 
        Average & \textbf{92.59} & 62.45 & 1,717 & 92.50 & \textbf{46.33} & \textbf{11,233} (6.54x) \\ \hline
        \textit{$n$=25,000} &  & & & \\
        Sine & \textbf{98.64} & 134.98 & 741 & 98.58 & \textbf{9.47} & \textbf{10,560} (14.25x) \\ 
        SEA & \textbf{88.62} & 127.36 & 785 & 88.58 & \textbf{8.56} & \textbf{11,682} (14.88x) \\ 
        Agrawal & \textbf{92.01} & \textbf{244.94} & \textbf{408} & 91.71 & 341.89 & 292 (0.72x) \\ 
        Average & \textbf{93.09} & 169.09 & 645 & 92.96 & \textbf{119.97} & \textbf{7,511} (11.64x) \\ \hline
        \textit{$n$=50,000} & & & & \\
        Sine & \textbf{98.77} & 275.08 & 363 & 98.75 & \textbf{8.95} & \textbf{11,173} (30.78x) \\ 
        SEA & \textbf{88.82} & 240.84 & 415 & 88.79 & \textbf{8.20} & \textbf{12,195} (29.39x) \\ 
        Agrawal & \textbf{92.23} & \textbf{526.91} & \textbf{190} & 91.90 & 589.17 & 170 (0.89x) \\ 
        Average & \textbf{93.27} & 347.61 & 323 & 93.15 & \textbf{202.11} & \textbf{7,846} (24.29x) \\
        \bottomrule
    \end{tabular}
\end{table}

On average, the Online K-d tree had the smallest processing time and the highest number of classified instances per second. Regarding performance, except for when $n=1,000$, the brute force \ac{kNN} consistently achieved the highest accuracy, but the difference to the K-d tree is negligible. Although \ac{IncA-DES} did not achieve a smaller processing time on the Agrawal dataset, the difference got smaller as the search space increased. The difference between the Sine and SEA datasets is quite significant. See that \ac{IncA-DES} managed to label more than 24 times the number of instances that the brute force \ac{kNN} could label in one second when $n=50,000$.

\begin{figure}[!htb]
\centering
    \subfloat[Sine.]{
    \label{subfig:sine-knn}
    \includegraphics[width=0.3\linewidth]{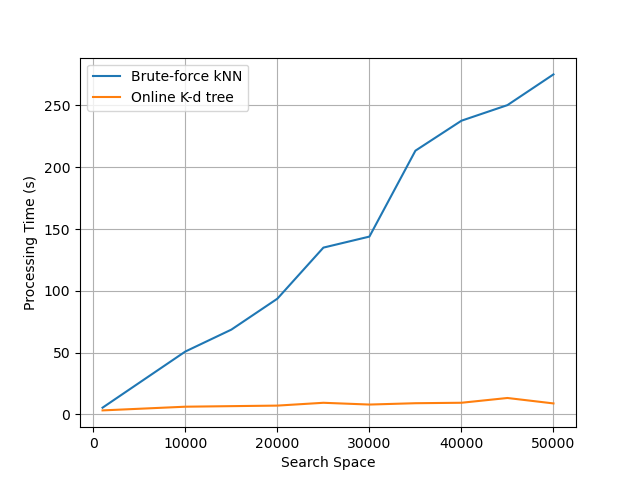}
    }
    \subfloat[SEA.]{
    \label{subfig:sea-knn}
    \includegraphics[width=0.3\linewidth]{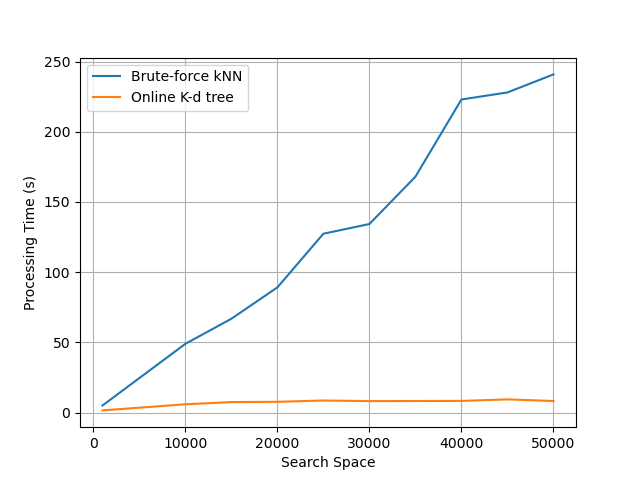}
    }
    \subfloat[Agrawal.]{
    \label{subfig:agrawal-knn}
    \includegraphics[width=0.3\linewidth]{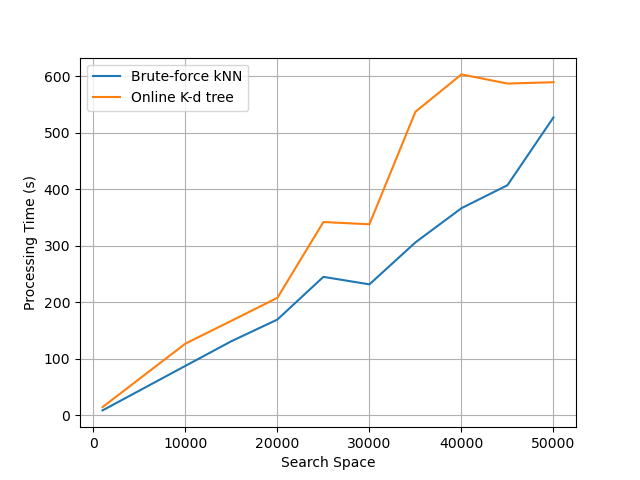}
    }\\
    \caption{Variation of the processing time from IncA-DES with the variation of the search space for the brute force kNN and the Online K-d tree.}
    \label{fig:knn-kdtree-incades}
\end{figure}

In Figure \ref{fig:knn-kdtree-incades}, we can see the variation in processing time as the search space increases. As seen previously, the Online K-d tree does not suffer on the Sine and SEA datasets, while on the Agrawal dataset, its processing time increases as $n$ grows. Possible explanations are that the data distribution or characteristics may not allow effective search space partitioning. If the number of distance calculations is close to the brute force \ac{kNN}, the overhead required by the K-d tree (e.g., construction and traversal) leads to increased processing time.

In conclusion, the proposed Online K-d tree algorithm presented quicker neighborhood definition compared to the brute force \ac{kNN}, mostly when the search space increases. However, its benefits are not guaranteed in every scenario, a constraint prior known from the K-d tree. Although the neighborhood definition can be sped up, the cost of maintaining a K-d tree can not be ignored. One must choose carefully when to use the K-d tree and when to remain with the brute force \ac{kNN}


\section{Conclusion}
\label{sec:conclusion}

In this work, we have presented a novel framework to deal with data streams with concept drift called \ac{IncA-DES}, which combines concept drift detectors with dynamic selection of classifiers. An adaptive mechanism is used in order to keep more instances from a stable concept in the \ac{DSEW}, and drift detectors assist in the maintenance of information when a concept drift is triggered by shrinking the validation set. Additionally, we introduce two components to promote effectiveness in neighborhood-based \ac{DS} applications: 1) an Online K-d tree Algorithm for approximate neighborhood search, which employs lazy deletion to avoid inconsistencies generated with the removal of nodes, and 2) an overlap-based classification filter that leverages the \ac{kNN} algorithm when the neighbors of the test instance mostly agree with the class.

Experimental analysis has shown that the proposed framework had best overall accuracy than the 7 tested methods from the state of the art on 22 datasets in four different scenarios of label availability and delay of instances in data stream, and presented statistically significant difference to four methods on most scenarios. \ac{ARE}, a method from the state of the art, also presented good resilience when label delay was introduced, but \ac{IncA-DES} still was the most accurate.

In addition, the Online K-d tree algorithm improved the processing time of the proposed framework in some scenarios with a negligible loss in accuracy, and the overlap-based classification led to improvements in both accuracy and processing time. However, an open issue still involves the curse of dimensionality, which is a common issue for both brute force \ac{kNN} and K-d tree, in which distance functions lose their effectiveness under high-dimensional problems. To address this issue, feature selection techniques can be employed, or \ac{DES} approaches that do not rely on the neighborhood of the test instance can also be considered.

For future works, we intend to evaluate different training strategies, pursuing the best one in terms of diversity and generation of local experts for \ac{DS} in drifting scenarios. Furthermore, aiming to address the natural trend of labels being delayed and partial in the real world, we will also explore options of unsupervised, self-supervised, and semi-supervised learning, which may bring best results in such scenarios. Since the neighborhood search still seems able to evolve in streaming scenarios, we also intend to test different ways to do it in data streams, and to better address the curse of dimensionality. Additionally,, imbalanced data streams is also a constant problem under high-scale data streams. Futurely, we will explore alternatives to overcome this issue, such as oversampling or undersampling techniques \citep{jiao2022, Han2023}.

\subsubsection*{Acknowledgments}

This research was enabled in part by NSRC-Canada (Natural Sciences and Engineering Research Council of Canada), CAPES (Coordenação de Aperfeiçoamento de Pessoal de N\'ivel Superior), CNPq (Conselho Nacional de Pesquisa) and by support provided by the Digital Research Alliance of Canada (alliancecan.ca).


\appendix

\section{Hyperparameter Analysis}
\label{app:hyp}

This section presents the experiments performed to tune the hyperparameters of \ac{IncA-DES}. We start with an initial set of hyperparameters, shown in Table \ref{tab:initial-set-hyp}, which will be tuned individually throughout this section. The primary metric that we used for evaluating them is accuracy. The drift detector is not exposed because it is the first hyperparameter to be tuned. The Hoeffding Tree \citep{domingos2000} was defined as the base classifier.

\begin{table}[ht]
    \centering
    \caption{Initial set of Hyperparameters before the tuning. $F$ is the maximum training size of each classifier, $D$ is the maximum size of the pool of classifiers, $k$ the neighborhood size of the \ac{RoC} for the KNORA-Eliminate, and $\omega$ is the rate of the majority class in the neighborhood.}
    \label{tab:initial-set-hyp}
    \begin{tabular}{l c}
    \toprule
        Parameter & Value \\
        \midrule
        $F$ & 200 \\
        $D$ & 25 \\
        $k$ & 5 \\
        $\omega$ & 1 \\
        \bottomrule
    \end{tabular}
\end{table}

We assume that, as we use a K-d tree to perform neighborhood search, we can let the \ac{DSEW} grow more and worry less about the processing time for defining the \ac{RoC}. Thus, we did not set a limit for the maximum number of instances in the validation set. The only constraint, in this case, would be the memory. Other hyperparameters, such as the \ac{DS} method and the pruning engine, were fixed. We used the KNORA-Eliminate \citep{ko2008}, which is widely used in the literature and usually presents good results. As the pruning engine, we used the age-based, since \cite{almeidaEtAl2018} showed that it is less costly and leads to results similar to an accuracy-based one. The overlap-based classification and the Online K-d tree for defining the \ac{RoC}, components presented in this work, are also set as default.

\subsection{Concept Drift Detectors}

In this Section, we have tested 8 different drift detectors on \ac{IncA-DES}. The drift detectors used in these tests were introduced in section \ref{sec:stateoftheart}. Table \ref{tab:acc-dets-full} shows the average accuracy on all datasets. We see that the \ac{RDDM} drift detector achieved the highest average accuracy, even though it did not have individual victories over any of the tested datasets. Nevertheless, as the results indicates that it is the most consistent, it is chosen as the default drift detector for \ac{IncA-DES}.

\begin{table}[htb]
    \centering
    \caption{Average Accuracies of IncA-DES With Different Concept Drift Detectors.}
    \setlength{\tabcolsep}{2pt}
    \scriptsize
    \begin{tabular}{l |  c c | c c | c c | c c | c c | c c | c c | c c }
    \toprule
        Dataset & \multicolumn{2}{c|}{DDM} & \multicolumn{2}{c|}{EDDM} & \multicolumn{2}{c|}{RDDM} & \multicolumn{2}{c|}{STEPD} & \multicolumn{2}{c|}{ADWIN} & \multicolumn{2}{c|}{HDDMA} & \multicolumn{2}{c|}{HDDMW} & \multicolumn{2}{c}{FHDDM} \\
         & Acc. & \# Det. & Acc. & \# Det. & Acc. & \# Det. & Acc. & \# Det. & Acc. & \# Det. & Accuracy & \# Det. & Accuracy & \# Det. & Accuracy & \# Det. \\
    \midrule
        Nursery & 93.84 & 25.1 & \textbf{93.99} & 19.6 & 93.82 & 21.9 & 93.54 & 18.8 & 92.72 & 5 & 92.55 & 4.6 & 91.74 & 2 & 91.81 & 3 \\
        Electricity & 85.78 & 57.1 & \textbf{89.15} & 133.5 & 88.17 & 82.6 & 88.79 & 106 & 85.85 & 13.5 & 86.68 & 32.7 & 87.44 & 35.3 & 86.78 & 20.5 \\
        NOAA & 75.76 & 6.6 & 74.59 & 76.6 & 75.17 & 20.1 & 75.03 & 34.3 & \textbf{76.16} & 1 & 75.98 & 3 & 74.92 & 25.3 & 75.19 & 23 \\
        Digits & 93.18 & 0 & 91.54 & 3.4 & 93.14 & 0 & 89.71 & 8.3 & 93.17 & 0 & 93.10 & 0 & \textbf{93.21} & 0 & 93.14 & 0 \\ \hline
        Average & 86.95 & 22.2 & 87.15 & 58.3 & \textbf{87.58} & 31.2 & 86.84 & 41.9 & 87.09 & 4.9 & 87.16 & 10.1 & 86.72 & 15.7 & 86.68 & 11.6 \\
        \bottomrule
    \end{tabular}
    \label{tab:acc-dets-full}
\end{table}


\subsection{Maximum Training Size of Classifiers -- F}

In this Section, we compare the accuracies of different values for $F$, which dictates the maximum number of instances a classifier can receive for training. When $C_k$ has received $F$ instances for training, its training is interrupted, and another online classifier is added to $C$ and starts to be trained. Minor values for $F$ will help to add more classifiers in $C$ earlier in the stream, but may not give enough information. On the other hand, higher values for $F$ may lead to a slower building of a diverse pool of classifiers. The results are exposed in Table \ref{tab:acc-f-parameter}. As we can see, $F=200$ got the best average accuracy and is set as the default.

\begin{table}[htb]
    \centering
    \caption{Accuracies of IncA-DES With Different Values for $F$.}
    \begin{tabular}{l c c c c c}
        \toprule
        Dataset & 50 & 100 & 200 & 500 & 1000 \\
        \midrule
        Nursery & 92.55 & 92.99 & 93.87 & \textbf{94.81} & 94.59 \\
        Electricity & 87.12 & 87.54 & \textbf{88.40} & 87.56 & 87.82 \\
        NOAA & \textbf{75.50} & 75.38 & 75.27 & 75.07 & 74.91 \\
        Digits & 91.86 & 92.74 & \textbf{93.15} & 92.83 & 92.70 \\
        \hline
        Average & 86.76 & 87.19 & \textbf{87.63} & 87.57 & 87.51 \\
        \bottomrule
    \end{tabular}
    \label{tab:acc-f-parameter}
\end{table}

\subsection{Size of Pool of Classifiers -- D}

The $D$ parameter also brings trade-offs regarding drift adaptation since we use an age-based pruning engine. Higher values of $D$ maintain older classifiers for longer, which still would be candidates for classifying incoming instances. Remember that this would not happen if we were using an accuracy-based pruning engine, but the need to check the performance of classifiers would require more processing time. Maintaining old classifiers for longer may be beneficial when we have recurrent concept drifts, in which their information would be useful again when the concept returns. The $D$ parameter aligned with the $F$ parameter dictates both the diversity and the longevity of a concept in $C$ (remember, if we use an age-based pruning engine). The results of different $D$ values are in Table \ref{tab:acc-d-parameter}. We see that $D=100$ got the higher accuracy. However, since the difference to $D=75$ was not significant, we will choose to use $D=75$ to provide quicker adaptation and processing time for the framework.


\begin{table}[htb]
    \centering
    \caption{Accuracies of IncA-DES With Different Values for D.}
    \begin{tabular}{l c c c c}
    \toprule
        Dataset & 25 & 50 & 75 & 100 \\
        \midrule
        Nursery & \textbf{93.87} & 92.86 & 93.81 & \textbf{93.87} \\
        Electricity & 88.40 & 88.51 & 88.86 & \textbf{88.90} \\
        NOAA & \textbf{75.27} & 75.26 & 75.17 & \textbf{75.27} \\
        Digits & 93.15 & 93.14 & 93.15 & \textbf{93.18} \\
        \hline
        Average & 87.67 & 87.69 & 87.75 & \textbf{87.81} \\
        \bottomrule
    \end{tabular}
    \label{tab:acc-d-parameter}
\end{table}

\subsection{Neighborhood Size -- k}

Now, we test the impact of the neighborhood size $k$ of the \ac{RoC} for the KNORA-Eliminate in the framework. This hyperparameter dictates the size of the local regions where the classifiers' competence is measured, which plays a crucial part in neighborhood-based \ac{DS} frameworks \citep{almeidaEtAl2018, cavalheiro2021}.

Table \ref{tab:acc-k-parameter} shows the average accuracy for each value, where we see that $k=5$ was the most accurate one. Thus, it is set as the default. However, we must mention that the optimal size for the region of interest can vary according to the data distribution of the dataset and on how the local regions are distributed.

\begin{table}[htb]
    \centering
    \caption{Accuracies of IncA-DES With Different Values for $k$.}
    \begin{tabular}{l c c c c}
    \toprule
        Dataset & 5 & 7 & 9 & 11 \\
        \midrule
        Nursery & \textbf{93.81} & 93.74 & 93.58 & 93.60 \\
        Electricity & \textbf{88.86} & 88.67 & 88.12 & 88.10 \\
        NOAA & 75.17 & 75.63 & 75.87 & \textbf{76.87} \\
        Digits & \textbf{93.15} & 92.68 & 92.22 & 91.90 \\
        \hline
        Average & \textbf{87.75} & 87.68 & 87.45 & 87.47 \\
        \bottomrule
    \end{tabular}
    \label{tab:acc-k-parameter}
\end{table}

\subsection{Rate of Overlap -- \texorpdfstring{$\omega$}{ }}

In this section, we analyse the $\omega$ parameter, which dictates when the \ac{DS} method will be used to classify an instance, or if the majority class of the \ac{RoC} will be used to label the test instance. The results are in Table \ref{tab:acc-omega-parameter}, where we see that $\omega=0.8$ has achieved the highest accuracy, i.e., if 4 out of the 5 neighbors in the \ac{RoC} belong to the same class, that class is used to label the test instance. Thus, allowing a bit of overlap in the \ac{RoC} in the classification filter seems to bring benefits, as we can better leverage the \ac{kNN} algorithm. That said, $\omega=0.8$ is set as default in \ac{IncA-DES}.

\begin{table}[htb]
    \centering
    \caption{Accuracies of IncA-DES With Different Values for $\omega$.}
    \begin{tabular}{l c c}
    \toprule
        Dataset & 1 & 0.8 \\
        \midrule
        Nursery & 93.81 & \textbf{93.94} \\
        Electricity & \textbf{88.86} & 88.25  \\
        NOAA & 75.17 & \textbf{75.57} \\
        Digits & 93.15 & \textbf{93.98} \\
        \hline
        Average & 87.75 & \textbf{87.94} \\
        \bottomrule
    \end{tabular}
    \label{tab:acc-omega-parameter}
\end{table}

\section{State-of-the-art Hyperparameters}
\label{app:art-hyp}

We have separated the hyperparameters from Dynse to other state-of-the-art methods for some reasons. Firstly, as the authors had different configurations for different data sizes, we have performed our own tuning for Dynse for some hyperparameters on the same datasets from \ac{IncA-DES}. The tuned hyperparameters were $D$, $M$, and $B$. Other values were left as the default value, as well as the usage of two different values of $M$ for different types of concept drift (\textit{real} or \textit{virtual}) done by the authors. Secondly, Dynse has more hyperparameters than the other state-of-the-art methods, which most of them do not share. Dynse's hyperparameters are shown in Table \ref{tab:dynse-hyp}.

The hyperparameters of the other state-of-the-art methods were maintained as the default configuration and are in Table \ref{tab:art-param}.

\begin{table}[htb]
    \centering
    \footnotesize
    \caption{Dynse Hyperparameters}
    \begin{tabular}{l c}
        \toprule
        Hyperparameter & Value \\
        \midrule
        $D$ & 100 \\
        $M$ & 2/32 \\
        $B$ & 200 \\
        Pruning Engine & Age-based \\
        DS Method & KNORAE \\
        $k$ & 5 \\
        \bottomrule
    \end{tabular}
    \label{tab:dynse-hyp}
\end{table}

\begin{table}[htb]
    \centering
    \footnotesize
    \caption{Hyperparameters of other state-of-the-art methods.}
    \begin{tabular}{c c c c}
    \toprule
        Method & \multicolumn{3}{c}{Hyperparameters} \\
        \midrule
        \multirow{2}{*}{\ac{DynEd}} & $D$ & $W$ & Drift Detector \\
         & 1000 & 1000 & ADWIN \\ \hline
        \multirow{2}{*}{\ac{ARF}} & $D$ & Drift Detector \\
         & 100 & ADWIN \\ \hline
         \multirow{2}{*}{OzaBag} & $D$ \\
         & 10 \\ \hline
         \multirow{2}{*}{LevBag} & $D$ \\
         & 10 \\ \hline
         \multirow{2}{*}{OAUE} & $D$ & $W$ \\
         & 10 & 500 \\ \hline
         
    \end{tabular}    
    \label{tab:art-param}
\end{table}

\section{Table of Processing Time of the State-of-the-art Methods}
\label{app:time-art}

Table \ref{tab:time-art} shows the processing time in seconds of all tested methods in this work.

\begin{table}[hb]
\centering
\footnotesize
\caption{Average instances per second of IncA-DES and methods of the state of the Art.}
\begin{tabular}{l c c c c c c c c c}
    \toprule
    Dataset & IncA-DES & ARE & Dynse & DynEd & ARF & OzaBag & LevBag & OAUE \\ \midrule
        Ozone & 1,743 & 550 & 3,708 & 42 & 142 & \textbf{10,618} & 2,539 & 11,124 \\
        Gas Sensor & 667 & 236 & 806 & 13 & 86 & \textbf{1,785} & 981 & 1,186 \\
        Adult & 202 & 900 & 3,875 & 174 & 301 & \textbf{24,737} & 6,660 & 10,303 \\ 
        Rialto & 574 & 761 & 1,169 & 47 & 235 & \textbf{7,000} & 3,719 & 4,127 \\ 
        Keystroke & 3,500 & 1,333 & 11,667 & 196 & 625 & \textbf{17,500} & 7,778 & 46,667 \\ 
        Insects-AB & 227 & 784 & 1,148 & 26 & 244 & \textbf{6,937} & 3,512 & 4,091 \\ 
        Insects-GB & 264 & 822 & 1,702 & 37 & 266 & \textbf{6,709} & 3,507 & 4,246 \\ 
        Insects-IB & 43 & 668 & 1,140 & 38 & 236 & \textbf{6,584} & 3,300 & 3,785 \\ 
        Yeast & 3,669 & 1,284 & 11,673 & 102 & 600 & 18,343 & 9,171 & \textbf{64,200} \\ 
        Asfault & 630 & 347 & 1,829 & 29 & 160 & \textbf{3,894} & 2,103 & 3,250 \\ 
        Covertype & 854 & 635 & 1,236 & 21 & 231 & \textbf{10,645} & 4,313 & 5,347 \\ 
        Pen Digits & 342 & 1,296 & 872 & 45 & 262 & \textbf{8,919} & 4,072 & 5,996 \\ 
        Dry Bean & 2,521 & 2,447 & 736 & 67 & 471 & \textbf{11,764} & 5,683 & 7,577 \\ 
        Rice & 10,939 & 3,505 & 3,471 & 232 & 1,068 & \textbf{36,100} & 14,440 & 32,818 \\ 
        Letters & 134 & 857 & 507 & 9 & 121 & 3,867 & 1,735 & \textbf{2,215} \\ 
        SineRec & 9,301 & 4,971 & 10,671 & 73 & 1,591 & \textbf{50,251} & 27,352 & 37,398 \\ 
        SeaRec & 15,029 & 3,270 & 6,773 & 129 & 958 & \textbf{67,568} & 31,827 & 49,776 \\ 
        StaggerRec & 6,987 & 16,396 & 10,436 & 57 & 5,230 & 93,545 & 69,881 & \textbf{98,814} \\ 
        Agrawal & 747 & 1,123 & 7,048 & 48 & 336 & \textbf{24,522} & 9,659 & 15,840 \\ 
        Hyperplane & 71 & 1,353 & 4,572 & 123 & 369 & \textbf{38,610} & 11,520 & 18,797 \\ 
        LED & 52 & 1,820 & 1,872 & 58 & 573 & \textbf{17,452} & 1,355 & 9,302.33 \\ 
        RandomRBF & 88 & 1,486 & 4,671 & 130 & 541 & 37,175 & 11,136 & 17,182 \\ \hline
        Average & 2663 & 2129 & 4163 & 77 & 669 & \textbf{22933} & 10738 & 20638 \\
        \bottomrule
\end{tabular}
\label{tab:time-art}
\end{table}

\section{Tables of Accuracies for Delayed Labels}
\label{app:acc-delay}

Tables \ref{tab:acc-art-delay50}, \ref{tab:acc-art-delay200} and \ref{tab:acc-art-delay1000} show the average accuracies for \ac{IncA-DES} and all of the tested state-of-the-art with delays equal to 50, 200, and 1,000, respectively.

\begin{table}[!h]
    \centering
    \caption{Accuracies of IncA-DES and Methods of the State of the Art With Delayed and Partial Labels (Delay = 50).}
    \small
    \begin{tabular}{l c c c c c c c c c c}
        \toprule
        Dataset & IncA-DES & ARE & Dynse & DynEd & ARF & OzaBag & LevBag & OAUE \\ 
        \midrule
        Ozone & 91.79 & 93.75 & \textbf{93.96} & 93.00 & 93.84 & \textbf{93.96} & 93.16 & \textbf{93.96} \\ 
        Gas Sensor & \textbf{71.53} & 64.86 & 69.08 & 66.15 & 61.32 & 58.12 & 61.90 & 55.57 \\ 
        Adult & 83.05 & 82.94 & 82.54 & 82.91 & 83.45 & 84.21 & 84.08 & \textbf{84.25} \\ 
        Rialto & \textbf{40.94} & 40.76 & 39.70 & 39.14 & 34.23 & 28.41 & 30.60 & 30.59 \\ 
        Keystroke & \textbf{94.71} & 92.24 & 78.99 & 80.10 & 90.79 & 71.43 & 83.07 & 56.21 \\ 
        Insects-AB & \textbf{73.85} & 71.97 & 70.08 & 72.37 & 71.43 & 54.86 & 66.43 & 62.43 \\ 
        Insects-GB & \textbf{76.23} & 74.58 & 71.74 & 73.36 & 74.23 & 56.92 & 68.43 & 60.15 \\ 
        Insects-IB & 60.59 & 64.51 & 59.63 & 62.39 & \textbf{64.53} & 49.54 & 59.52 & 55.40 \\ 
        Yeast & \textbf{48.86} & 43.99 & 45.74 & 38.60 & 34.35 & 45.95 & 44.55 & 34.81 \\ 
        Asfault & 85.39 & 83.40 & 81.63 & 84.25 & \textbf{85.51} & 70.60 & 78.19 & 69.99 \\ 
        Covertype & 88.82 & \textbf{92.42} & 91.90 & 91.68 & 91.76 & 80.30 & 88.57 & 85.06 \\ 
        Pen Digits & \textbf{92.72} & 90.22 & 89.05 & 83.06 & 89.40 & 81.55 & 85.73 & 81.32 \\ 
        Dry Bean & 88.17 & \textbf{88.88} & 87.84 & 88.33 & 87.99 & 88.78 & 88.38 & 85.76 \\ 
        Rice & 90.71 & 91.39 & 89.32 & 90.21 & \textbf{91.82} & 90.52 & 91.53 & 83.22 \\ 
        Letters & 74.76 & 38.29 & \textbf{74.81} & 55.24 & 49.20 & 59.15 & 58.98 & 59.31 \\ 
        SineRec & 92.44 & 91.49 & 88.22 & \textbf{98.20} & 95.02 & 55.95 & 90.77 & 84.99 \\ 
        SEARec & 86.34 & 86.12 & 86.29 & \textbf{88.64} & 86.84 & 84.26 & 86.34 & 85.97 \\ 
        StaggerRec & \textbf{99.58} & 99.07 & 97.23 & 99.20 & 99.29 & 71.40 & 98.42 & 95.51 \\ 
        Agrawal & 81.95 & 78.59 & 77.75 & \textbf{93.08} & 77.31 & 61.32 & 77.71 & 81.38 \\ 
        Hyperplane & 86.72 & 87.32 & \textbf{89.20} & 83.75 & 86.05 & 87.41 & 87.09 & 87.92 \\
        LED & 68.92 & \textbf{73.66} & 69.22 & 72.32 & 73.65 & 73.56 & 73.62 & 73.43 \\
        RandomRBF & 93.28 & \textbf{93.48} & 86.11 & 91.64 & 93.73 & 86.38 & 91.91 & 89.58 \\ \hline
        Average & \textbf{80.40} & 78.27 & 77.85 & 78.42 & 78.03 & 69.13 & 76.70 & 72.36 \\
        \bottomrule
    \end{tabular}
    \label{tab:acc-art-delay50}
\end{table}

\begin{table}[!htb]
    \centering
    \caption{Accuracies of IncA-DES and Methods of the State of the Art With Delayed and Partial Labels (Delay = 200).}
    \small
    \begin{tabular}{l c c c c c c c c c c}
        \toprule
        Dataset & IncA-DES & ARE & Dynse & DynEd & ARF & OzaBag & LevBag & OAUE \\
        \midrule
        Ozone & 91.94 & \textbf{93.96} & 93.95 & 93.40 & \textbf{93.96} & \textbf{93.96} & 93.87 & \textbf{93.96} \\ 
        Gas Sensor & 56.06 & 51.39 & \textbf{57.22} & 51.85 & 48.52 & 55.50 & 50.75 & 52.82 \\ 
        Adult & 82.91 & 82.89 & 82.47 & 82.84 & 83.89 & 84.13 & 84.03 & \textbf{84.15} \\ 
        Rialto & 34.07 & \textbf{35.05} & 33.01 & 35.04 & 27.41 & 27.46 & 24.38 & 28.16 \\ 
        Keystroke & \textbf{94.42} & 89.59 & 74.01 & 69.80 & 88.36 & 67.64 & 77.14 & 49.14 \\ 
        Insects-AB & \textbf{72.10} & 69.94 & 67.99 & 70.52 & 70.65 & 54.53 & 65.69 & 61.63 \\ 
        Insects-GB & \textbf{74.62} & 73.04 & 69.98 & 70.89 & 72.93 & 55.31 & 67.30 & 59.48 \\ 
        Insects-IB & 60.22 & \textbf{64.30} & 59.46 & 62.62 & 64.07 & 49.14 & 58.87 & 55.14 \\ 
        Yeast & 47.27 & \textbf{47.70} & 45.97 & 31.20 & 35.44 & 45.79 & 43.61 & 35.12 \\ 
        Asfault & \textbf{82.72} & 80.70 & 78.87 & 80.17 & 82.55 & 68.13 & 76.01 & 66.51 \\ 
        Covertype & 88.88 & \textbf{92.10} & 91.48 & 91.32 & 91.68 & 80.07 & 88.53 & 84.94 \\ 
        Pen Digits & \textbf{91.04} & 87.88 & 87.64 & 79.57 & 86.84 & 79.50 & 83.43 & 79.17 \\ 
        Dry Bean & 86.59 & 88.02 & 85.24 & 87.39 & 86.15 & \textbf{88.03} & 87.20 & 84.70 \\ 
        Rice & 88.66 & \textbf{89.89} & 85.23 & 88.27 & 89.48 & 88.82 & 88.35 & 80.16 \\ 
        Letters & 72.08 & 30.43 & \textbf{73.30} & 49.73 & 43.90 & 57.34 & 56.57 & 57.89 \\ 
        SineRec & 91.27 & 90.28 & 87.12 & \textbf{96.91} & 93.78 & 55.31 & 89.56 & 83.79 \\ 
        SeaRec & 86.19 & 85.99 & 86.18 & \textbf{88.51} & 86.69 & 84.24 & 86.21 & 85.85 \\ 
        StaggerRec & \textbf{98.67} & 98.15 & 96.31 & 98.26 & 98.37 & 71.24 & 97.52 & 94.59 \\ 
        Agrawal & 81.44 & 78.08 & 77.31 & \textbf{92.49} & 76.85 & 61.30 & 77.18 & 80.82 \\ 
        Hyperplane & 86.73 & 87.30 & 89.15 & 83.84 & 86.04 & 87.49 & 87.08 & \textbf{87.86} \\ 
        LED & 68.91 & \textbf{73.64} & 69.28 & 72.58 & 73.61 & 73.58 & 73.61 & 73.36 \\ 
        RandomRBF & 93.23 & 93.46 & 86.10 & 92.25 & \textbf{93.72} & 86.25 & 91.98 & 89.52 \\\hline 
        Average & \textbf{78.64} & 76.54 & 76.24 & 75.88 & 76.13 & 68.85 & 74.95 & 71.31 \\ 
        \bottomrule
    \end{tabular}
    \label{tab:acc-art-delay200}
\end{table}

\begin{table}[!htb]
    \centering
    \caption{Accuracies of IncA-DES and Methods of the State of the Art With Delayed and Partial Labels (Delay = 1000).}
    \small
    \begin{tabular}{l c c c c c c c c c}
        \toprule
        Dataset & IncA-DES & ARE & Dynse & DynEd & ARF & OzaBag & LevBag & OAUE \\
        \midrule
        Ozone & 92.14 & 93.63 & \textbf{93.99} & 92.65 & 93.83 & 93.96 & 93.83 & 93.96 \\ 
        Gas Sensor & 39.35 & 41.15 & 46.98 & 36.52 & 39.81 & \textbf{51.77} & 44.30 & 48.36 \\ 
        Adult & 83.01 & 82.93 & 82.50 & 82.34 & 83.81 & \textbf{84.05} & 84.04 & 84.03 \\ 
        Rialto & 25.20 & \textbf{31.67} & 23.84 & 28.63 & 19.16 & 26.71 & 18.50 & 26.90 \\ 
        Keystroke & \textbf{85.26} & 63.59 & 48.45 & 25.00 & 57.79 & 57.14 & 51.36 & 25.00 \\ 
        Insects-AB & \textbf{63.50} & 61.93 & 60.12 & 62.00 & 61.84 & 53.18 & 59.02 & 56.50 \\ 
        Insects-GB & \textbf{67.36} & 66.13 & 62.28 & 64.23 & 65.80 & 53.17 & 61.61 & 57.77 \\ 
        Insects-IB & 58.96 & \textbf{63.14} & 58.48 & 60.47 & 62.89 & 48.21 & 57.79 & 53.87 \\ 
        Yeast & \textbf{47.11} & 46.94 & 46.29 & 17.00 & 35.83 & 46.73 & 46.18 & 30.61 \\ 
        Asfault & \textbf{73.86} & 71.34 & 69.30 & 61.66 & 73.69 & 60.71 & 67.38 & 50.77 \\ 
        Covertype & 82.76 & \textbf{86.56} & 85.45 & 83.88 & 86.10 & 77.06 & 84.12 & 81.39 \\ 
        Pen Digits & \textbf{84.48} & 81.84 & 81.21 & 73.27 & 79.80 & 74.04 & 77.74 & 73.03 \\ 
        Dry Bean & 82.90 & 82.80 & 83.53 & 72.52 & 82.38 & \textbf{83.78} & 83.60 & 79.89 \\ 
        Rice & 80.13 & 81.30 & 80.64 & 76.24 & 79.12 & \textbf{82.51} & 80.96 & 68.88 \\ 
        Letters & 68.45 & 29.35 & \textbf{70.30} & 47.94 & 42.02 & 55.34 & 54.04 & 55.49 \\ 
        SineRec & 84.97 & 83.85 & 81.17 & \textbf{88.78} & 87.11 & 51.63 & 82.96 & 77.30 \\ 
        SeaRec & 85.43 & 85.29 & 85.55 & \textbf{87.69} & 85.89 & 84.14 & 85.46 & 85.20 \\ 
        StaggerRec & \textbf{93.78} & 93.26 & 91.42 & 93.27 & 93.48 & 70.41 & 92.63 & 89.68 \\ 
        Agrawal & 78.67 & 75.47 & 74.94 & \textbf{88.41} & 74.48 & 60.75 & 74.60 & 77.81 \\ 
        Hyperplane & 86.59 & 87.12 & \textbf{88.94} & 83.15 & 85.79 & 87.22 & 86.93 & 87.42 \\ 
        LED & 68.85 & \textbf{73.55} & 69.17 & 72.13 & 73.39 & 73.51 & 73.48 & 72.86 \\ 
        RandomRBF & 93.18 & 93.29 & 85.94 & 92.06 & \textbf{93.54} & 86.31 & 91.78 & 89.16 \\ \hline
        Average & \textbf{73.91} & 71.64 & 71.39 & 67.72 & 70.80 & 66.47 & 70.56 & 66.63 \\ 
        \bottomrule
    \end{tabular}
    \label{tab:acc-art-delay1000}
\end{table}

\end{document}